\documentclass{article}

\usepackage{float}
\usepackage{jheppub}
\usepackage{amsmath}
\usepackage{graphicx}
\usepackage{amssymb}
\usepackage{subfig}
\usepackage{natbib}
\usepackage{afterpage}

\usepackage[utf8]{inputenc}

\title{Learning Particle Physics by Example:\newline Location-Aware Generative Adversarial Networks for Physics Synthesis}
\author{Luke de Oliveira${}^{a}$,}
\author{Michela Paganini${}^{a,b}$, and}
\author{Benjamin Nachman${}^a$}

\affiliation{$^{a}$Lawrence Berkeley National Laboratory, 1 Cyclotron Rd, Berkeley, CA, 94720, USA}  
\affiliation{$^{b}$Department of Physics, Yale University, New Haven, CT 06520, USA}

\emailAdd{lukedeoliveira@lbl.gov, michela.paganini@yale.edu, bnachman@cern.ch}

\abstract {
We provide a bridge between generative modeling 
in the Machine Learning community and simulated physical processes in High Energy 
Particle Physics by applying a novel Generative Adversarial Network (GAN) architecture to the production of \textit{jet images} -- 2D representations of energy depositions from particles interacting with a calorimeter.
We propose a simple architecture, the Location-Aware Generative Adversarial Network, that learns to produce realistic radiation patterns from simulated high energy particle collisions. The pixel intensities of GAN-generated images faithfully span over many orders of magnitude and exhibit the desired low-dimensional physical properties (\textit{i.e.}, jet mass, n-subjettiness, etc.). We shed light on limitations, and provide a novel empirical validation of image quality and validity of GAN-produced simulations of the natural world. This work provides a base for further explorations of GANs for use in faster simulation in High Energy Particle Physics. 
}

\begin{document}

\maketitle

\section{Introduction}
\label{sec:intro}
The task of learning a generative model of a data distribution has long been a very difficult, yet rewarding research direction in the statistics and machine learning communities. Often positioned as a complement to discriminative models, generative models face a more difficult challenge than their discriminative counterparts, \textit{i.e.}, reproducing rich, structured distributions such as natural language, audio, and images. 
Deep learning-based generative models, with the promise of hierarchical, end-to-end learned features, are seen as one of the most promising avenues towards building generative models capable of handling the most rich and non-linear of data spaces. Generative Adversarial Networks~\cite{goodfellow2014generative}, a relatively new framework for learning a deep generative model, cast the task as a two player non-cooperative game between a generator network, $G$, and a discriminator network, $D$. The generator tries to produce samples that the discriminator cannot distinguish as fake when compared to real samples drawn from the data distribution, while the discriminator tries to correctly identify if a sample it is shown originates from the generator (fake) or if it was sampled from the data distribution (real).
When we relax $G$ and $D$ to be from the space of all functions, there exists a unique equilibrium where $G$ reproduces the true data distribution, and $D$ is 1/2 everywhere~\cite{goodfellow2014generative}. Most recent work on GANs has focused on the task of recovering data distributions from natural images; many recent advances in image generation using GANs have shown great promise in producing photo-realistic natural images at high resolution~\cite{goodfellow2014generative,odena_acgan,info_gan,improved_gan,conditional_gan,semisup_gan,dcgan,whatwheredraw,text2im,stackgan} while attempting to solve many known problems with stability~\cite{improved_gan,dcgan} and lack of convergence guarantees~\cite{distinguishability}.

With the growing complexity of theoretical models and power of computers, many scientific and engineering projects increasingly rely on detailed simulations and generative modeling. This is particularly true for High Energy Particle Physics where precise Monte Carlo (MC) simulations are used to model physical processes spanning distance scales from $10^{-25}$ meters all the way to the macroscopic distances of detectors. For example, both the ATLAS~\cite{Aad:2010ah} and CMS~\cite{Bayatian:922757} collaborations model the detailed interactions of particles with matter to accurately describe the detector response. These {\it full simulations} are based on the \textsc{Geant4} package~\cite{Agostinelli:2002hh} and are time\footnote{Full simulation can take up to $\mathcal{O}(\text{min/event})$.} and CPU intensive, a significant challenge given that $\mathcal{O}(10^9)$ events/year must be simulated. In fact, simulating this expansive dynamical range with the required precision is very expensive. Various approximations can be used to make faster simulations~\cite{Edmonds:2008zz,ATLAS:1300517,Abdullin:2011zz}, but the time is still non-negligible and they are not applicable in all physics applications. Another potential bottleneck is the modeling of quark and gluon interactions on the smallest distance scales (matrix element calculations).  Calculations with a large number of final state objects (e.g. higher order in the perturbative series) are very time consuming and can compete with the detector simulation for the total event generation time. Recent work focusing on algorithmic improvements that leverage High Performance Computing capabilities have helped combat long generation times~\cite{Childers:2015pwh}. However, not all processes are massively parallelizable and time at supercomputers is a scarce resource.  

Our goal is to develop a new paradigm for fast event generation in high energy physics through the use of GANs. We start by tackling a constrained version of the larger problem, where we use the concept of a {\it jet image}~\cite{Cogan:2014oua} to show that the idealized 2D radiation pattern from high energy quarks and gluons can be efficiently and effectively reproduced by a GAN. This work builds upon recent developments in jet image classification with deep neural networks~\cite{deOliveira:2015xxd,Almeida:2015jua,Komiske:2016rsd,Barnard:2016qma,Baldi:2016fql}. By showing that generated jet images resemble the true simulated images in physically meaningful ways, we demonstrate the aptitude of GANs for future applications. 

This paper is organized as follows. Sections~\ref{sec:simulation} and~\ref{sec:gan} provide a brief introductions to the physics of jet images and structure of Generative Adversarial Networks, respectively.  New GAN architectures developed specifically for jet images are described in Sec.~\ref{sec:arch}.  The results from our model are shown in Sec.~\ref{sec:studies}, with an extensive discussion of what the neural network is learning. The paper concludes with Sec.~\ref{sec:conclusion}.

\section{Dataset}
\label{sec:simulation}

Jets are the observable result of quarks and gluons scattering at high energy. A collimated stream of protons and other hadrons forms in the direction of the initiating quark or gluon. Clusters of such particles are called jets.  A jet image is a two-dimensional representation of the radiation pattern within a jet: the distribution of the locations and energies of the jet's constituent particles.  Jet formation is finished well before it can be detected, so it is sufficient to consider the radiation pattern in a two dimensional surface spanned by orthogonal angles\footnote{While the azimuthal angle $\phi$ is a real angle, pseudorapidity $\eta$ is only approximately equal to the polar angle $\theta$.  However, the radiation pattern is nearly symmetric in $\phi$ and $\eta$ and so these standard coordinates are used to describe the jet constituent locations.} $\eta$ and $\phi$. The jet image consists of a regular grid of pixels in $\eta\times\phi$. This is analogous to a calorimeter without longitudinal segmentation (e.g. the CMS detector~\cite{Chatrchyan:2008aa}). Adding layers, either through longitudinal segmentation or combining information from multiple detectors, has recently been studied for classification~\cite{Komiske:2016rsd} and generation~\cite{Paganini:2017hrr}, but is beyond the scope of this paper. 

Jet images are constructed and pre-processed using the setup described in Ref.~\cite{deOliveira:2015xxd} and briefly summarized in this section.  The finite granularity of a calorimeter is simulated with a regular $0.1\times 0.1$ grid in $\eta$ and $\phi$.  The energy of each calorimeter cell is given by the sum of the energies of all particles incident on the cell.  Cells with positive energy are assigned to jets using the anti-$k_t$ clustering algorithm~\cite{antiktpaper} with a radius parameter of $R=1.0$ via the software package \textsc{FastJet} 3.2.1~\cite{fastjet}.  To mitigate the contribution from the underlying event, jets are are trimmed~\cite{trimming} by re-clustering the constituents into $R=0.3$ $k_t$ subjets and dropping those which have less than $5\%$ of the transverse momentum of the parent jet. Trimming also reduces the impact of \textit{pileup}: multiple proton-proton collisions occurring in the same event as the hard-scatter process.  Jet images are formed by translating the $\eta$ and $\phi$ of all constituents of a given jet so that its highest $p_\text{T}$ subjet is centered at the origin.  A rectangular grid of $\eta\times\phi\in[-1.25,1.25]\times[-1.25,1.25]$ with $0.1\times 0.1$ pixels centered at the origin forms the basis of the jet image.  The intensity of each pixel is the $p_\text{T}$ corresponding to the energy and pseudorapditiy of the constituent calorimeter cell, $p_\text{T}=E_\text{cell}/\cosh(\eta_\text{cell})$.  The radiation pattern is symmetric about the origin of the jet image and so the images are rotated\footnote{For more details about this rotation, which slightly differs from Ref.~\cite{deOliveira:2015xxd}, see Appendix~\ref{app:image_process}.}. The subjet with the second highest $p_\text{T}$ (or, in its absence, the direction of the first principle component) is placed at an angle of $-\pi/2$ with respect to the $\eta-\phi$ axes.  Finally, a parity transform about the vertical axis is applied if the left side of the image has more energy than the right side.

\begin{figure}[h!]
    \centering
    \includegraphics[width=0.5\textwidth]{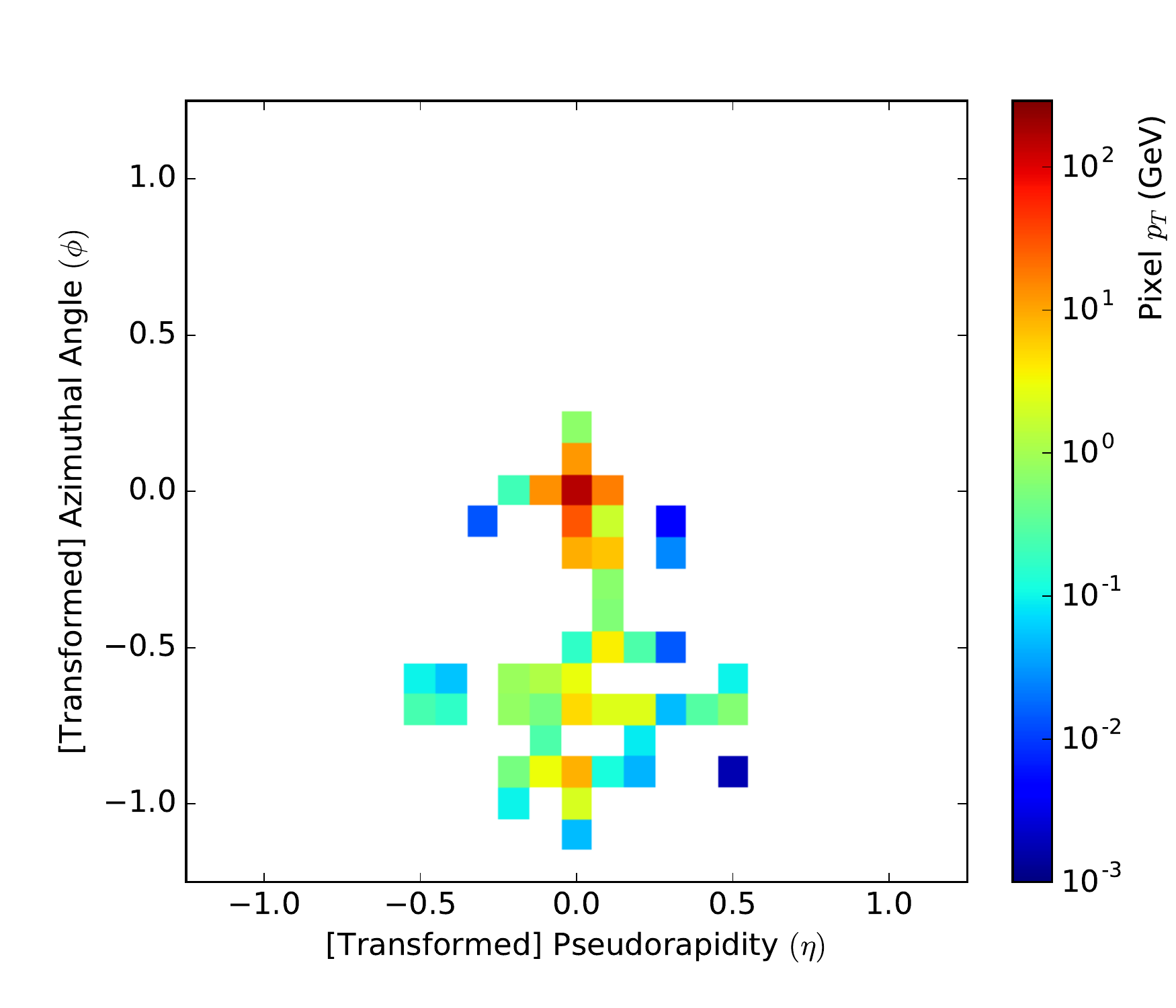}
    \caption{A typical jet image.}
    \label{fig:jetimage}
\end{figure}

Jet images vary significantly depending on the process that produced them. One high profile classification task is the separation of jets originating from high energy $W$ bosons (signal) from generic quark and gluon jets (background). Both signal and background are simulated using \textsc{Pythia} 8.219~\cite{Pythia8,Pythia} at $\sqrt{s}=14$ TeV.  In order to mostly factor the impact of the jet transverse momentum ($p_\text{T}$), we focus on a particular range: $250$~GeV~$<p_\text{T}^\text{jet}<300$~GeV.  A typical jet image from the simulated dataset~\cite{dataset} is shown in Fig.~\ref{fig:jetimage}. The image has already been processed so that the leading subjet is centered at the origin and the second highest $p_\text{T}$ subjet is at $-\pi/2$ in the translated $\eta-\phi$ space. Unlike natural images from commonly studied datasets, jet images do not have smooth features and are highly sparse (typical occupancy $\sim 10\%$).  This will necessitate a dedicated GAN setup, described in Sec.~\ref{sec:arch}.

\section{Generative Adversarial Networks}
\label{sec:gan}

Let $\mathcal{I}$ be the sample space -- in this application, the space of grey-scale square images of given dimensions. Let the space of naturally
occurring samples, according to a data distribution $f$, be $\mathcal{N}\subseteq\mathcal{I}$ where we do not know the true form of $\mathcal{N}\sim f$ (i.e., $f$ is the distribution we wish to recover).

A generator is a (potentially stochastic) function $G: z\mapsto \mathcal{S}$, where $z\sim \mathcal{N}(\mu, \sigma^2)$ is a 
latent vector governing the generation process, and $\mathcal{S}$ is the space 
of synthesized examples, described as 

\begin{equation*}
\label{eqn:synth_space}
  \mathcal{S}\subseteq\mathcal{I},\ \mathcal{S} = \{I\in\mathcal{I} : \exists z,\  G(z) = I\}.
\end{equation*}

A discriminator is a map 
$D: \mathcal{I}\mapsto (0, 1)$, which assigns to any sample $I\in\mathcal{I}$ a probability of being fake, i.e., $D(I) = 0$, or real, i.e., $D(I) = 1$. The loss of the system can be expressed as 

\begin{equation}
\label{eqn:gan_formulation}
  \mathcal{L} = 
    \underbrace{
        \mathbb{E}[\log(\mathbb{P}(D(I)=0\ \vert\  I\in \mathcal{S}))]
    }_{\begin{subarray}{l} 
        \text{term associated with the discriminator}\\
        \text{perceiving a generated sample as fake}
       \end{subarray}}
    + 
    \underbrace{
        \mathbb{E}[\log(\mathbb{P}(D(I)=1\ \vert\  I\in \mathcal{N}))]
    }_{\begin{subarray}{l}
        \text{term associated with the discriminator}\\
        \text{perceiving a real sample as real}
       \end{subarray}}
\end{equation}

The first term of the loss function is associated with the discriminator classifying a GAN-generated sample as fake, while the second term is associated with it classifying a sample drawn from the data distribution as real. Equation \ref{eqn:gan_formulation} is minimized by the generator and maximized by the discriminator. Note that the generator only directly affects the first term during backpropagation. At training time, gradient descent steps are taken in an alternating fashion between the generator and the discriminator. Due to this and to the fact that GANs are usually parametrized by non-convex players, GANs are considered to be quite unstable without any guarantees of convergence~\cite{distinguishability}. To combat this, a variety of ad-hoc procedures for controlling convergence have emerged in the field, in particular, relating to the generation of natural images. Architectural constraints and optimizer configurations introduced in~\cite{dcgan} have provided a well studied set of defaults and starting points for training GANs. 

Many newer improvements also help avoid \textit{mode collapse}, a very common point of failure for GANs where a generator learns to generate a single element of $\mathcal{S}$ that is maximally confusing for the discriminator. For instance, mini-batch discrimination and feature matching~\cite{improved_gan} allow the discriminator to use batch-level features and statistics, effectively rendering mode collapse suboptimal for the generator. It has also been empirically shown ~\cite{whatwheredraw, odena_acgan, info_gan, improved_gan} that adding auxiliary tasks seems to reduce such a tendency and improve convergence stability; \textit{side information} is therefore often used as either additional information to the discriminator / generator~\cite{improved_gan, conditional_gan, whatwheredraw}, as a quantity to be reconstructed by the discriminator~\cite{semisup_gan, info_gan}, or both~\cite{odena_acgan}.

\section{Location-Aware Generative Adversarial Network(LAGANs)}
\label{sec:arch}

We introduce a series of architectural modifications to the DCGAN~\cite{dcgan} frameworks in order to take advantage of the pixel-level symmetry properties of jet images while explicitly inducing location-based feature detection. 

We build an auxiliary task into our system following the ACGAN~\cite{odena_acgan} formulation. In addition to the primary task where the discriminator network must learn to identify fake jet images from real ones, the discriminator is also tasked with jointly learning to classify boosted $W$ bosons (signal) and QCD (background), with the scope of learning the conditional data distribution. 

\begin{figure}[h!]
    \centering
    \includegraphics[width=0.6\textwidth]{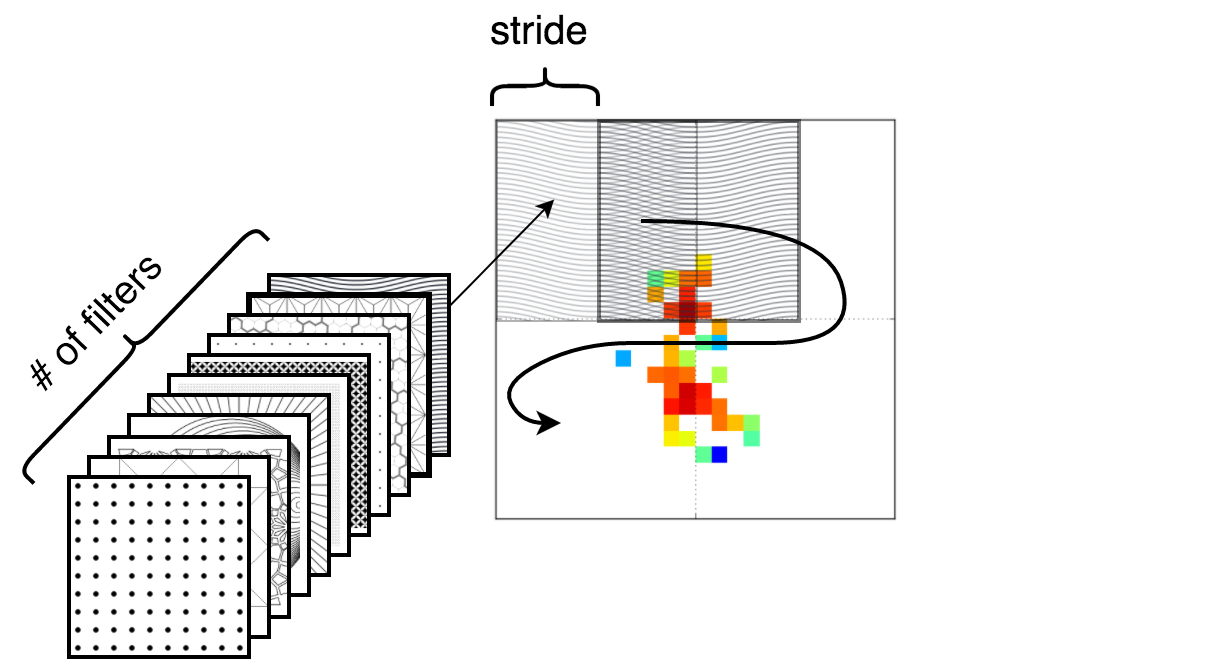}
    \caption{In the simplest (\textit{i.e.}, all-square) case, a convolutional layer consists of $N$ filters of size $F\times F$ sliding across an $L\times L$ image with stride $S$. For a \texttt{valid} convolution, the dimensions of the output volume will be $W\times W\times N$, where $W = (L-F)/S +1$.}
    \label{fig:cnn}
    \vspace*{\floatsep}
    \centering
    \includegraphics[width=0.7\textwidth]{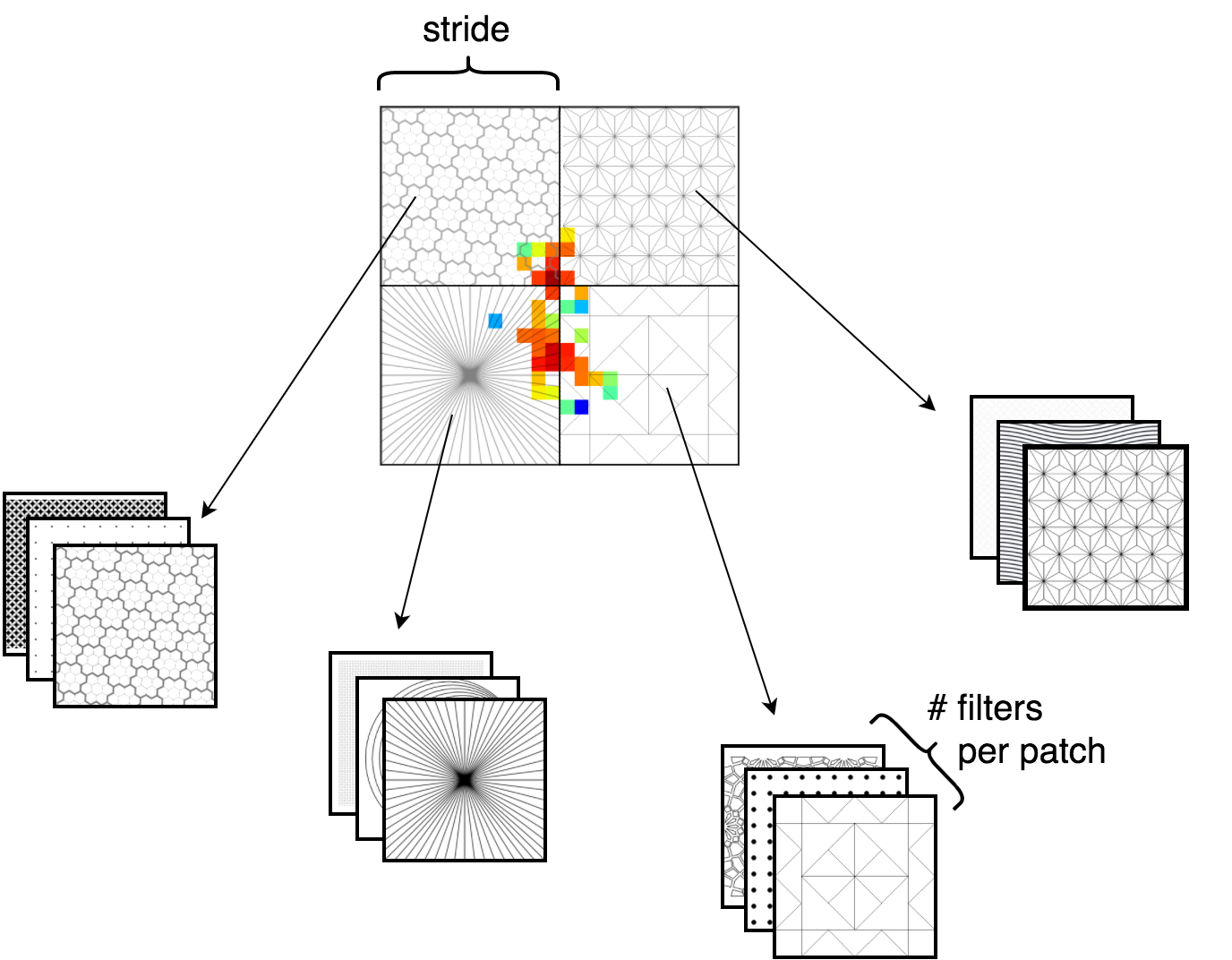}
    \caption{A locally connected layer consists of $N$ unique filters applied to \emph{each} individual patch of the image. Each group of $N$ filters is specifically learned for one patch, and no filter is slid across the entire image. The diagram shows the edge case in which the stride $S$ is equal to the filter size $F$, but in general patches would partially overlap. A convolution, as described above, is simply a locally connected layer with a weight sharing constraint.}
    \label{fig:lcn}
\end{figure}

We design both the generator and discriminator using a single 2D convolutional layer followed by 2D locally connected layers without weights sharing. Though locally connected layers are rarely seen as useful components of most computer vision systems applied to natural images due to their lack of translation invariance and parameter efficiency, we show that the location specificity of locally connected layers allows these transformations to outperform their convolutional counterparts on tasks relating to jet-images, as previously shown in~\cite{deOliveira:2015xxd,Baldi:2016fql}. Diagrams to better understand the nature of convolutional and locally connected layers are available in Figures~\ref{fig:cnn} and~\ref{fig:lcn}.

In preliminary investigations, we found experimental evidence showing the efficacy of \emph{fully connected networks} in producing the central constituents of jet images (which is reminiscent of the findings of \cite{deOliveira:2015xxd} on the discriminative power of fully connected networks applied to jet images). We also observed that fully convolutional systems excel at capturing the less location-specific low energy radiation dispersion pattern. This informed the choice of 2D locally connected layers to obtain a parameter-efficient model while retaining location-specific information.

As an additional experiment, we also trained a multi-headed model, essentially employing a \emph{localization network} and a \emph{dispersion network} where each stream learns a portion of the data distribution that reflects its comparative advantage. Though we do not present results from this model due to its ad-hoc construction, we provide code and model weights to analyze~\cite{code}.

Since jet images represent a fundamental challenge for most GAN-inspired architectures due to the extreme levels of sparsity and unbounded nature of pixel levels~\cite{Cogan:2014oua}, we apply additional modifications to the standard DCGAN framework. To achieve sparsity, we utilize Rectified Linear Units~\cite{maas2013rectifier} in the last layer of the generator, even though these activations are not commonplace in most GAN architectures due to issues with sparse gradients~\cite{how_to_train}. However, we remain consistent with~\cite{dcgan,how_to_train} and use Leaky Rectified Linear Units~\cite{leaky} throughout both the generator and discriminator. We also apply minibatch discrimination~\cite{improved_gan} on the last layer of features learned by the discriminator, which we found was crucial to obtain both stability and sample diversity in the face of an extremely sparse data distribution. Both batch normalization~\cite{batchnorm} and label flipping~\cite{improved_gan,how_to_train} were also essential in obtaining stability in light of the large dynamic range.

In summary, a Location Aware Generative Adversarial Network (LAGAN) is a set of guidelines for learning GANs designed specifically for applications in a sparse regime, when location within the image is critical, and when the system needs to be end-to-end differentiable, as opposed to requiring hard attention. Examples of such applications, in addition to the field of High Energy Physics, could include medical imaging, geological data, electron microscopy, etc. The characteristics of a LAGAN can be summarized as follows:

\begin{itemize}
    \item \textbf{Locally Connected Layers} - or any attentional component where we can attend to location specific features - to be used in the generator and the discriminator
    
    \item \textbf{Rectified Linear Units} in the last layer to induce sparsity
    
    \item \textbf{Batch normalization}, as also recommended in~\cite{dcgan}, to help with weight initialization and gradient stability
    
    \item \textbf{Minibatch discrimination}\cite{improved_gan}, which experimentally was found to be crucial in modeling both the high dynamic range and the high levels of sparsity
    
\end{itemize}

\subsection{Architecture Details, Implementation, and Training} 
\label{ssub:implementation_and_training} 
\begin{figure}[h!]
    \centering
    \includegraphics[width=\textwidth]{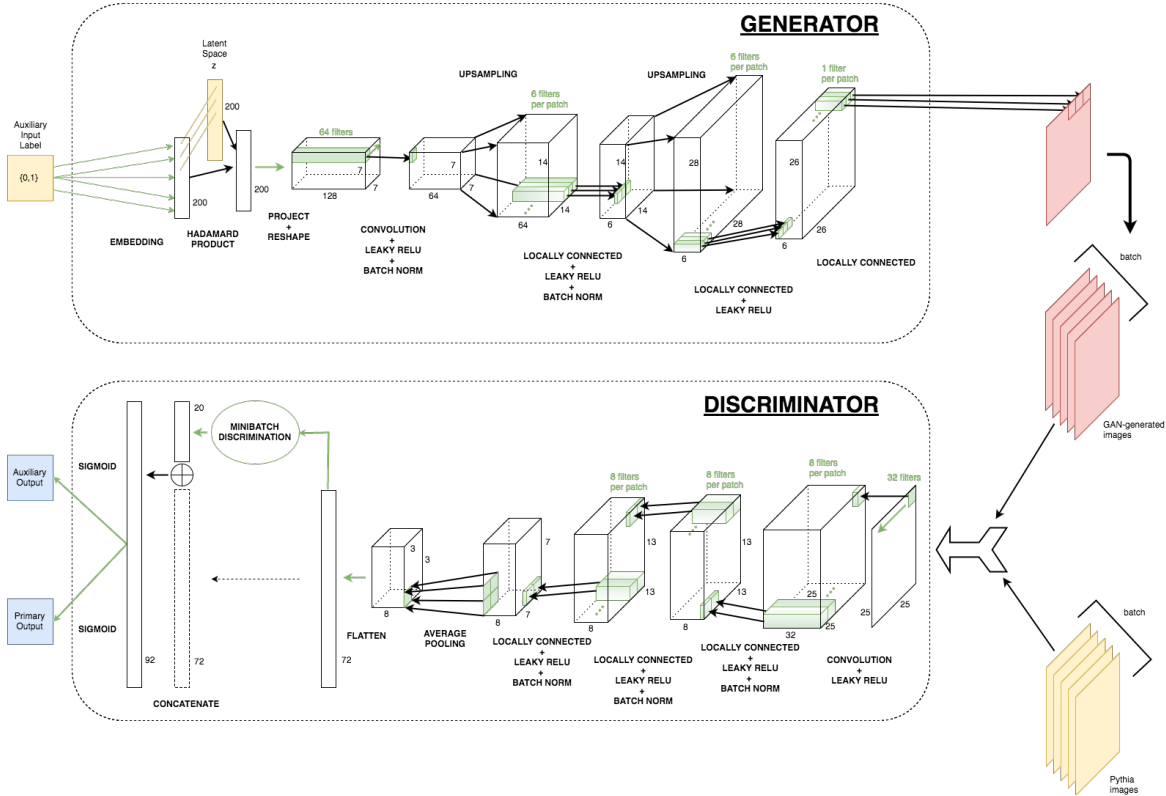}
    \caption{LAGAN architecture}
    \label{fig:diagram}
\end{figure}
A diagram of the architecture is available in Fig.~\ref{fig:diagram}.

We utilize low-dimensional vectors $z\in\mathbb{R}^{200}$ as our latent space, where, $z\sim\mathcal{N}(0, I)$, with final generated outputs occupying $\mathbb{R}^{25\times 25}_{\geq 0}$. 

Before passing in $z$, we perform a hadamard product between $z$ and a trainable lookup-table embedding of the desired class (boosted $W$, or QCD), effectively conditioning the generation procedure~\cite{odena_acgan}.

The generator consists of a \texttt{same}-bordered 2D convolution, followed by two \texttt{valid}-bordered 2D locally connected layer, with 64, 6, and 6 feature maps respectively. We use receptive fields of size $5\times 5$ in the convolutional layer, and $5\times 5$ and $3\times 3$ in the two locally connected layers respectively. Sandwiched between the layers are 2x-upsampling operations and channel-wise batch normalization~\cite{batchnorm} layers. On top of the last layer, we place a final ReLU-activated locally connected layer with 1 feature map, a $2\times 2$ receptive field, and no bias term. 

The discriminator consists of a \texttt{same}-bordered 2D convolutional layer with 32 $5 \times 5$ unique filters, followed by three \texttt{valid}-bordered 2D locally connected layers all with 8 feature maps with receptive fields of size $5\times 5$, $5\times 5$, and $3\times 3$. After each locally connected layer, we apply channel-wise batch normalization. We use the last feature layer as input to a minibatch discrimination operation with twenty 10-dimensional kernels. These batch-level features are then concatenated with the feature layer before being mapped using sigmoids for both the primary and auxiliary tasks.

At training time, label flipping alleviates the tendency of the GAN to produce very signal-like and very background-like samples, and attempts to prevent saturation at the label extremes. We flip labels according to the following scheme: when training the discriminator, we flip 5\% of the target labels for the primary classification output, as well as 5\% of the target labels for the auxiliary classification output on batches that were solely fake images, essentially tricking it into misclassifying fake versus real images in the first case, and signal versus background GAN-generated images in the second case; in addition, while training the generator, 9\% of the time we ask it to produce signal images that the discriminator would classify as background, and vice versa.

We train the system end-to-end by taking alternating steps in the gradient direction for both the generator and the  discriminator. We employ the Adam~\cite{adam} optimizer, utilizing the sensible parameters outlined in~\cite{dcgan} with a batch size of 100 for 40 epochs. We construct all models using Keras~\cite{keras} and TensorFlow~\cite{tensorflow2015-whitepaper}, and utilize two NVIDIA\textsuperscript{\textregistered} Titan X (Pascal) GPUs for training. 

\section{Generating Jet Images with Adversarial Networks}
\label{sec:studies}
The proposed LAGAN architecture is validated through quantitative and qualitative means on the task of generating realistic looking jet images. 
In this section, we generate 200k jet images and compare them to 200k Pythia images to evaluate - both quantitatively and qualitatively - their content (\ref{ssec:images}), to explore the powerful information provided by some of the most representative images (\ref{ssec:500}), to dig deeper into the inner workings of the generator (\ref{ssec:gen}) and discriminator (\ref{ssec:disc}), to monitor the development of the training procedure (\ref{ssec:train}), to compare with other architectures (\ref{ssec:arch}), and to briefly evaluate computational efficiency of the proposed method (\ref{ssec:inference}).

\subsection{Image Content Quality Assessment}
\label{ssec:images}

Quantifying the efficacy of the generator is challenging because of the dimensionality of the images. However, we can assess the performance of the network by considering low-dimensional, physically inspired features of the 625 dimensional image space. Furthermore, by directly comparing images, we can visualize what aspects of the radiation pattern the network is associating with signal and background processes, and what regions of the image are harder to generate via adversarial training.

\begin{figure}[H]
\centering
\includegraphics[width=0.45\textwidth]{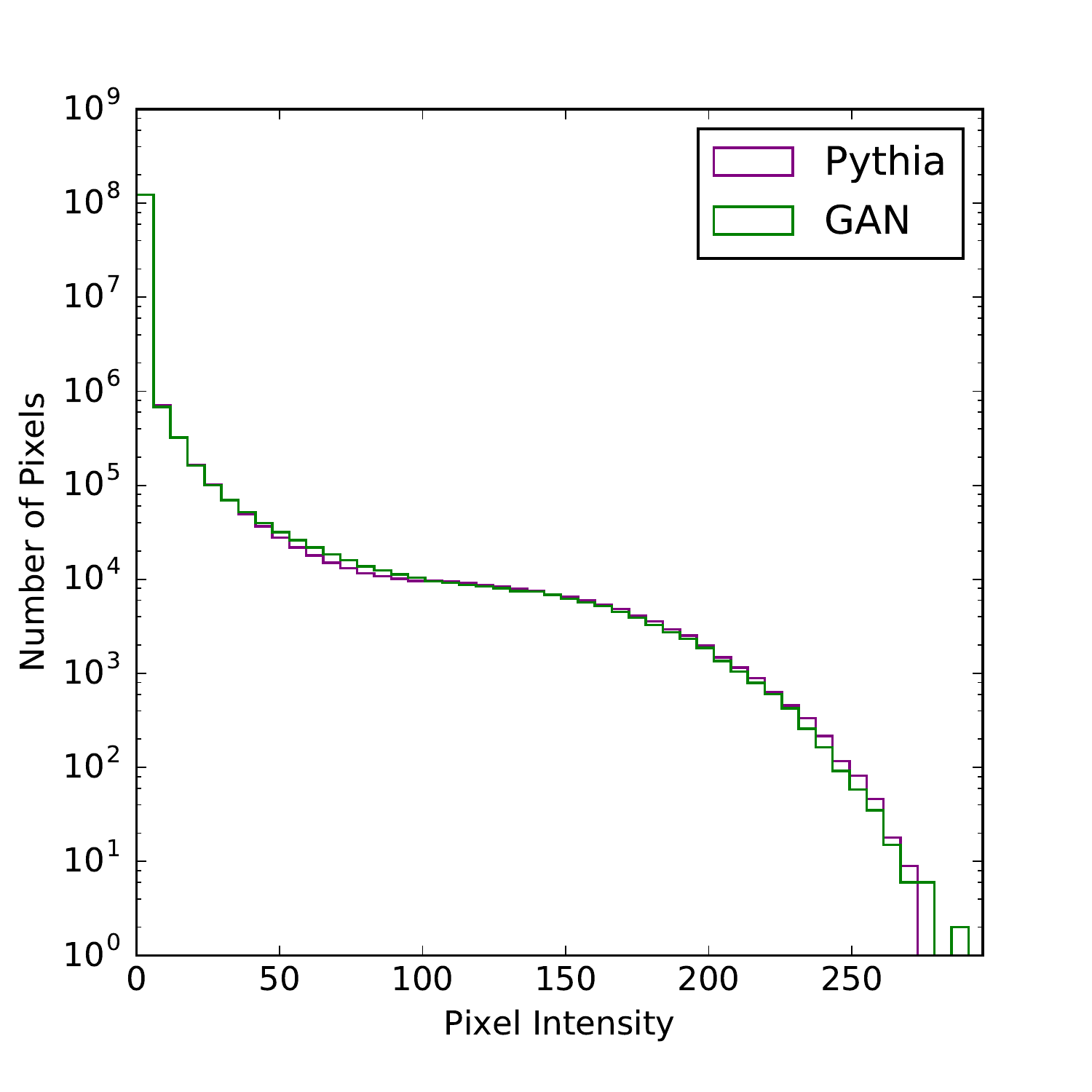}
\caption{The distribution of the pixel intensities for both fake (GAN) and real (Pythia) jet images, aggregated over all pixels in the jet image.}
\label{fig:energy_of_pixels}
\end{figure}

The first one-dimensional quantity to reproduce is the distribution of the pixel intensities aggregated over all pixels in the image. Intensities span a wide range of values, from the energy scale of the jet $p_\text{T}$ ($\mathcal{O}(100)$ GeV) down to the machine epsilon for double precision arithmetic. Because of inherent numerical degradation in the preprocessing steps
\footnote{Bicubic spline interpolation in the rotation process causes a large number of pixels to be interpolated between their original value and zero, the most likely intensity value of neighboring cells. Though a zero-order interpolation would solve sparsity problems, we empirically determine that the loss in jet-observable resolution is not worth the sparsity preservation. A more in-depth discussion can be found in Appendix~\ref{app:image_process}.}, 
images acquire unphysical low intensity pixel values. We truncate the distribution at $10^{-3}$ and discard all unphysical contributions.
Figure~\ref{fig:energy_of_pixels} shows the distribution of pixel intensities for both Pythia and GAN-generated jet images. The full physical dynamic range is explored by the GAN and the high $p_T$ region is faithfully reproduced.

One of the unique properties of high energy particle physics is that we have a library of useful {\it jet observables} that are physically motivated functions $f:\mathbb{R}^{25\times 25}\rightarrow\mathbb{R}$ whose features are qualitatively (and in some cases, quantitatively) well-understood. We can use the distributions of these observables to assess the abilty of the GAN to mimic Pythia.  Three such features of a jet image $I$ are the mass $m$, transverse momentum $p_\text{T}$, and $n$-subjettiness $\tau_{21}$~\cite{nsub}:

\begin{align}
p_\text{T}^2(I) &= \left(\sum_i I_i\cos(\phi_i)\right)^2+\left(\sum_i I_i\sin(\phi_i)\right)^2\\
m^2(I) &= \left(\sum_i I_i\right)^2-p_\text{T}^2(I)-\left(\sum_iI_i\sinh(\eta_i)\right)^2\\
\tau_{n}(I)&\propto\sum_i I_i\text{min}_a\left(\sqrt{\left(\eta_i-\eta_a\right)^2+\left(\phi_i-\phi_a\right)^2}\right),\\
\tau_{21}(I)&=\tau_2(I)/\tau_1(I),
\end{align}

\noindent where $I_i$, $\eta_i$, and $\phi_i$ are the pixel intensity, pseudorapidity, and azimuthal angle, respectively.  The sums run over the entire image.  The quantities $\eta_a$ and $\phi_a$ are axis values determined with the one-pass $k_t$ axis selection using the winner-take-all combination scheme~\cite{Larkoski:2014uqa}.  

\begin{figure}[H]
\centering
\includegraphics[width=0.333\textwidth]{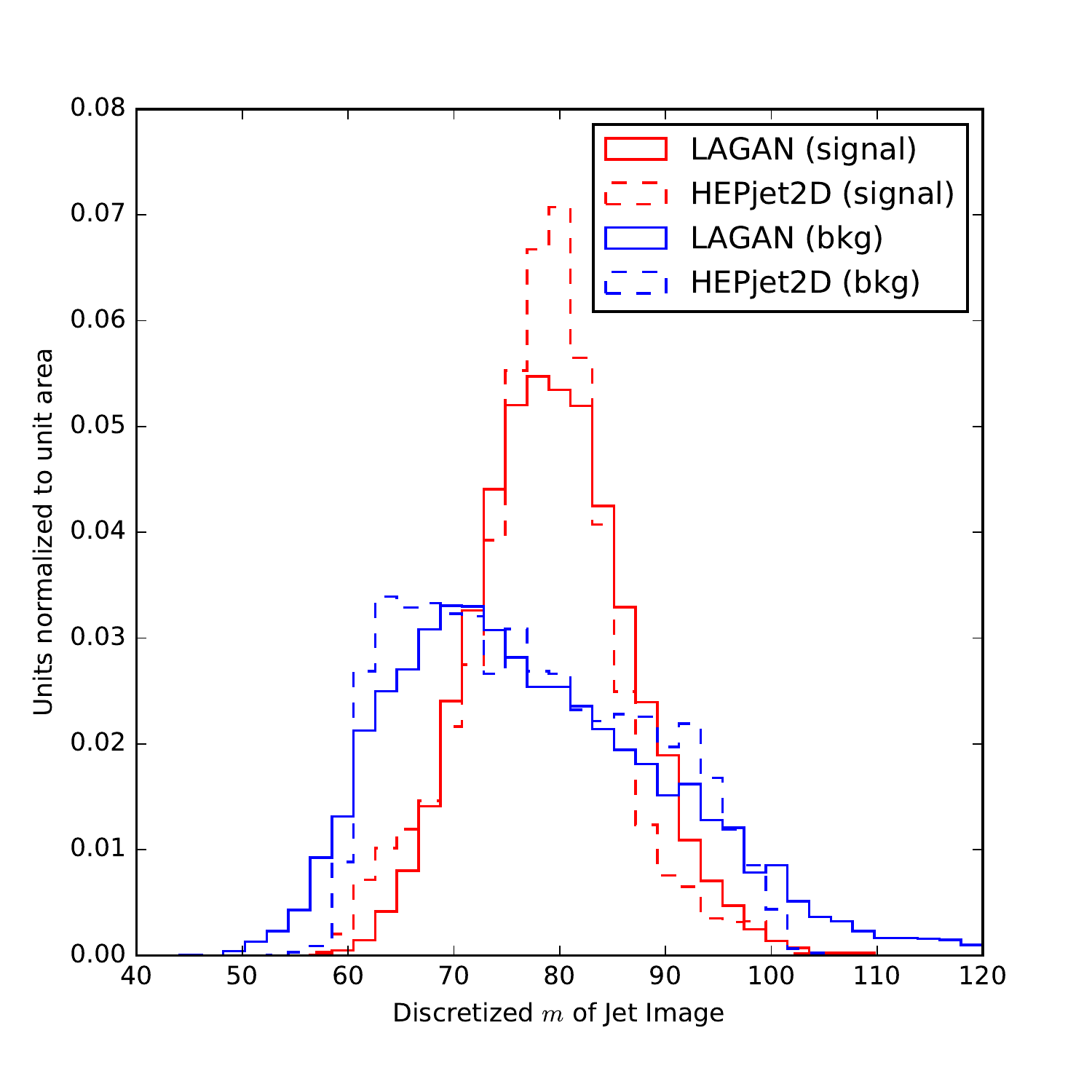}\includegraphics[width=0.333\textwidth]{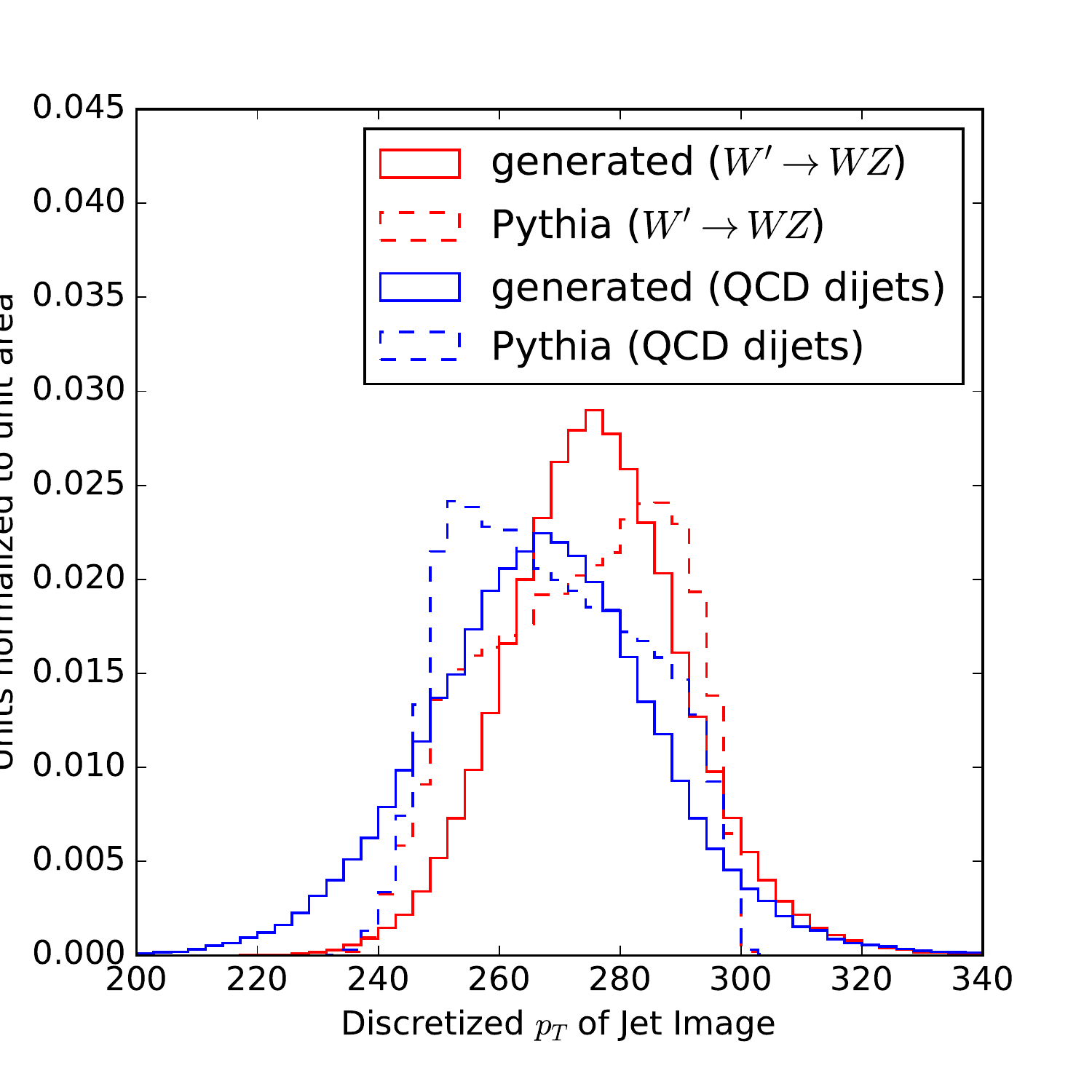}\includegraphics[width=0.333\textwidth]{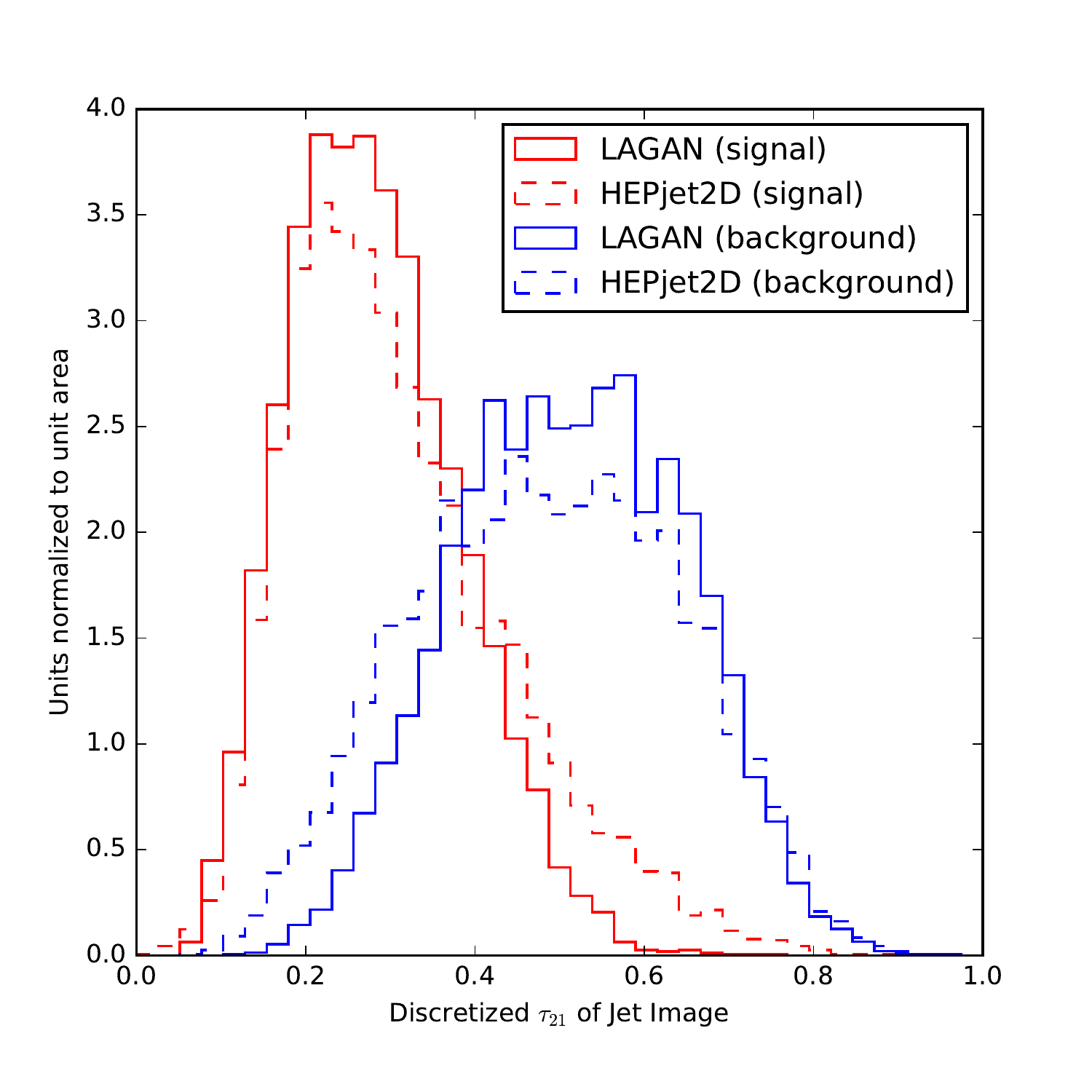}
\caption{The distributions of image mass $m(I)$, transverse momentum $p_\text{T}(I)$, and $n$-subjettiness $\tau_{21}(I)$.  See the text for definitions.}
\label{fig:m_pt_tau21}
\end{figure}

The distributions of $m(I), p_\text{T}(I)$, and $\tau_{21}(I)$ are shown in Fig.~\ref{fig:m_pt_tau21} for both GAN and Pythia images. 
These quantities are highly non-linear, low dimensional manifolds of the 625-dimensional space in which jet images live, so there is no guarantee that these non-trivial mappings will be preserved under generation. However this property is desirable and easily verifiable.
The GAN images reproduce many of the jet-observable features of the Pythia images. Shapes are nearly matched, and, for example, signal mass exhibits a peak at $\sim 80$GeV, which corresponds to the mass of the $W$ boson that generates the hadronic shower. This is an emergent property - nothing in the training or architecture encourages this. Importantly, the generated GAN images are as diverse as the true Pythia images used for training - the fake images do not simply occupy a small subspace of credible images.

We claim that the network is not only learning to produce samples with a diverse range of $m$, $p_T$ and $\tau_{21}$, but it's also internally learning these projections of the true data distribution and making use of them in the discriminator. To provide evidence for this claim, we explore the relationships between the $D$'s primary and auxiliary outputs, namely $P(\mathrm{real})$ and $P(\mathrm{signal})$, and the physical quantities that the generated images possess, such as mass $m$ and transverse momentum $p_T$.

\begin{figure}[h]
\centering
\includegraphics[width=0.3\textwidth]{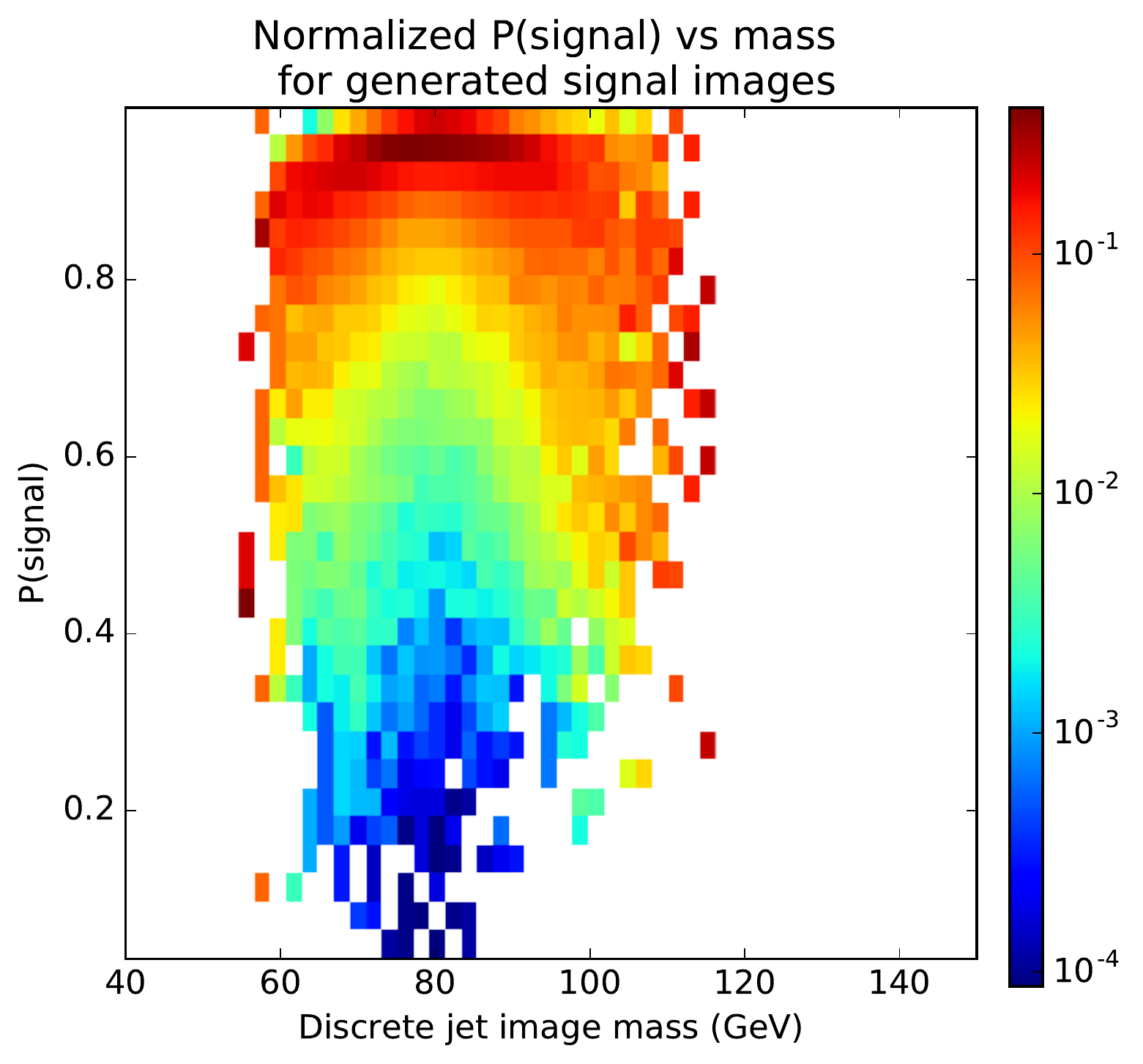}\hspace{10mm}\includegraphics[width=0.3\textwidth]{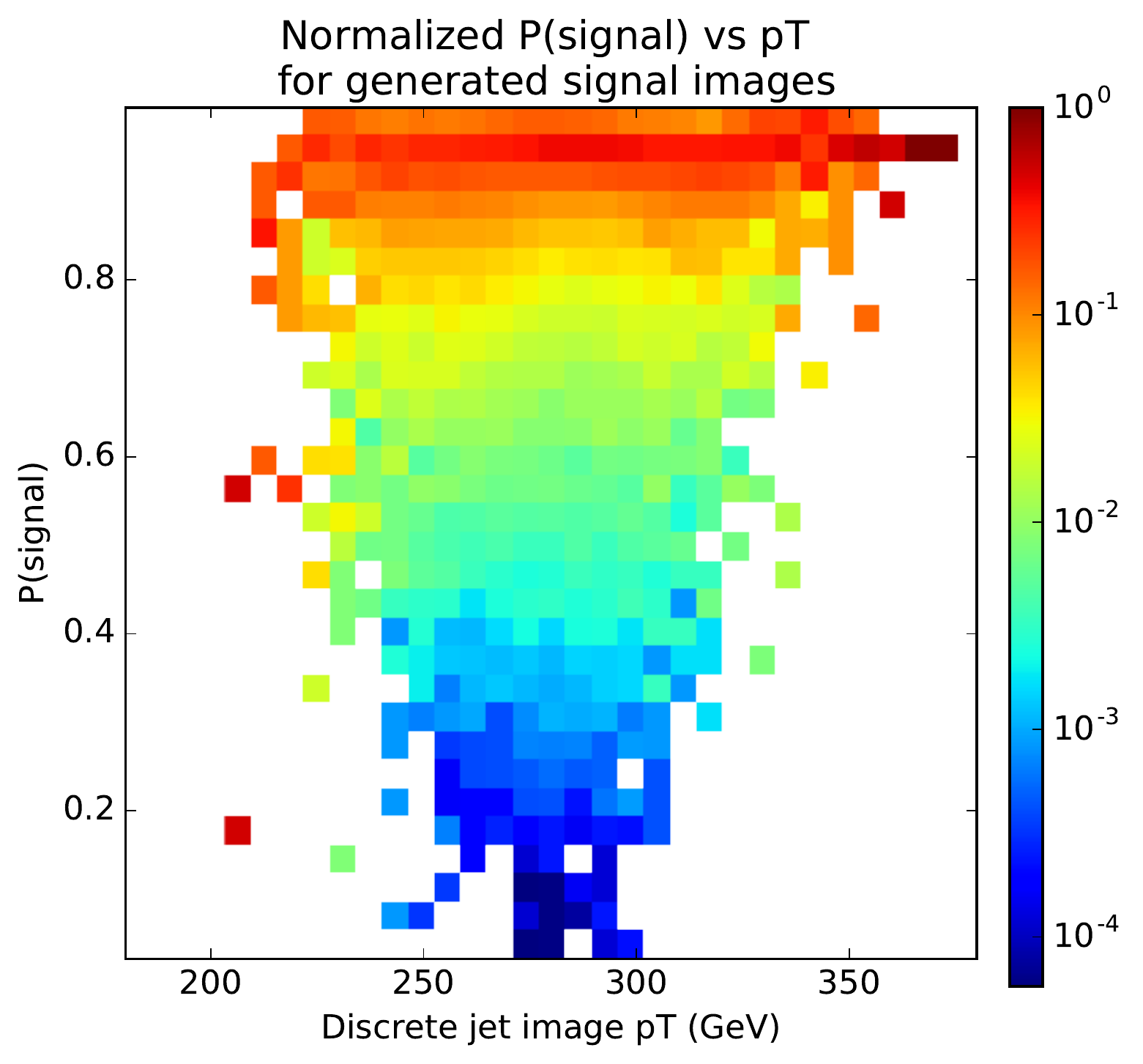} \\
\includegraphics[width=0.3\textwidth]{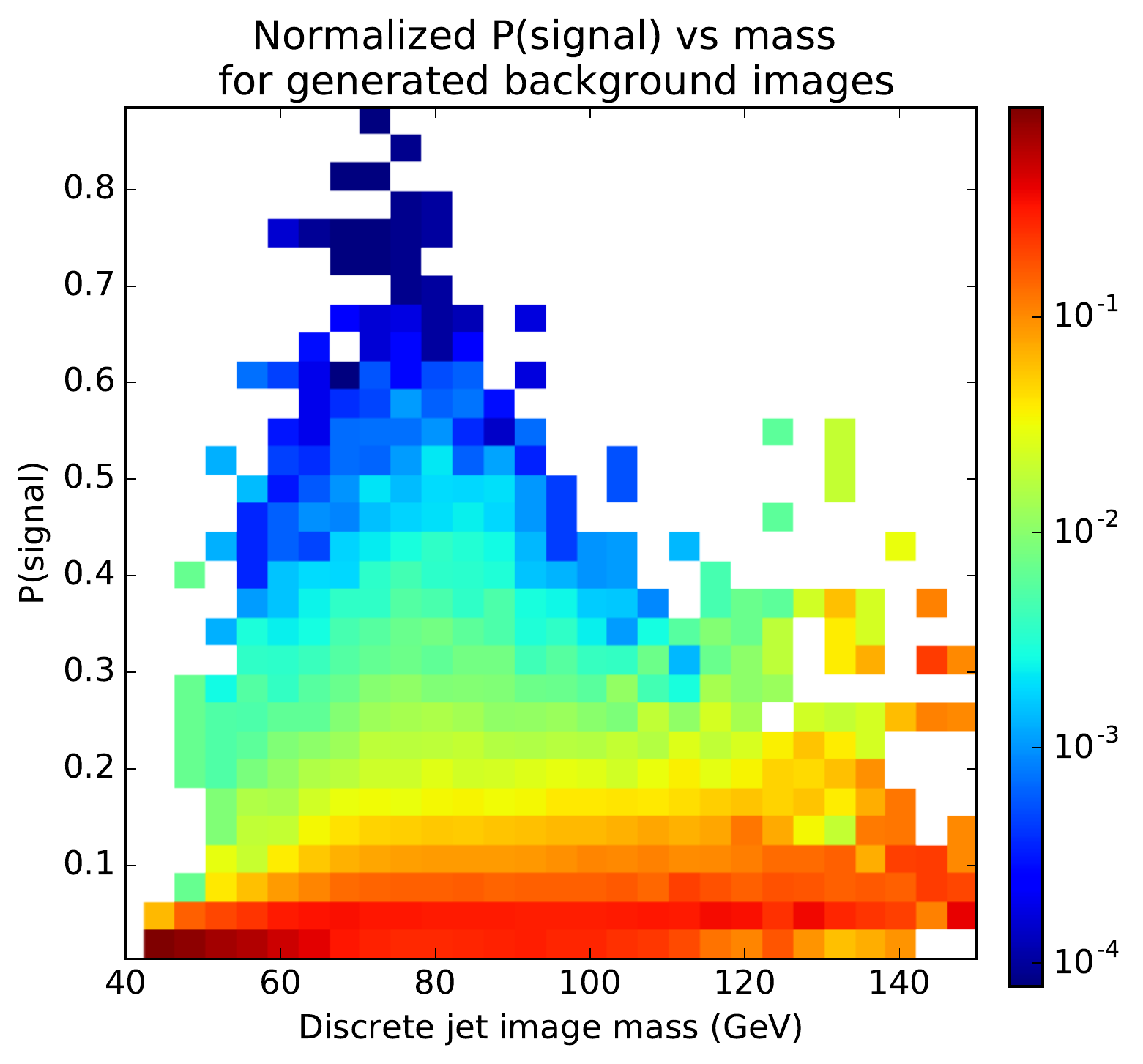}\hspace{10mm}\includegraphics[width=0.3\textwidth]{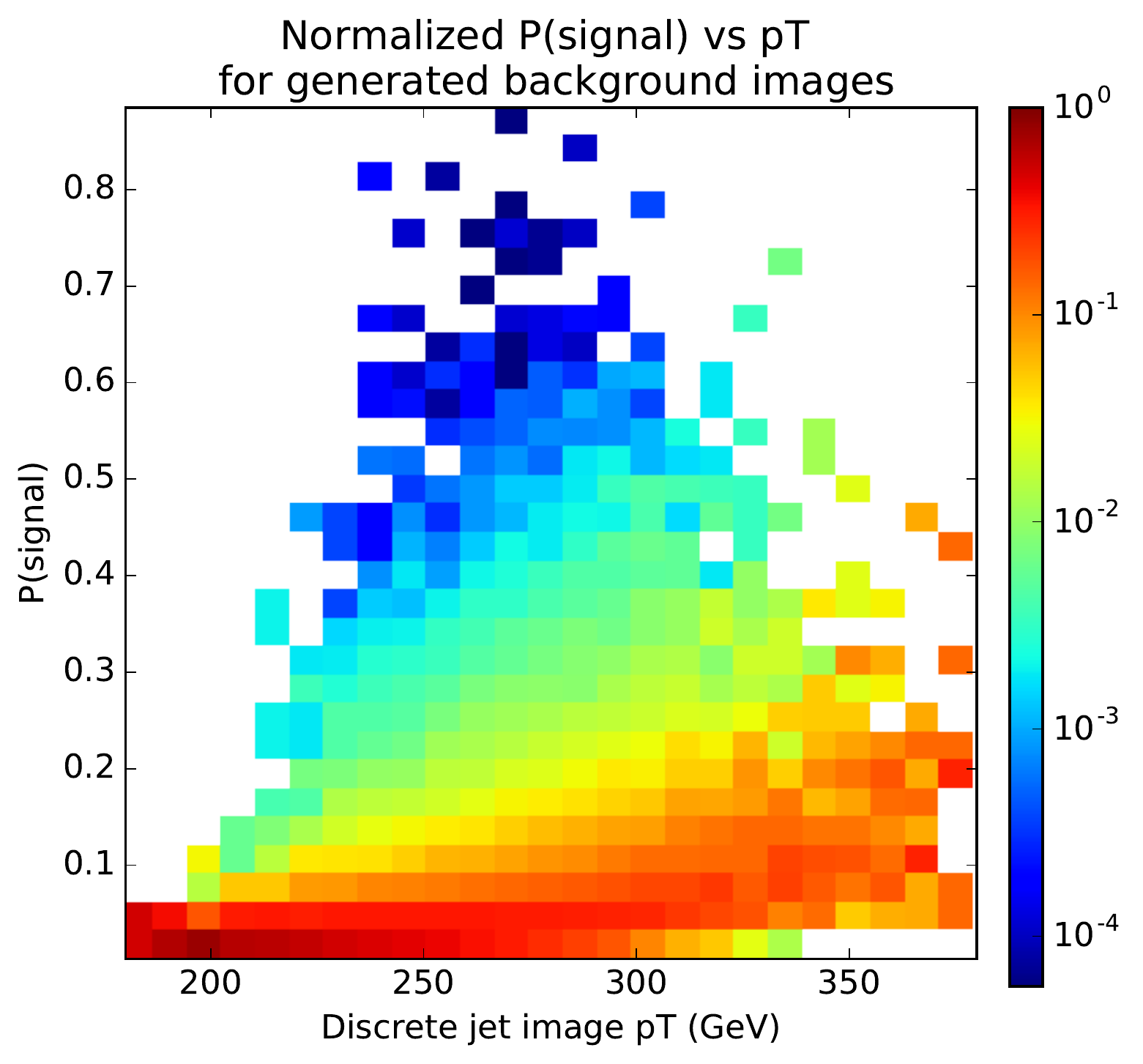} \\
\caption{Auxiliary discriminator output as a function of mass, on the left, and as a function of transverse momentum, on the right. The top plots refer to signal (boosted $W$ bosons from $W'$ decay) images, while the bottom plots refer to background (QCD) images. All plots have been normalized to unity in each $m$ or $p_T$ bin, such that the $z$-axis represents the percentage of images generated in a specific $m$ or $p_T$ bin that have been assigned the value of $P(\mathrm{signal})$ indicated on the $y$-axis.}
\label{fig:signal_vs_mas_pt}
\end{figure}

The auxiliary classifier is trained to achieve optimal performance in discriminating signal from background images. 
Fig. \ref{fig:signal_vs_mas_pt} confirms its ability to correctly identify the class most generated images belong to. Here, we can identify the response's dependence on the kinematic variables. Notice how $D$ is making use of its internal representation of mass to identify signal-like images: the peak of the $m$ distribution for signal events is located around 80 GeV, and indeed images with mass around that point have a higher $P(\mathrm{signal})$ than the ones at very low or very high mass. Similarly, low $p_T$ images are more likely to be classified as background, while high $p_T$ ones have a higher probability of being categorized as signal images. This behavior is well understood from a physical standpoint and can be easily cross-checked with the $m$ and $p_T$ distribution for boosted $W$ and QCD jets displayed in Fig.~\ref{fig:m_pt_tau21}. Although mass and transverse momentum influence the label assignment, $D$ is only partially relying on these quantities; there is more knowledge learned by the network that allows it, for example, to still manage to correctly classify the majority of signal and background images regardless of their $m$ and $p_T$ values.

\begin{figure}[h]
\centering
\includegraphics[width=0.3\textwidth]{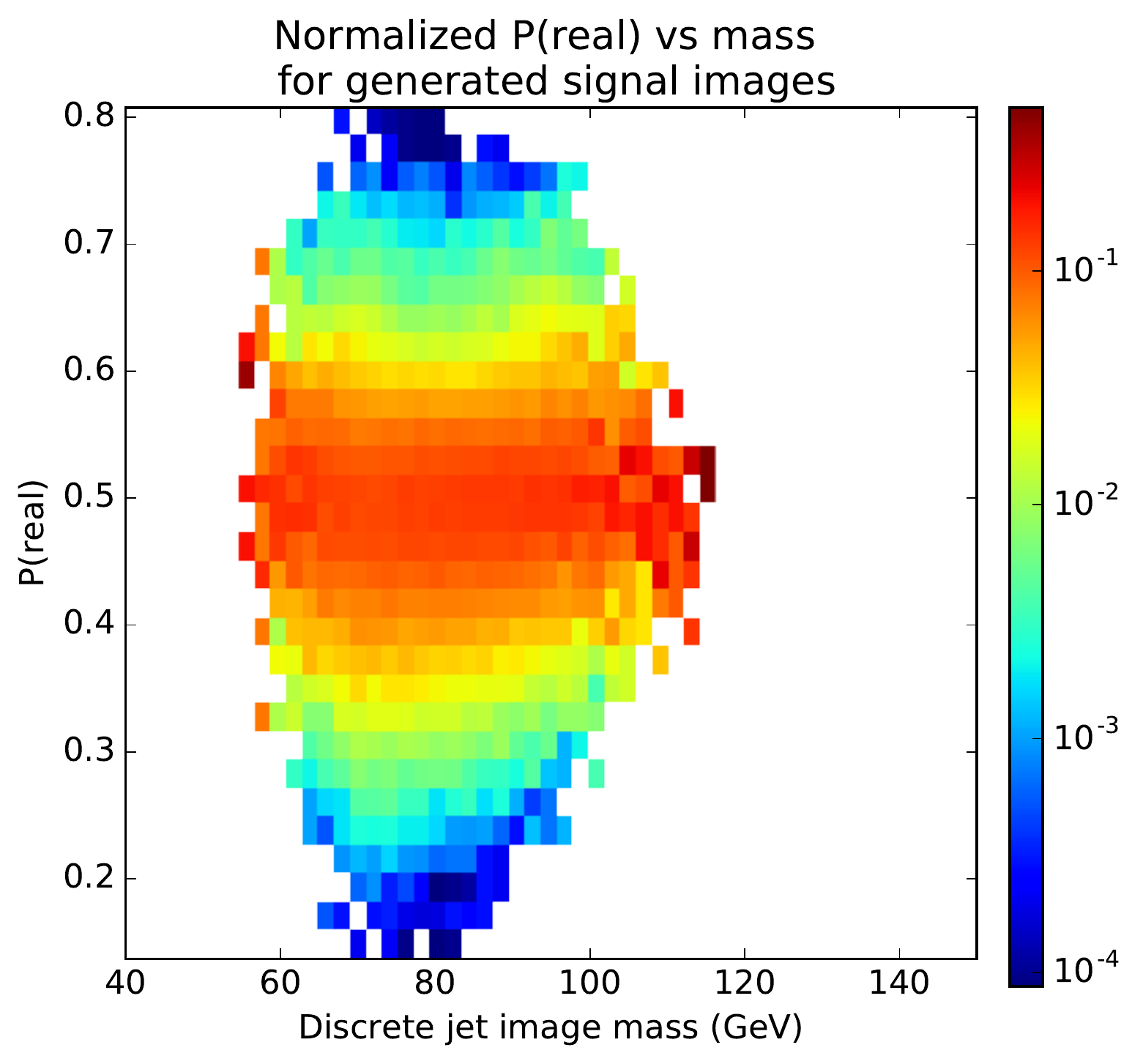}\hspace{10mm}\includegraphics[width=0.3\textwidth]{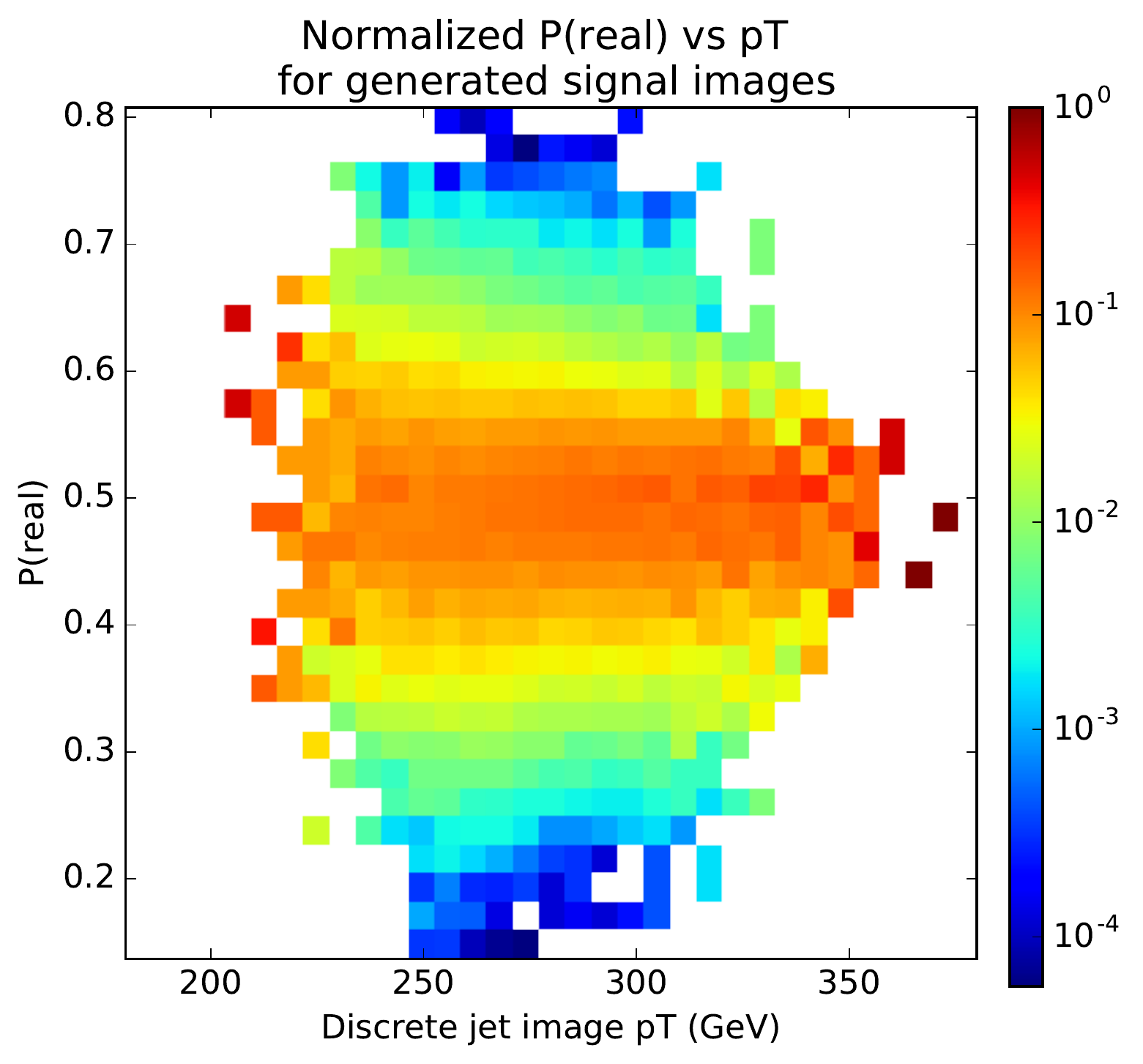} \\
\includegraphics[width=0.3\textwidth]{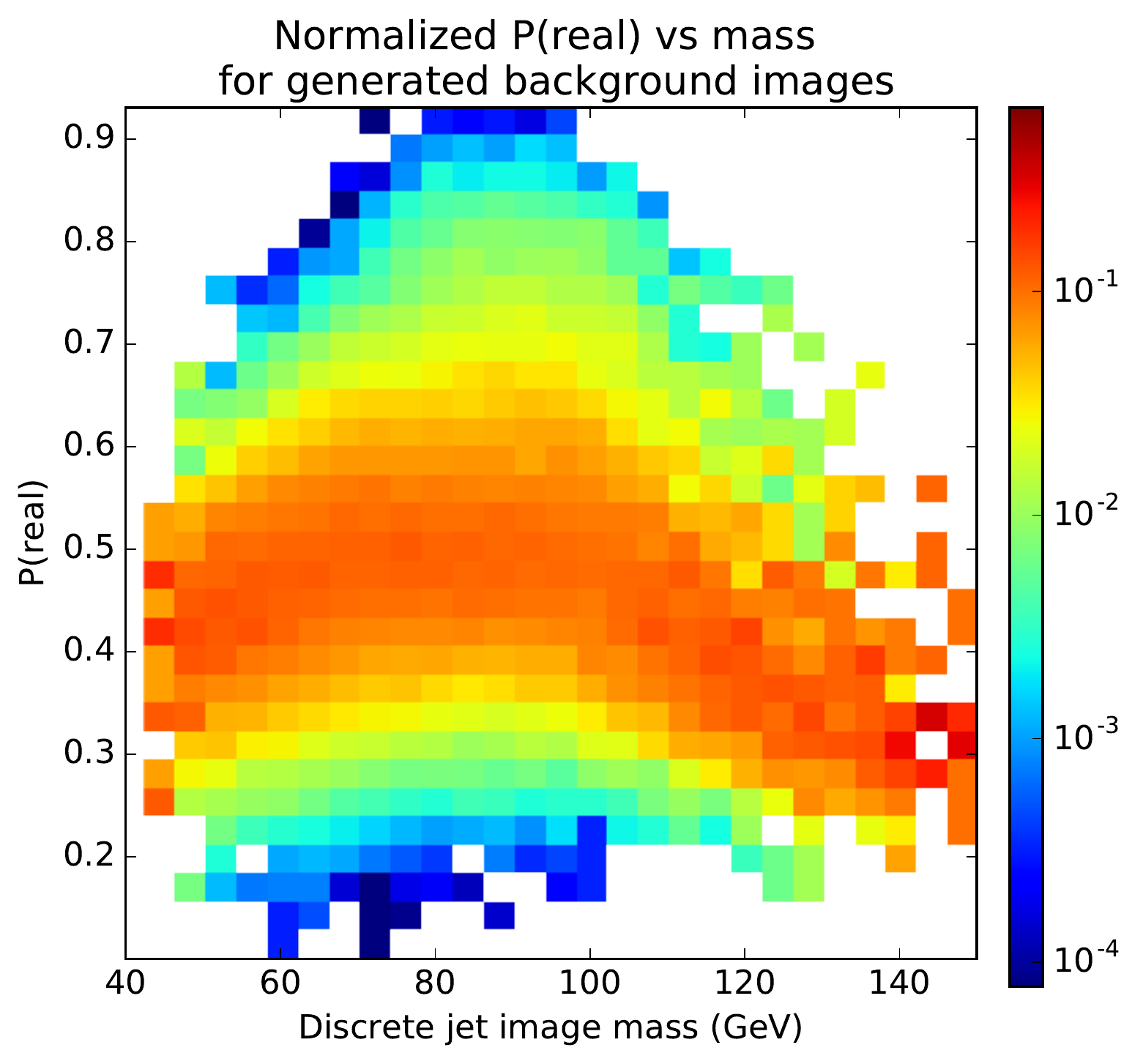}\hspace{10mm}\includegraphics[width=0.3\textwidth]{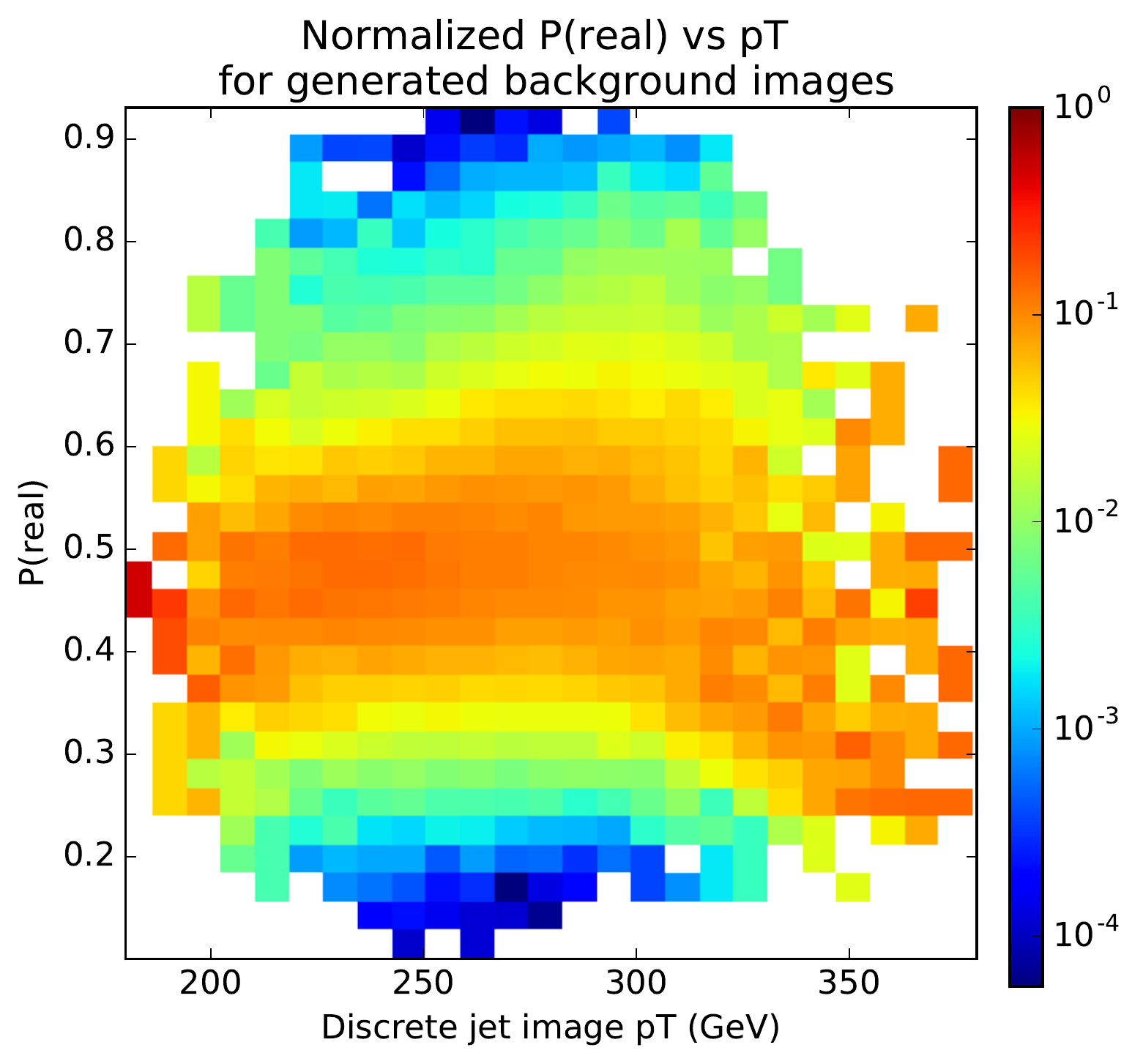} \\
\caption{Discriminator output as a function of mass, on the left, and as a function of transverse momentum, on the right. The top plots refer to signal (boosted $W$ bosons from $W'$ decay) images, while the bottom plots refer to background (QCD) images. All plots have been normalized to unity in each $m$ or $p_T$ bin, such that the $z$-axis represents the percentage of images generated in a specific $m$ or $p_T$ bin that have been assigned the value of $P(\mathrm{real})$ indicated on the $y$-axis.}
\label{fig:real_vs_mass_pt}
\end{figure}

On the other hand, Fig.~\ref{fig:real_vs_mass_pt} shows that the training has converged to a stable point such that $D$ outputs a $P(\mathrm{real}) \sim 1/2$ for almost all generated images. A high level of confusion from the discriminator is not just expected -- it's one of the goals of the adversarial training procedure. In addition, we notice no strong $m$ or $p_T$ dependence of the output, except for background images produced with $m$ and $p_T$ values outside the ranges of the training set.

Studying one-dimensional projections onto powerful physically-inspired variables is useful, but we can also try to understand the network behavior by looking directly at images. This is similar to the visualization techniques used in Ref.~\cite{deOliveira:2015xxd} except now there is not only signal and background classes, but also fake and real categories. 

\begin{figure}[h!]
\centering
\includegraphics[width=0.8\textwidth]{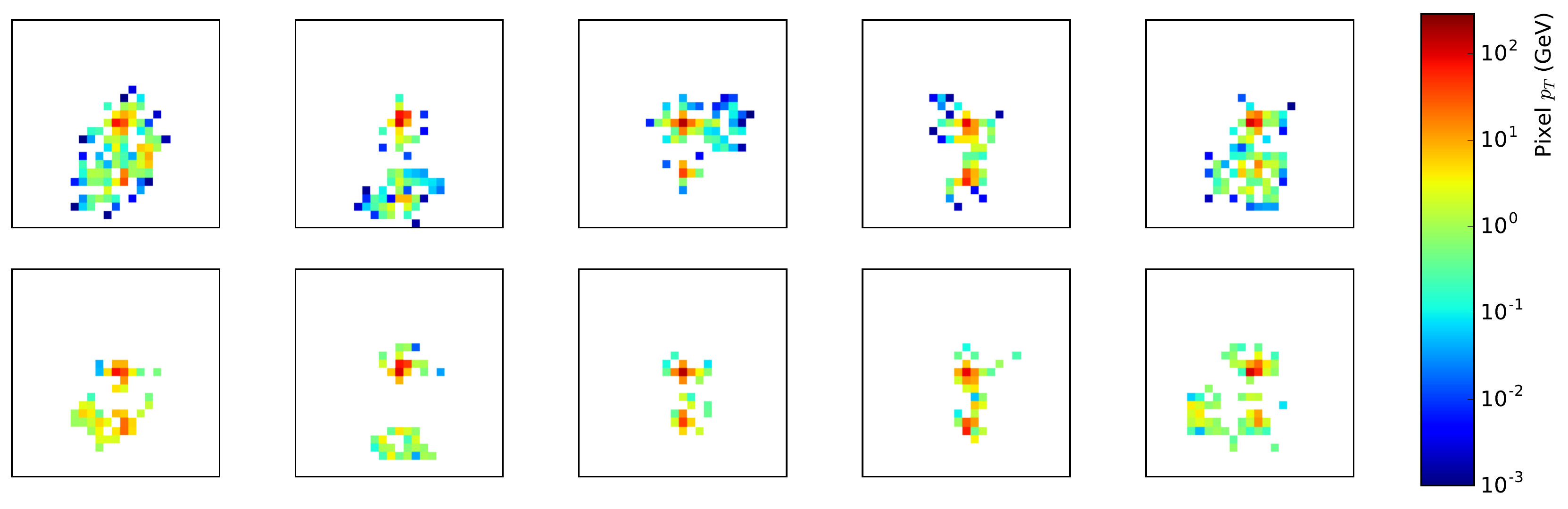}
\caption{Randomly selected Pythia images (top row) and their nearest generated neighbor (bottom row).}
\label{fig:nearest_neighbors}
\end{figure}

As is commonly done in GAN literature, we examine randomly selected images from the data distribution and show a comparison with their nearest generated neighbor to show that we have not memorized the training set (Figure~\ref{fig:nearest_neighbors}). The dispersion patterns appear realistic, and, although our generative model seems inadequate in the very low intensity region, this turns out to be inconsequential because most of the important jet information is carried by the higher $p_\text{T}$ particles.

\begin{figure}[h!]
\centering
\includegraphics[width=0.333\textwidth]{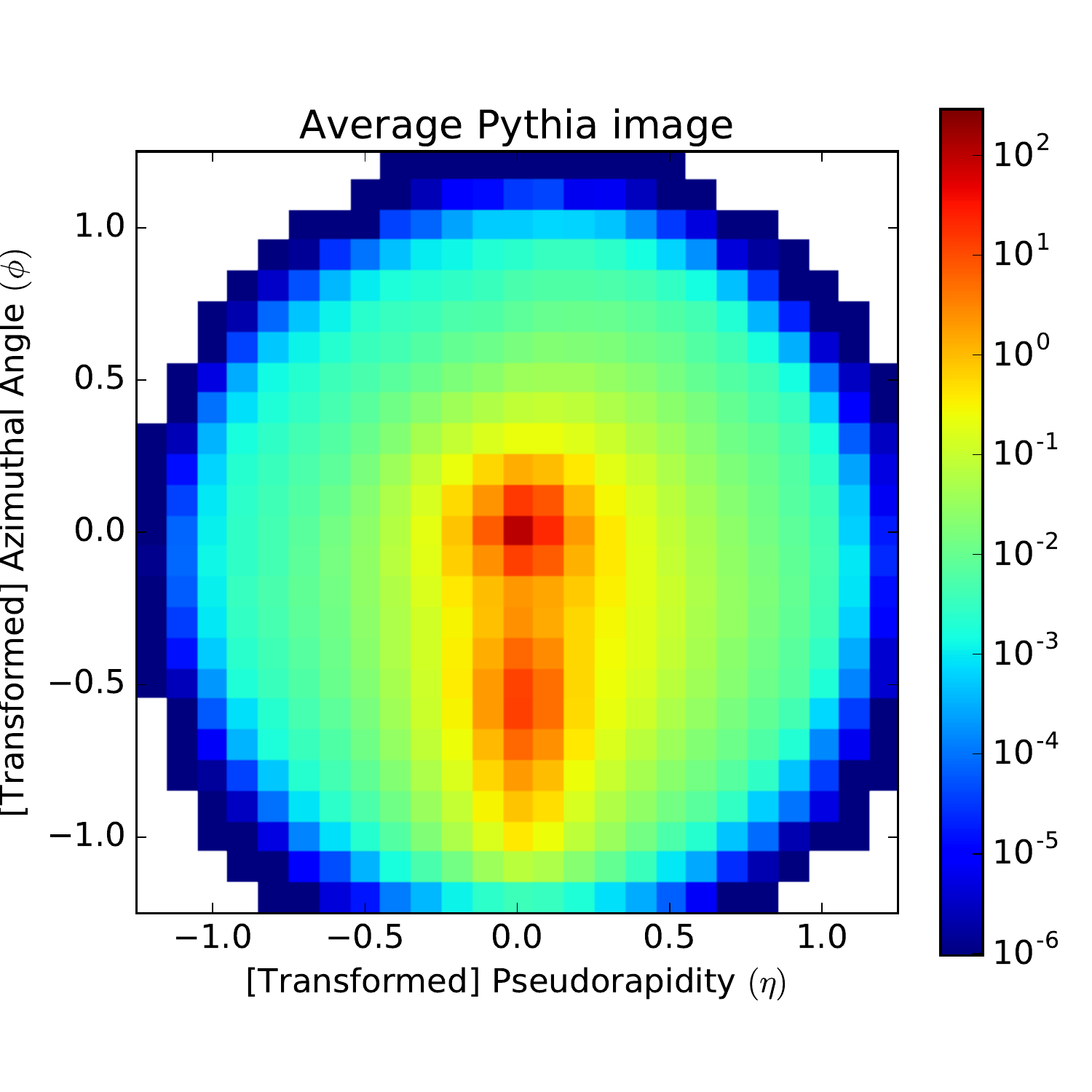}\includegraphics[width=0.333\textwidth]{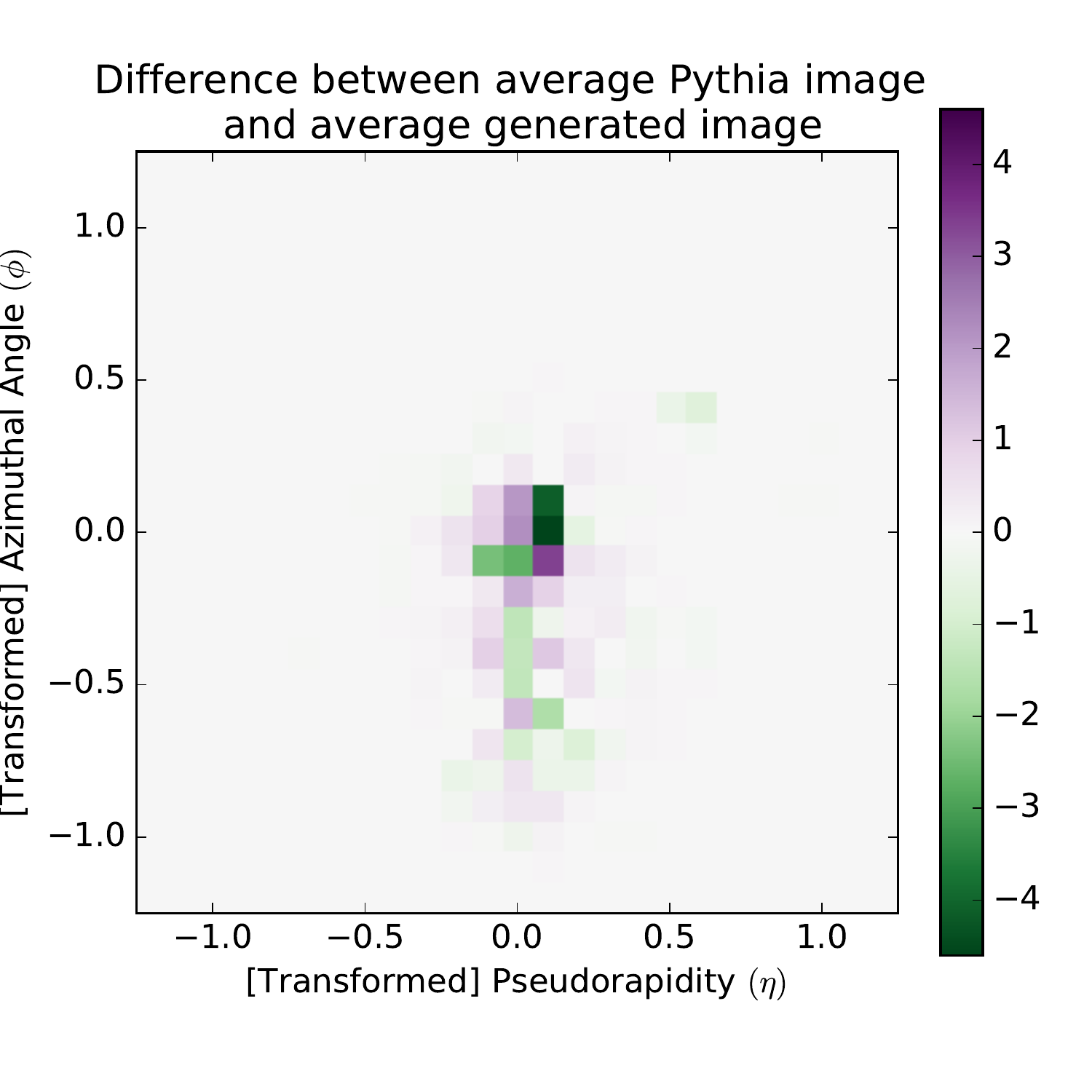}\includegraphics[width=0.333\textwidth]{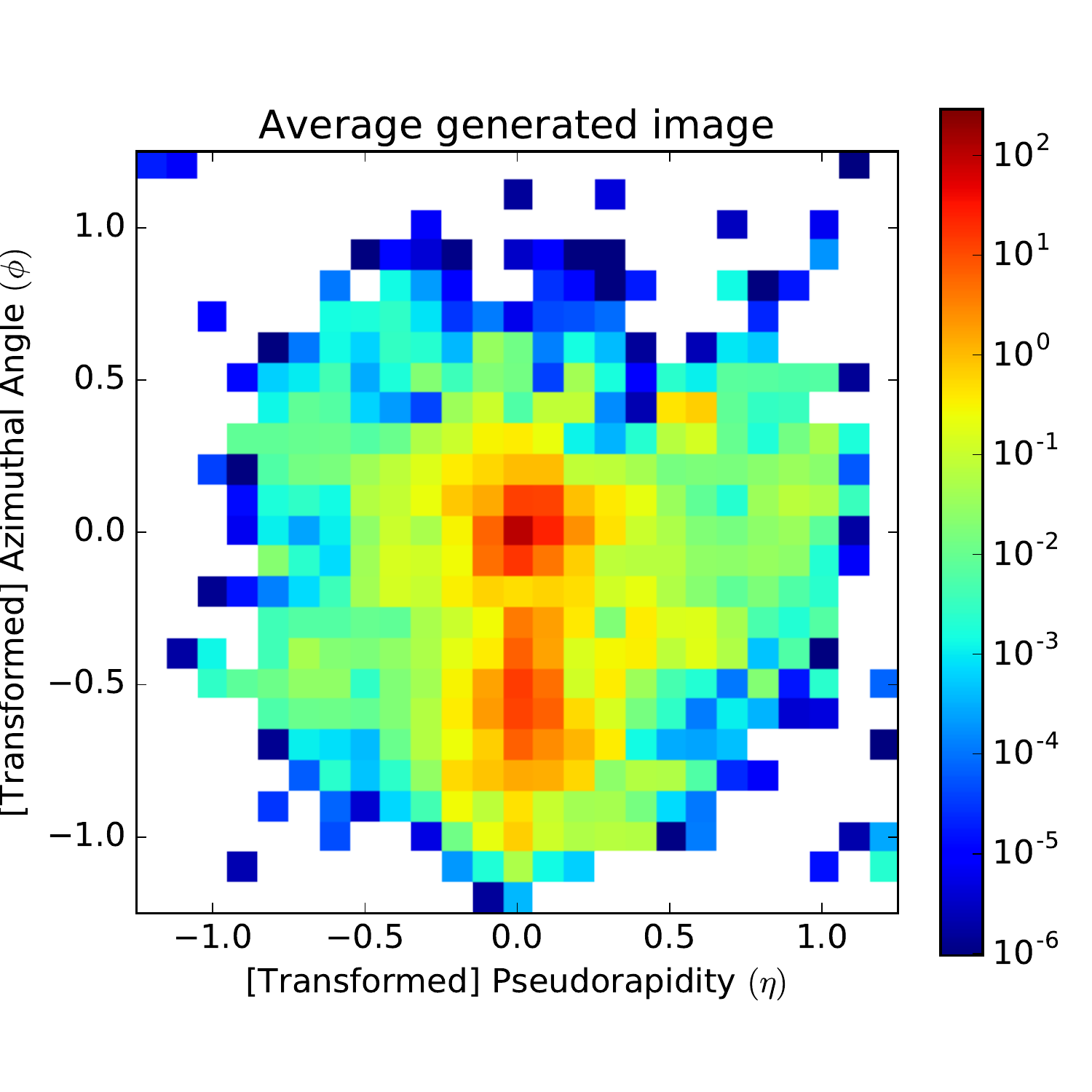}
\caption{Average image produced by Pythia (left) and by the GAN (right), displayed on log scale, with the difference between the two (center), displayed on linear scale.}
\label{fig:pythia-gan}
\end{figure}

The topology of jet images becomes more visible when examining averages over jet images with specific characteristics. Figure~\ref{fig:pythia-gan} shows the average Pythia image, the average generated image, and their difference\footnote{Similar plots for the average signal and background images are shown in Fig.~\ref{fig:signal_pythia-gan}, \ref{fig:bkg_pythia-gan}.}. On average, the generative model produces large, spread out, roughly circular dispersion patterns that approximate those of Pythia jet images. The generated images span multiple orders of magnitude, and differ from Pythia images in the pixel intensity of the central regions by only a few percent.

\begin{figure}[h!]
\centering
\includegraphics[width=0.333\textwidth]{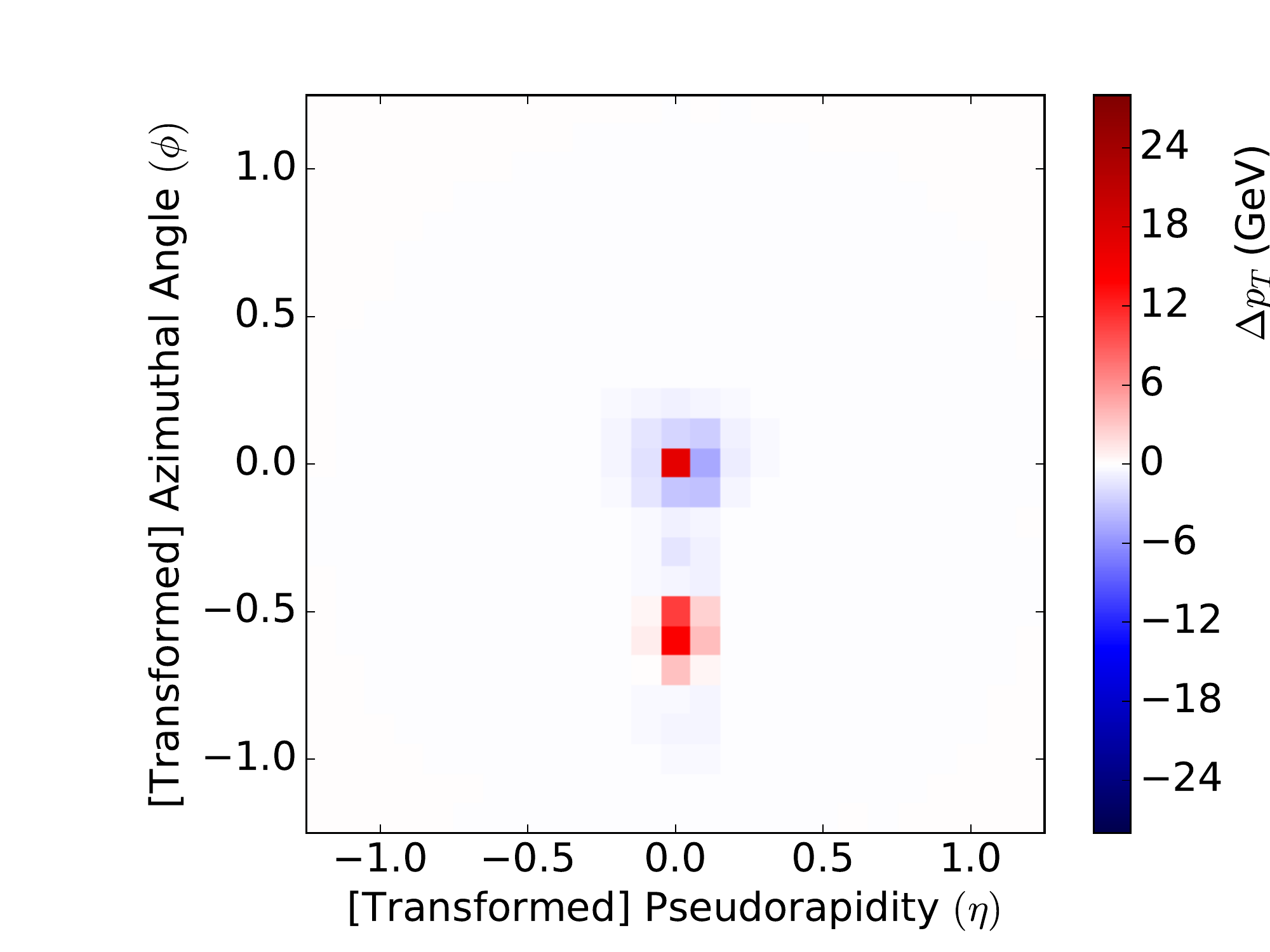}\hspace{5mm}\includegraphics[width=0.333\textwidth]{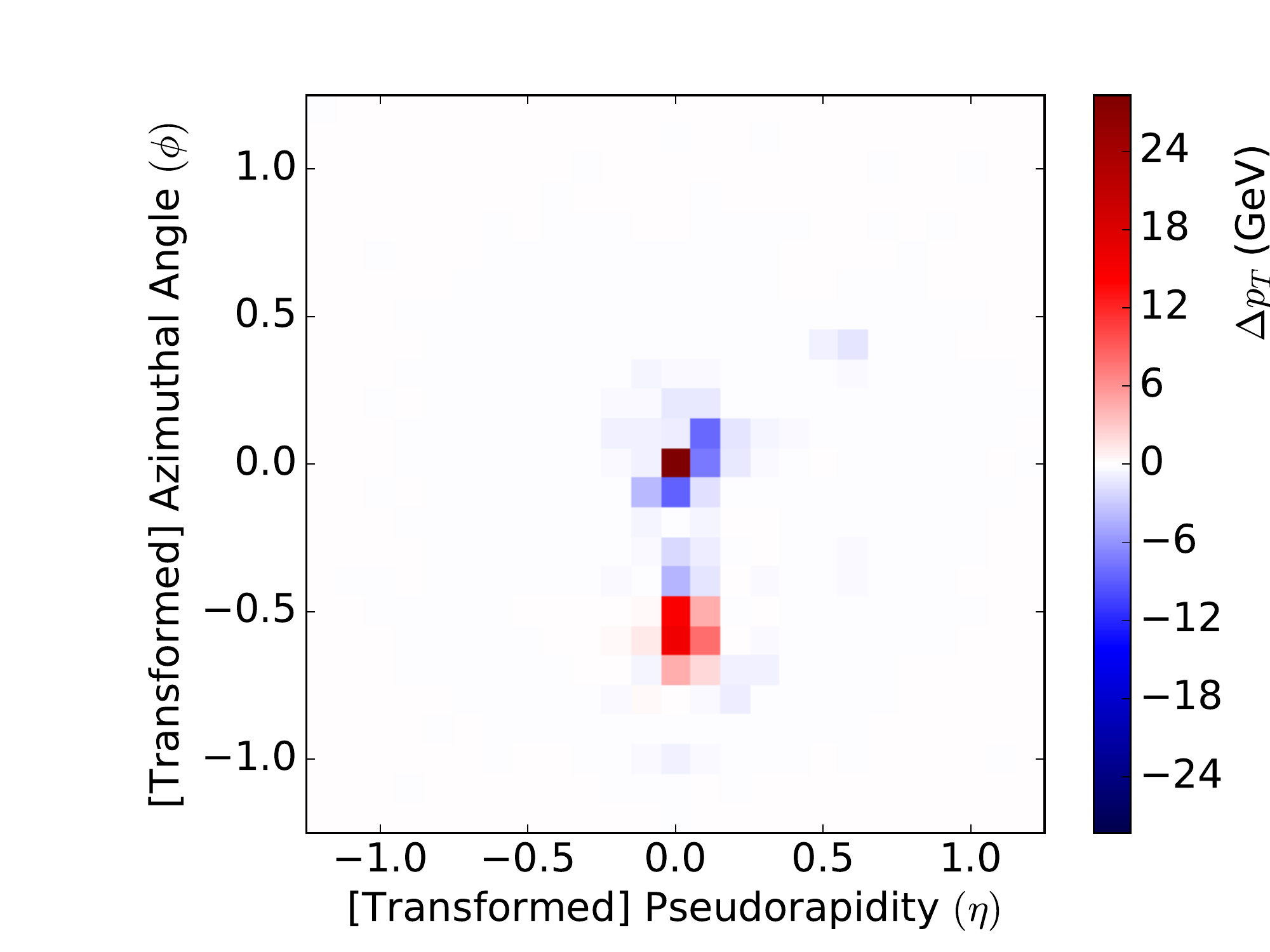}
\caption{The difference between signal and background images generated by Pythia (left) and GAN (right). Similar images are available in Appendix~\ref{app:auxmaterial} (Fig.~\ref{fig:real_signal-bkg},~\ref{fig:fake_signal-bkg}) after conditioning on the discriminator's output, which classifies images into real and fake.}
\label{fig:signal-bkg}
\end{figure}

For classification, the most important property of the GAN images is that the difference between signal and background should be faithfully reproduced.  This is shown qualitatively on average in Fig.~\ref{fig:signal-bkg}. The polarity of the difference between signal and background for both Pythia and GAN-generated images is consistent at the individual pixel level. However, the magnitude difference is stronger in GAN-generated images, leading us to believe that the GAN overestimates the importance of individual pixel contributions to the categorization of signal and background.

\begin{figure}[h!]
\centering
\includegraphics[width=0.333\textwidth]{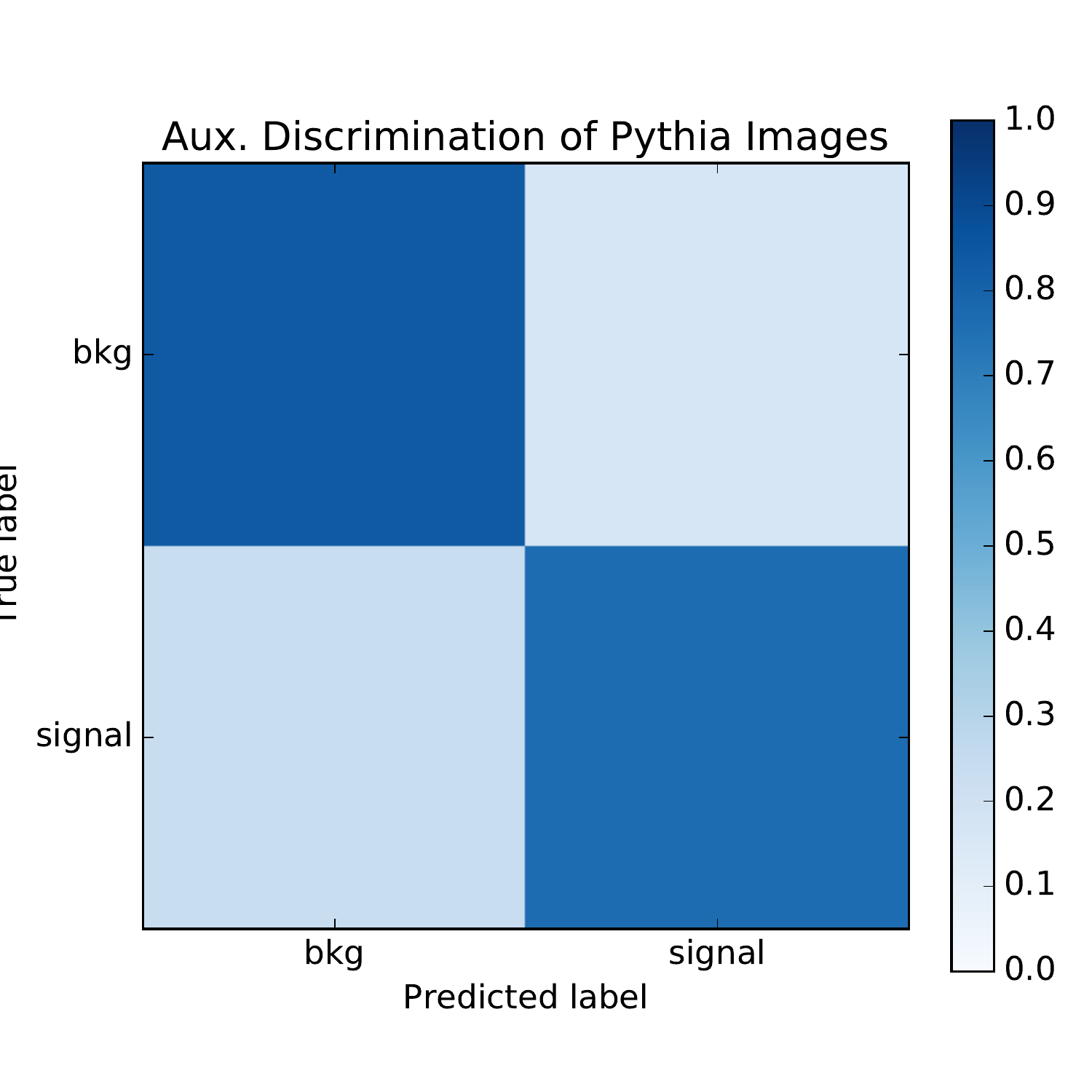}\hspace{5mm}\includegraphics[width=0.333\textwidth]{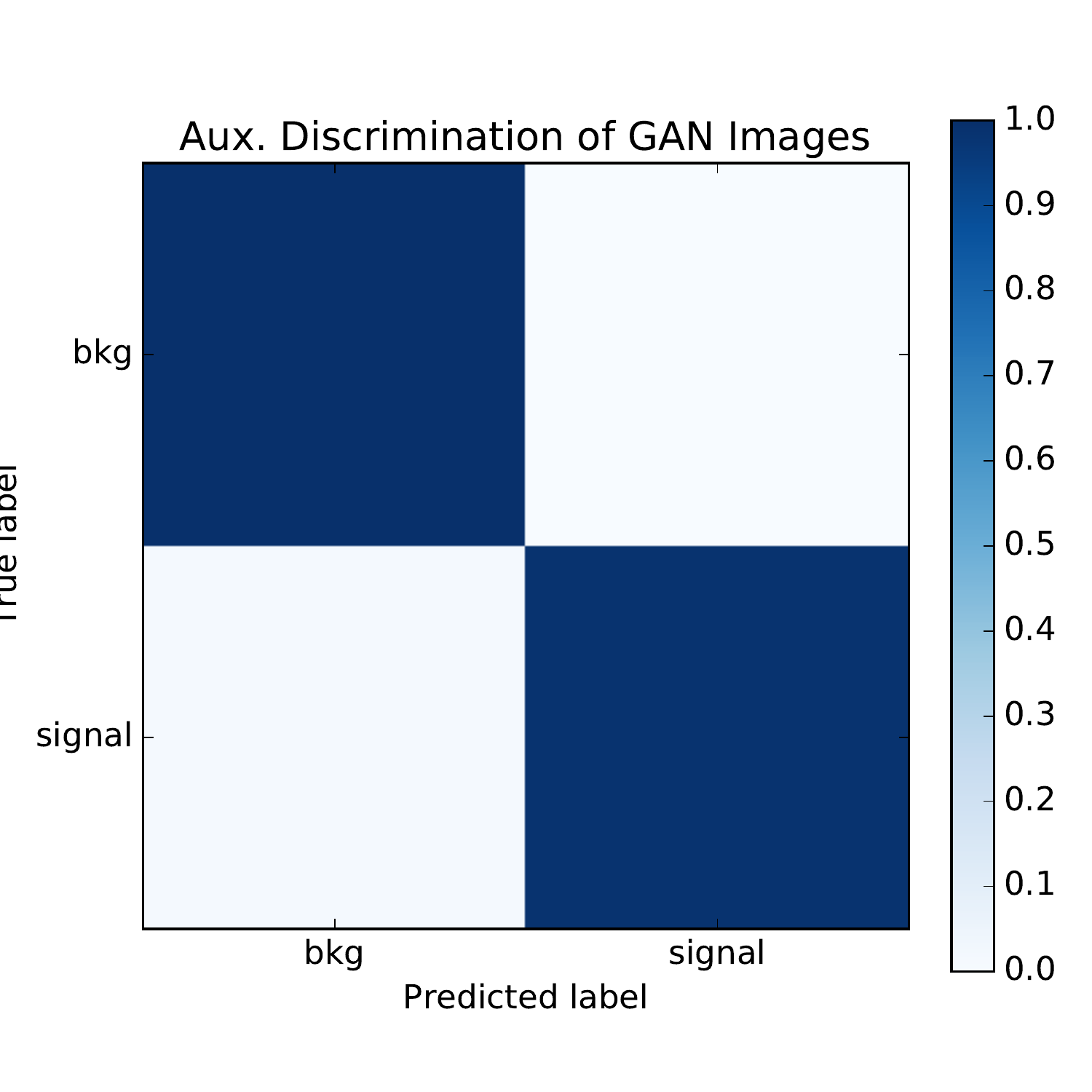}
\caption{We plot the normalized confusion matrices showing the percentage of signal and background images that the auxiliary classifier successfully labels. The matrices are plotted for Pythia images (left) and GAN images (right).}
\label{fig:confusion_bkg_signal}
\end{figure}

Another shortcoming of GANs when applied to the generation of images that, like ours, are divided into inherently overlapping and indistinguishable classes, is their inability to carefully explore the grey area represented by the subspace of images that are hard to classify with the correct label.
There exists an entire literature \cite{Cogan:2014oua, deOliveira:2015xxd, Baldi:2016fql} on the effort of classifying $W$ boson from QCD jet images using machine learning, and the topic itself is an active field of research in high energy physics. 
In particular, inserting this auxiliary classification task into the GAN's loss function may induce the generator to produce very distinguishably $W$-like or QCD-like jet images. The production of ambiguous images is unfavorable under the ACGAN loss formulation. Evidence of the occurrence of this phenomenon can be found by analyzing the normalized confusion matrices (Fig.~\ref{fig:confusion_bkg_signal}) generated from the auxiliary classifier's output evaluated on Pythia and GAN-generated images. The plots show that classification appears to be easier for GAN-generated images, which $D$ correctly labels with higher success rate that Pythia images.  

\begin{figure}[h!]
\centering
\includegraphics[width=0.333\textwidth]{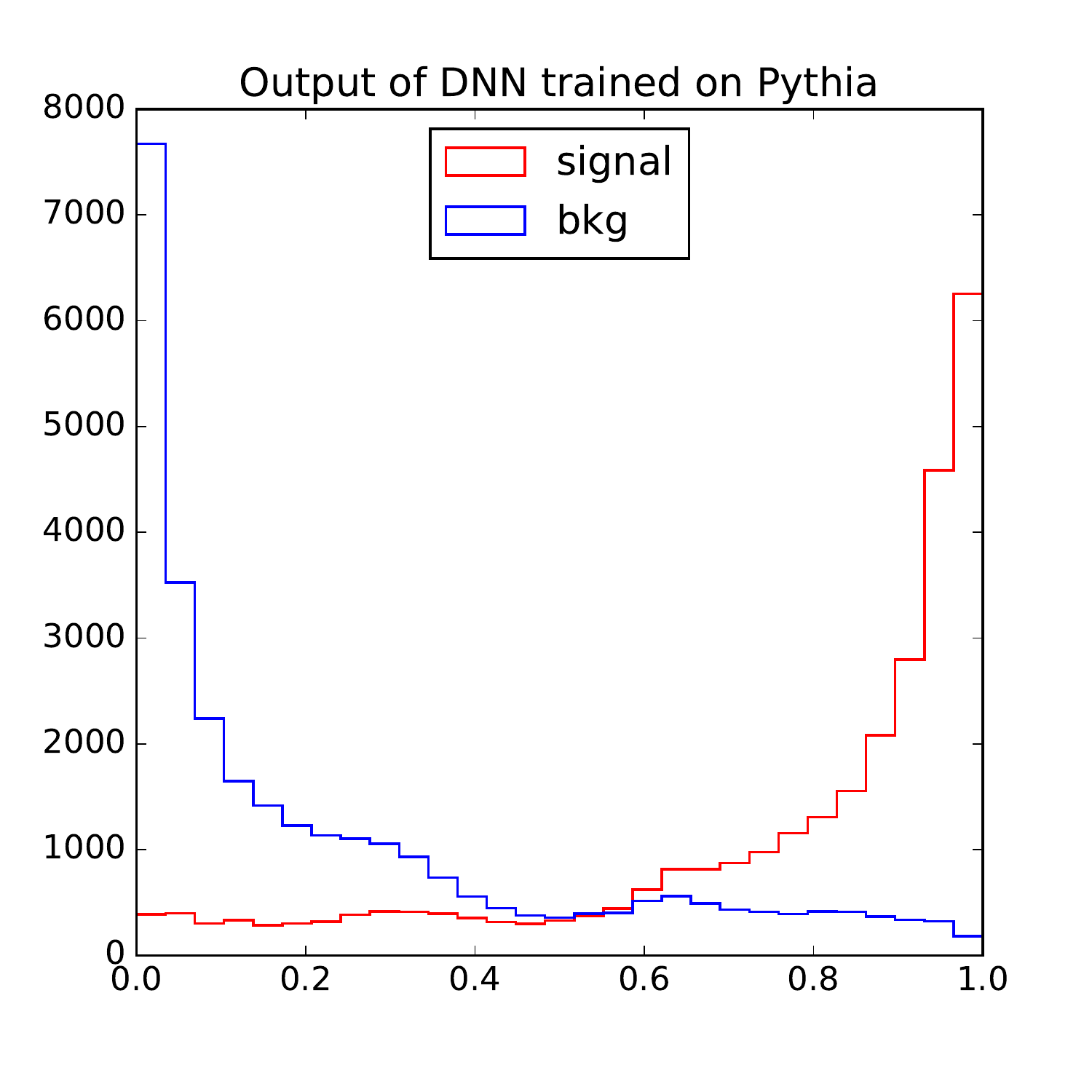}\hspace{5mm}\includegraphics[width=0.333\textwidth]{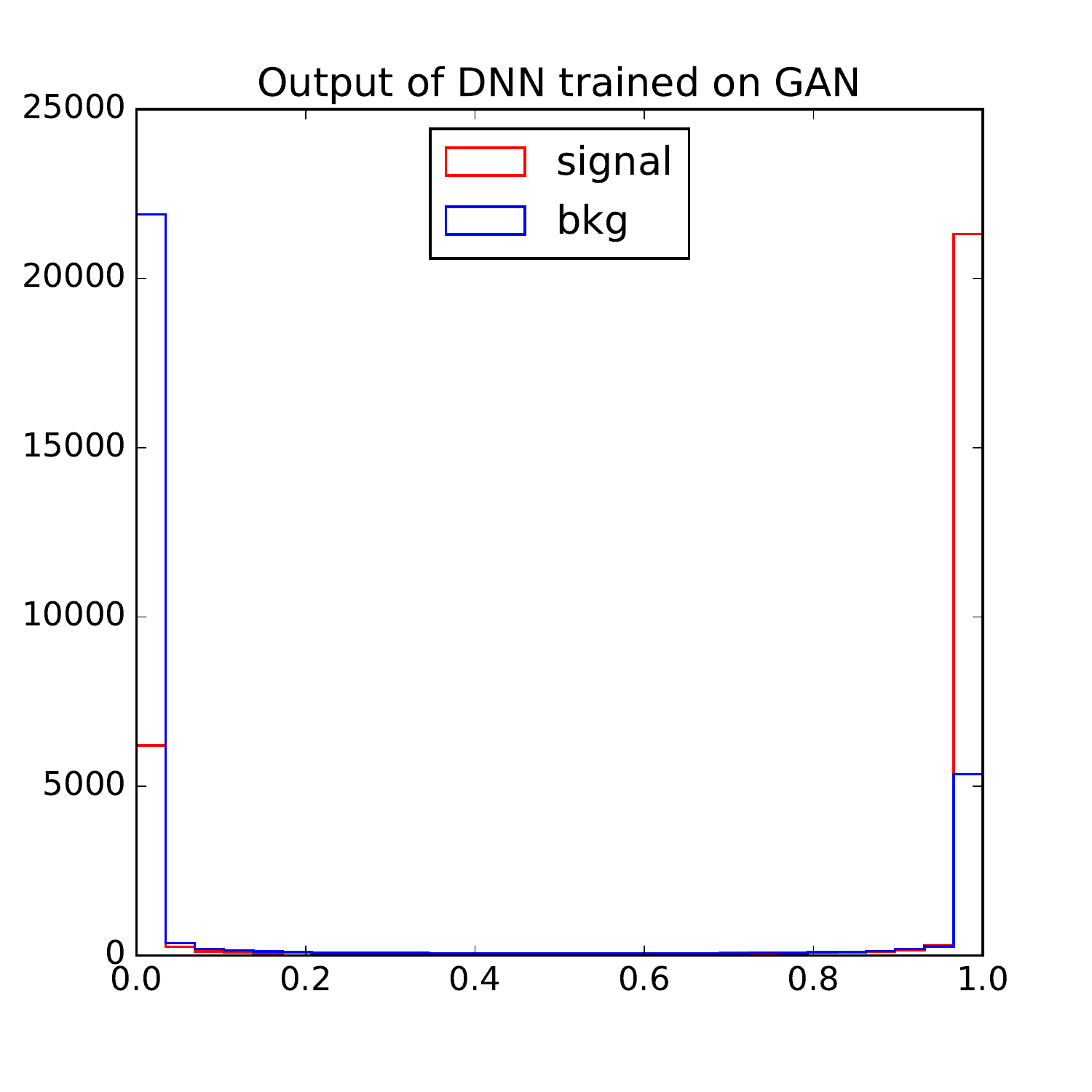}
\caption{Output of the 2 fully connected nets - one trained on Pythia, one trained on GAN images - evaluated on Pythia images to discriminate boosted $W$ bosons (signal) from QCD (bkg).}
\label{fig:classifier_output}
\end{figure}

A consequence of the inability to produce equivocal images is that GAN-generated images currently do not represent a viable, exclusive substitute to Pythia images as a training set for a $W$ vs QCD classifier that is to be applied for the discrimination of Pythia images. We check this hypothesis by training a fully-connected `MaxOut' network in the same vogue as the one in \cite{deOliveira:2015xxd}. Two trainings are performed: one on a subset of Pythia images, one on GAN images. Both are evaluated on a test set composed exclusively of Pythia images. Fig.~\ref{fig:classifier_output} clearly shows how, unlike Pythia images, GAN images cause the network to easily create two distinct representations for signal and background, which in turn leads to a higher misclassification rate when the model is applied to real images. We expect that with more research in this direction, coupled with better theoretical understanding, this problem with be ameliorated. Nonetheless, generated images can still be useful for data augmentation.

\begin{figure}[h]
    \centering
    \includegraphics[width=0.333\textwidth]{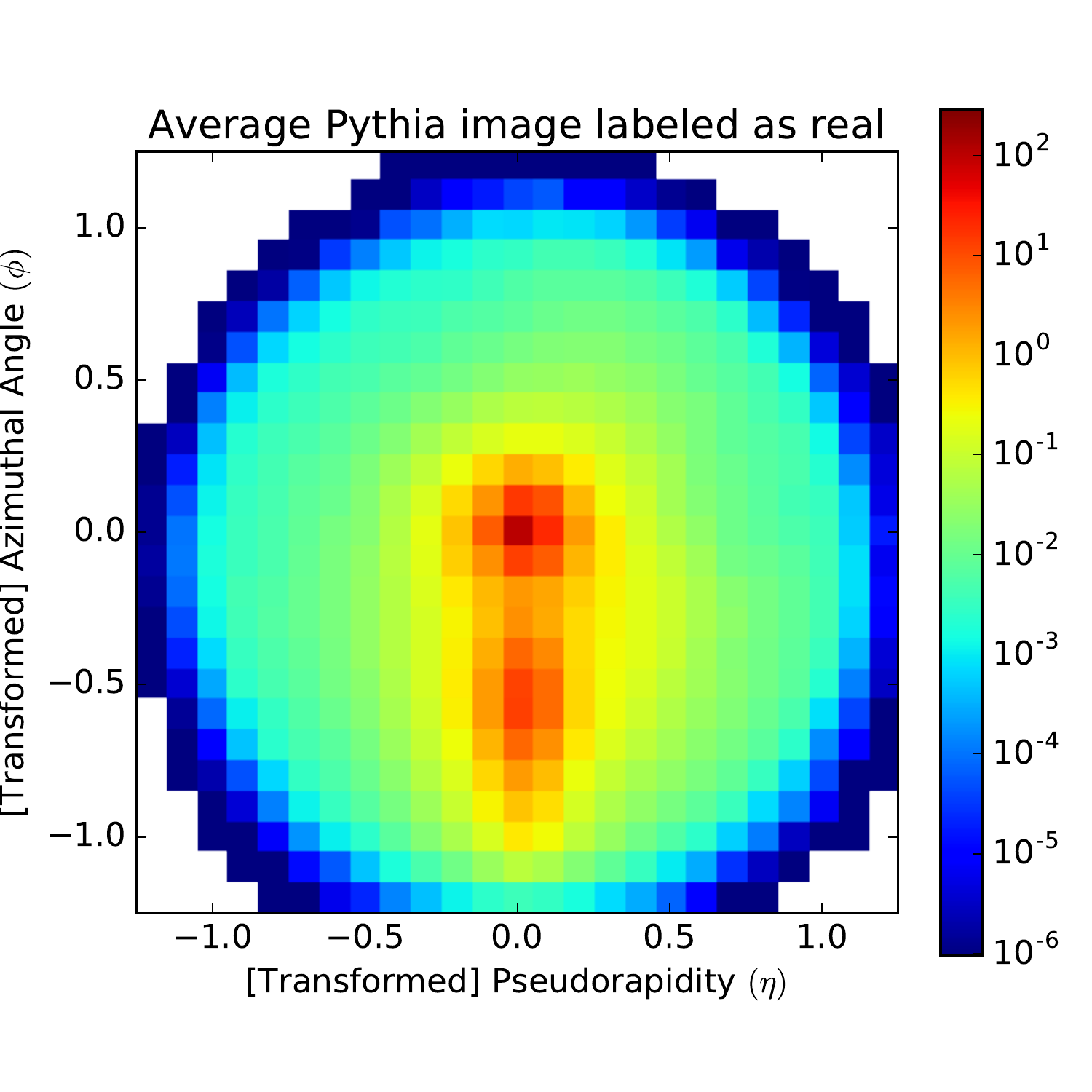}\includegraphics[width=0.333\textwidth]{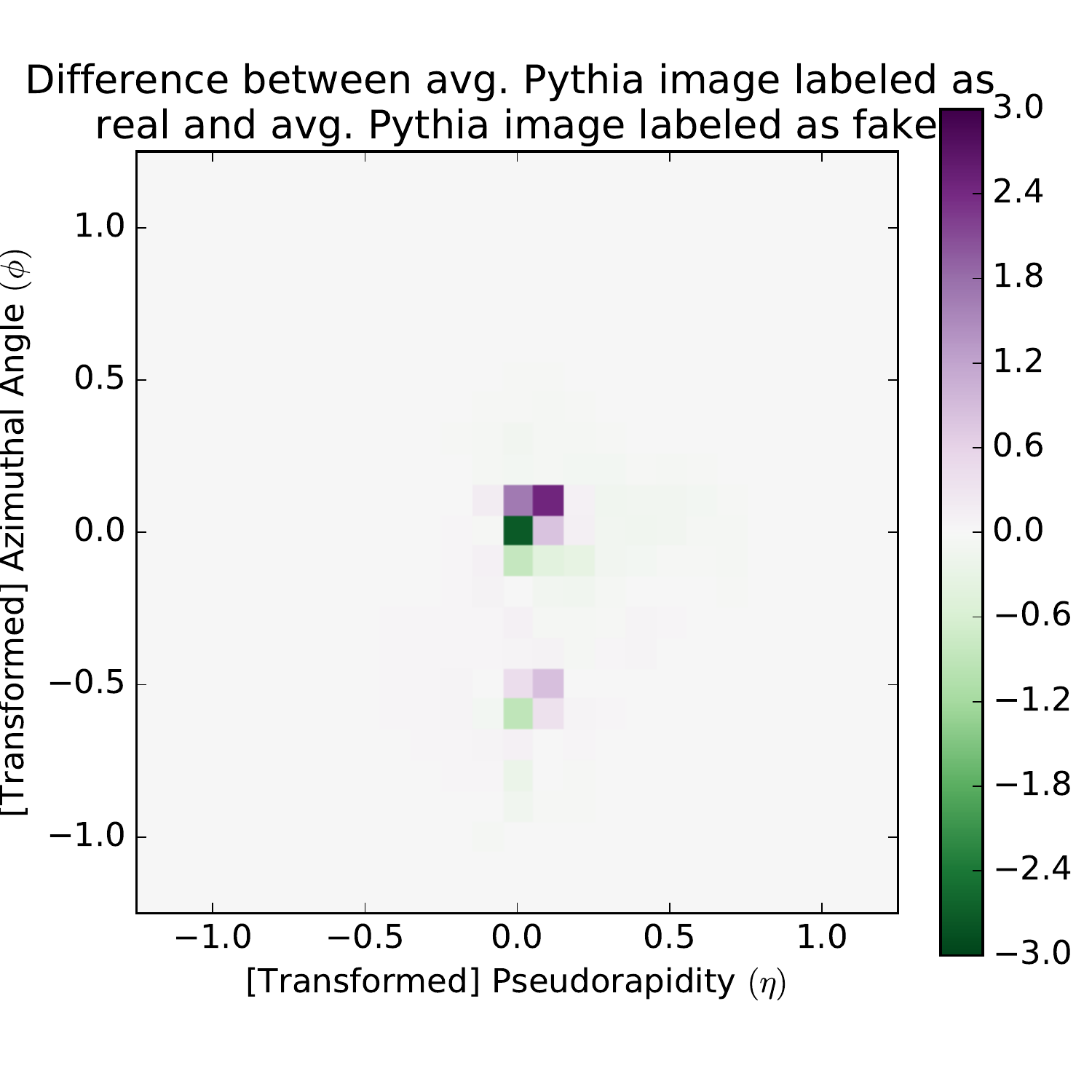}\includegraphics[width=0.333\textwidth]{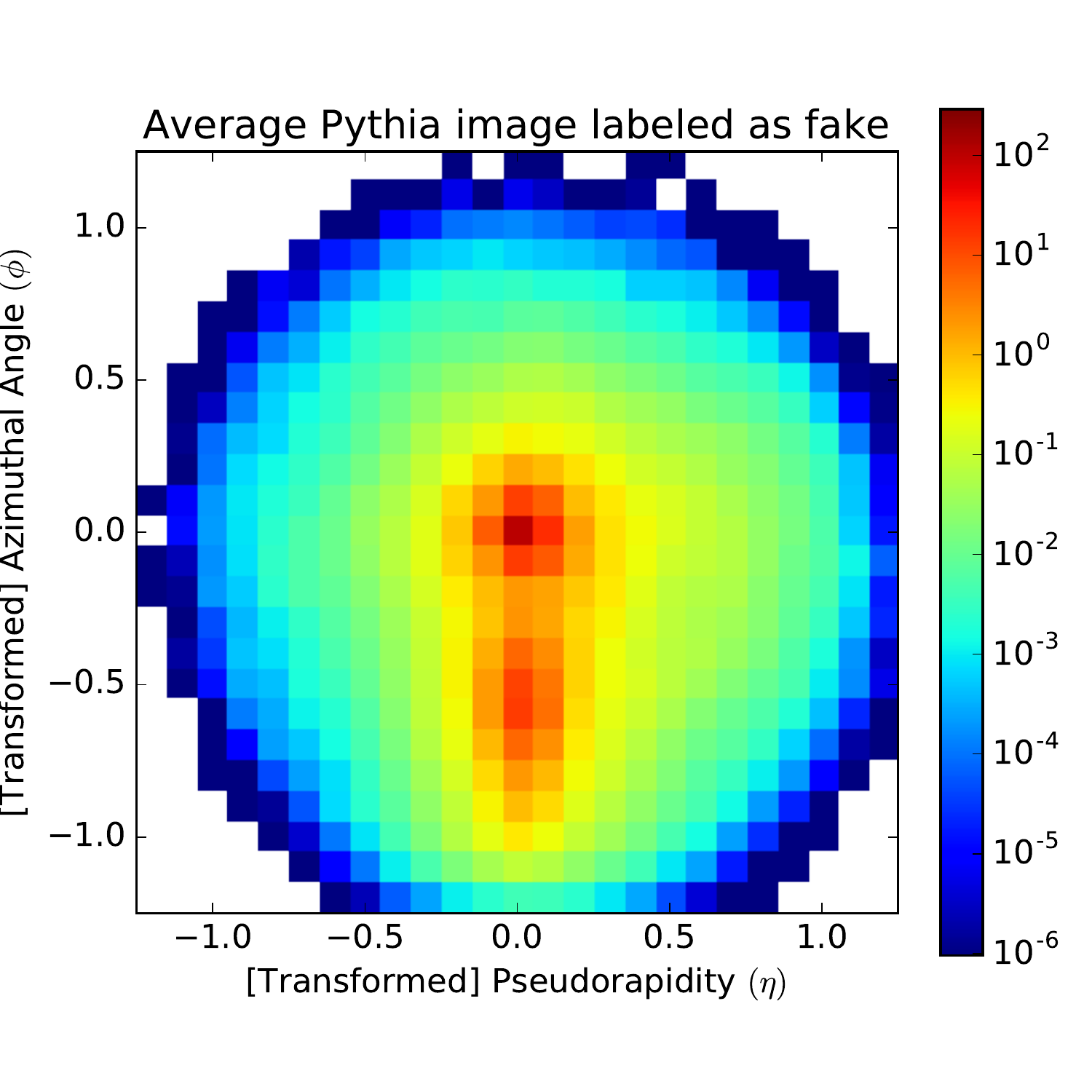}
    \caption{
    The average Pythia image labeled as real (left), as fake (right), and the difference between these two (middle) plotted on linear scale. All plots refer to aggregated signal and background jets; similar plots for individual signal and background only jets are shown in Fig.~\ref{fig:pythia_signal_real-fake}, \ref{fig:pythia_bkg_real-fake}.
    }
    \label{fig:pythia_real-fake}

    \vspace*{\floatsep}
    \includegraphics[width=0.333\textwidth]{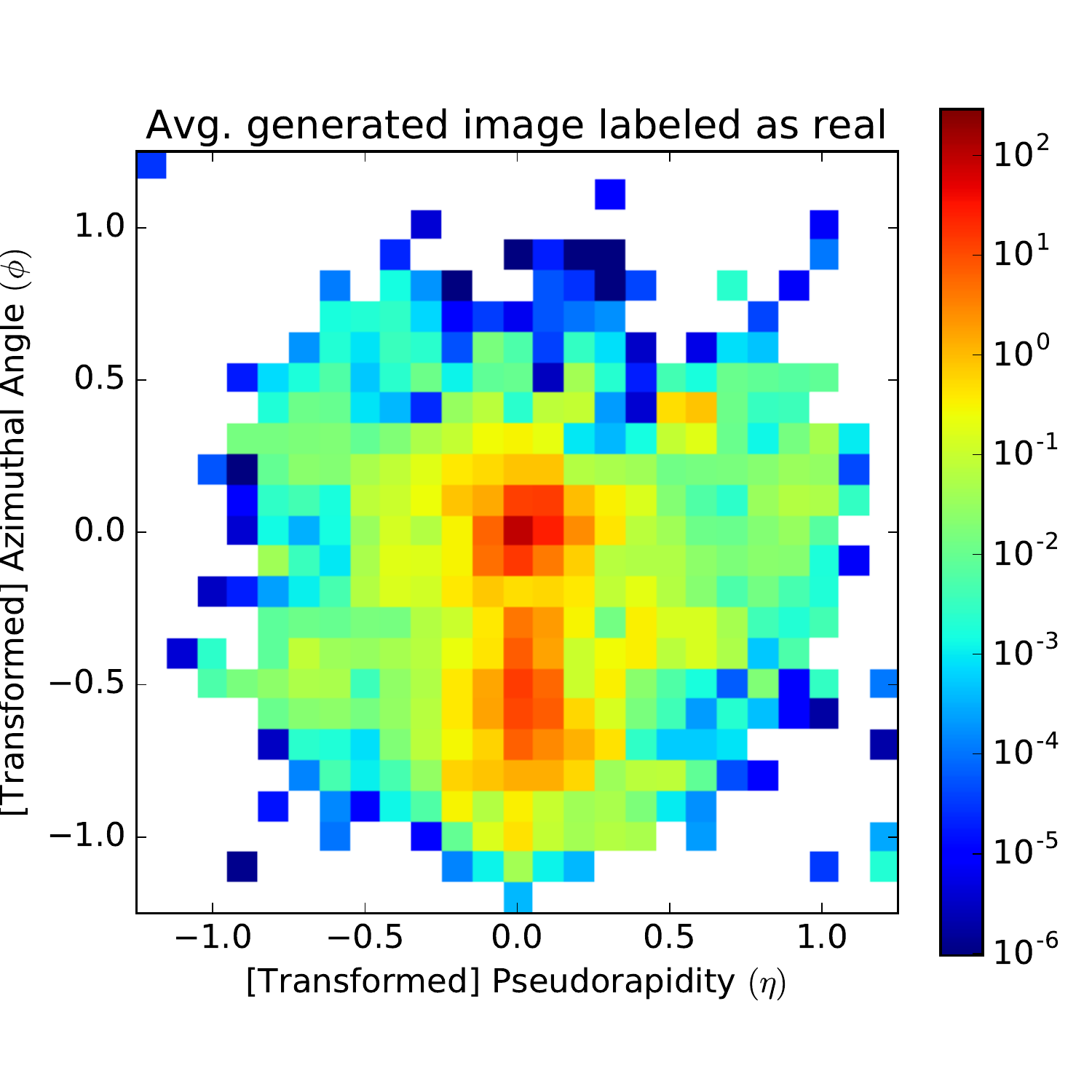}\includegraphics[width=0.333\textwidth]{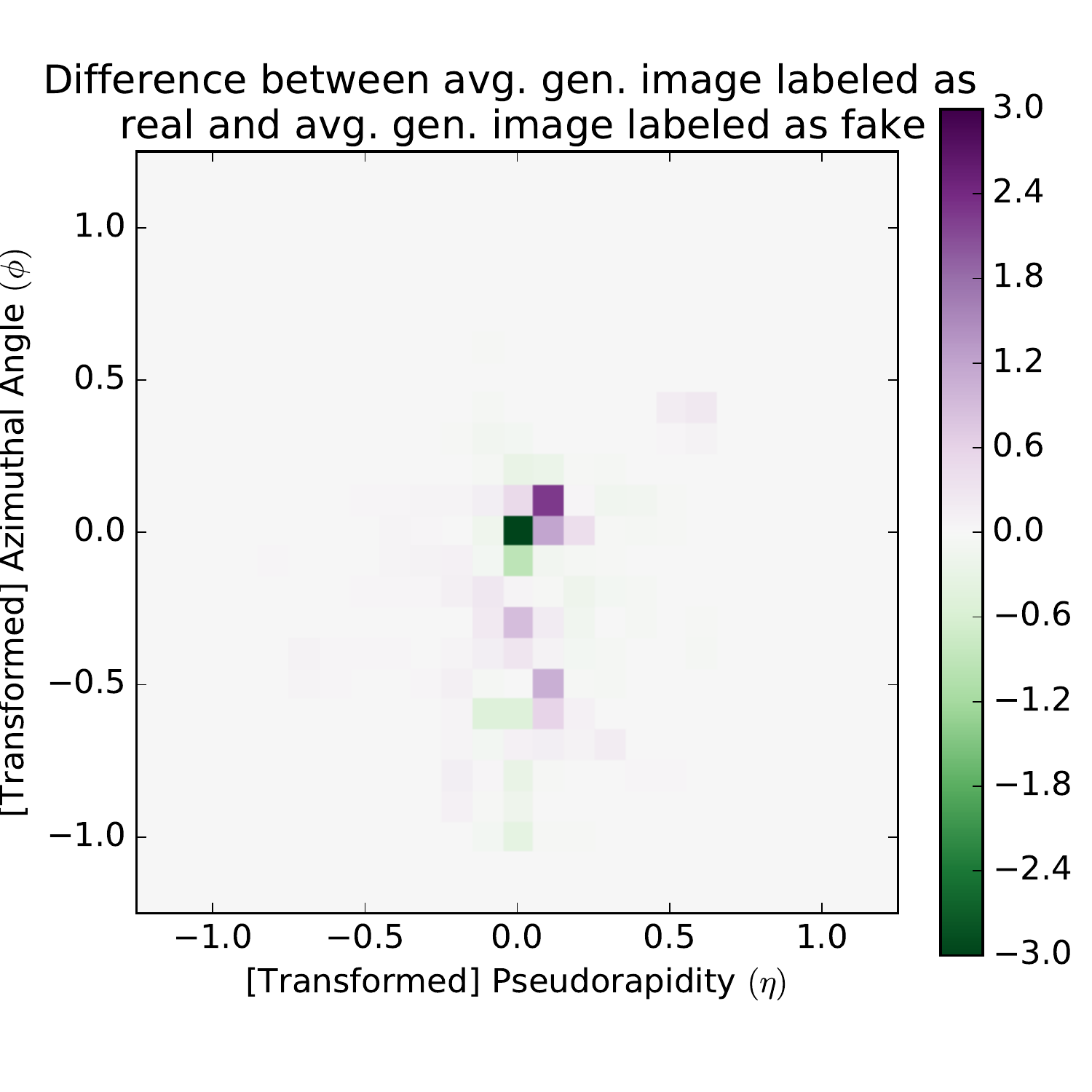}\includegraphics[width=0.333\textwidth]{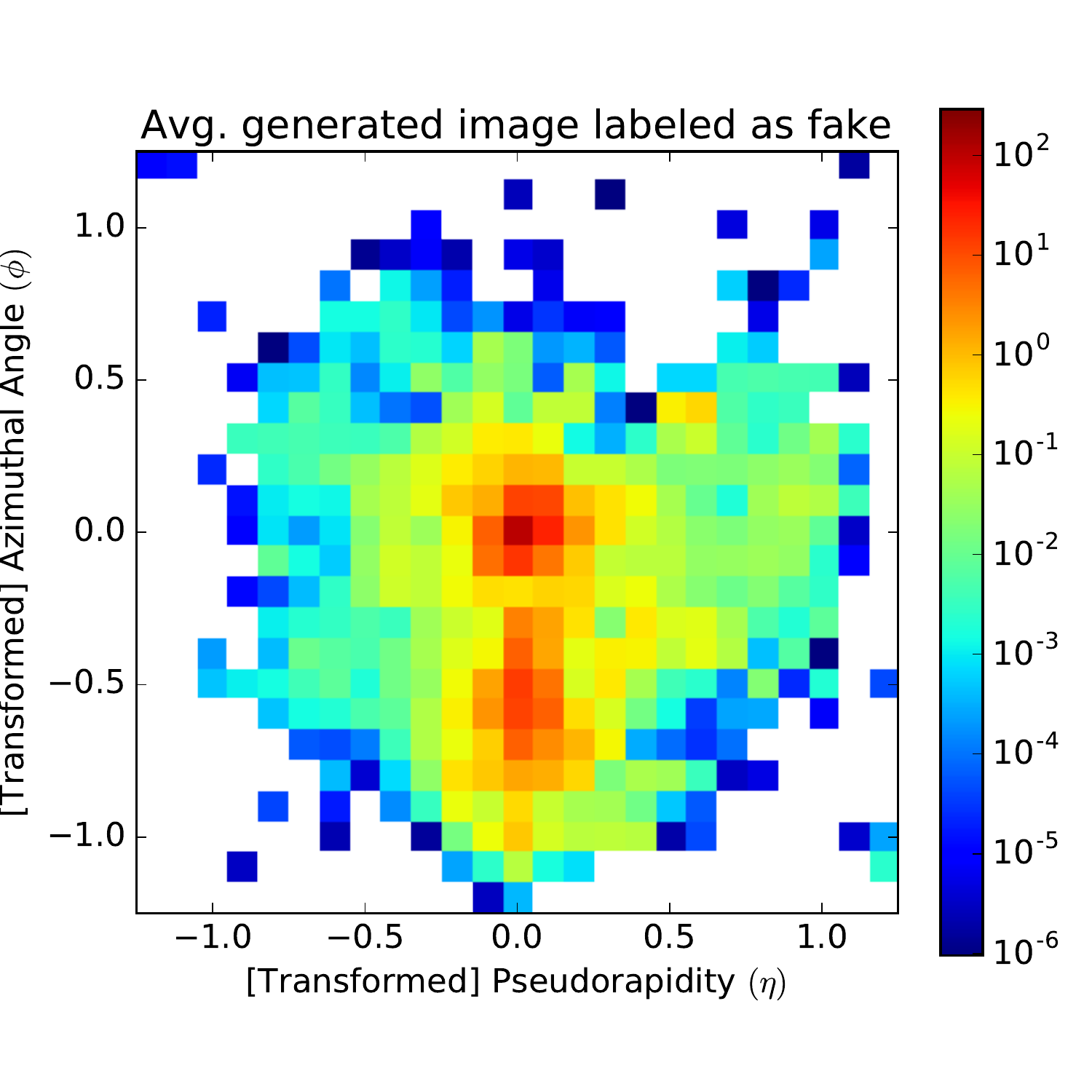}
    \caption{The average GAN-generated image labeled as real (left), as fake (right), and the difference between these two (middle) plotted on linear scale. All plots refer to aggregated signal and background jets; similar plots for individual signal and background only jets are shown in Fig.~\ref{fig:gan_signal_real-fake}, \ref{fig:gan_bkg_real-fake}.
    }
    \label{fig:gan_real-fake}
\end{figure}

Finally, we provide further visual analysis of Pythia and GAN images, this time aimed at identifying what features the discriminator network $D$ is using to differentiate real and fake images. Fig.~\ref{fig:pythia_real-fake} shows the average radiation pattern in truly real and falsely real images, and the difference between the two. Conversely, Fig.~\ref{fig:gan_real-fake} shows the average pattern in falsely fake and truly fake images, and their difference. In both cases, the most striking feature is that a higher intensity in the central pixel is associated with a higher probability of the image being classified as fake, while real-looking images tend to have a more diffuse radiation around the leading subjet, especially concentrated in the top right area adjacent to it.

\subsection{Most Representative Images}
\label{ssec:500}
All plots so far were produced using 200k Pythia images and 200k GAN-generated images, averaged over all examples with specific characteristics. Now, to better understand what's being learned by the generative model, we only consider the 500 most representative jet images of a particular kind, and average over them. This helps identifying and avoids washing out striking and unique features.

\afterpage{
\begin{figure}[h!]
\centering
\includegraphics[width=0.3\textwidth]{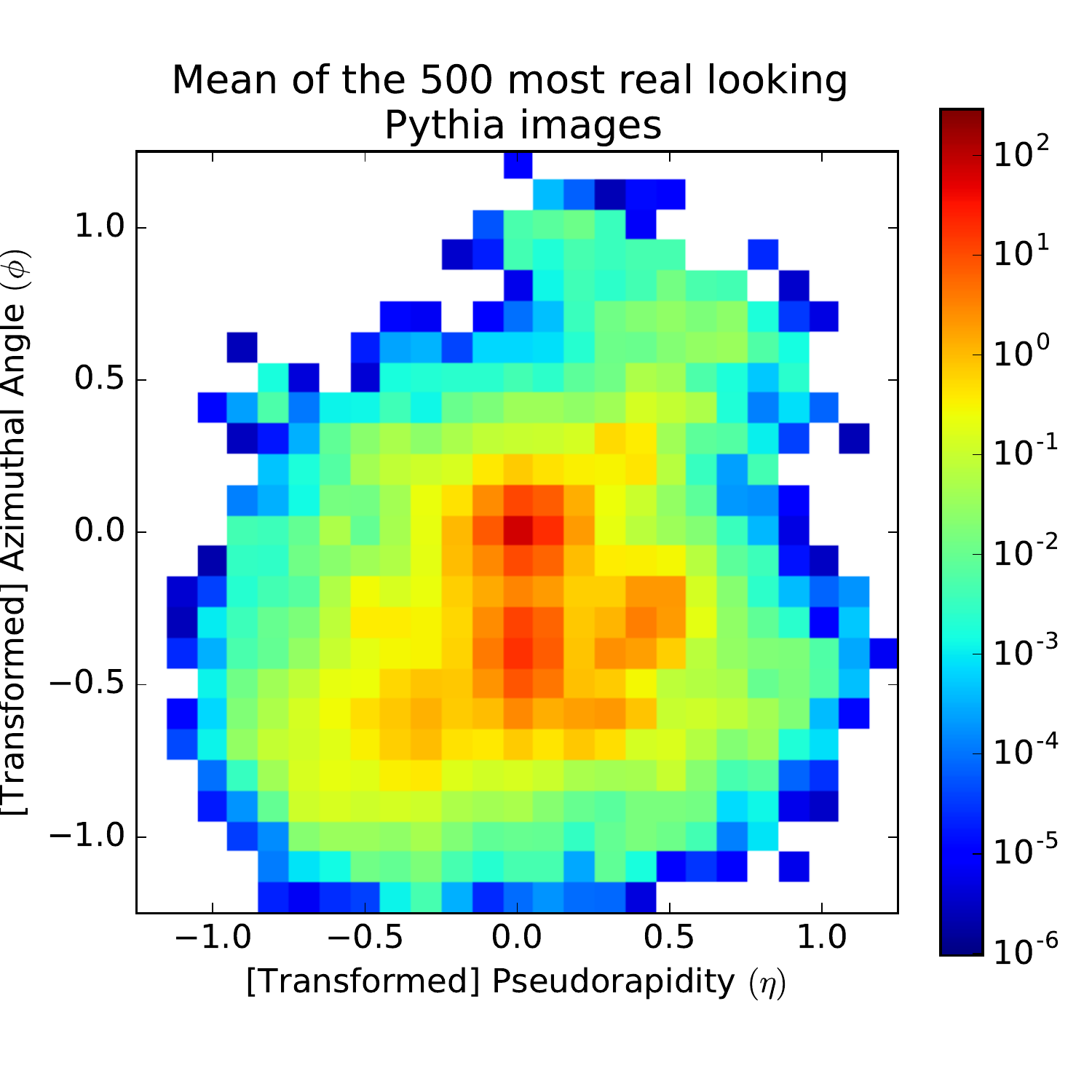}\includegraphics[width=0.3\textwidth]{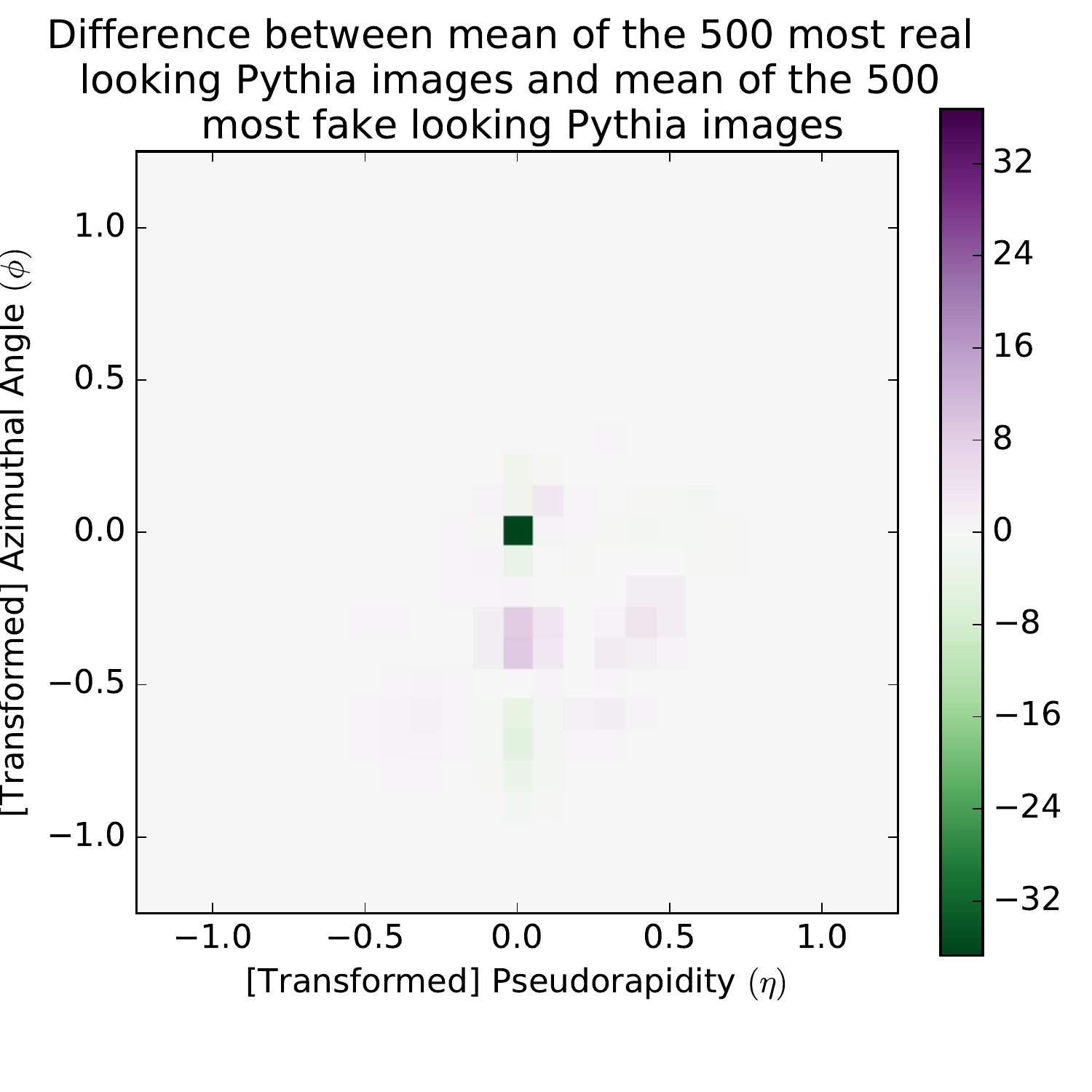}
\includegraphics[width=0.3\textwidth]{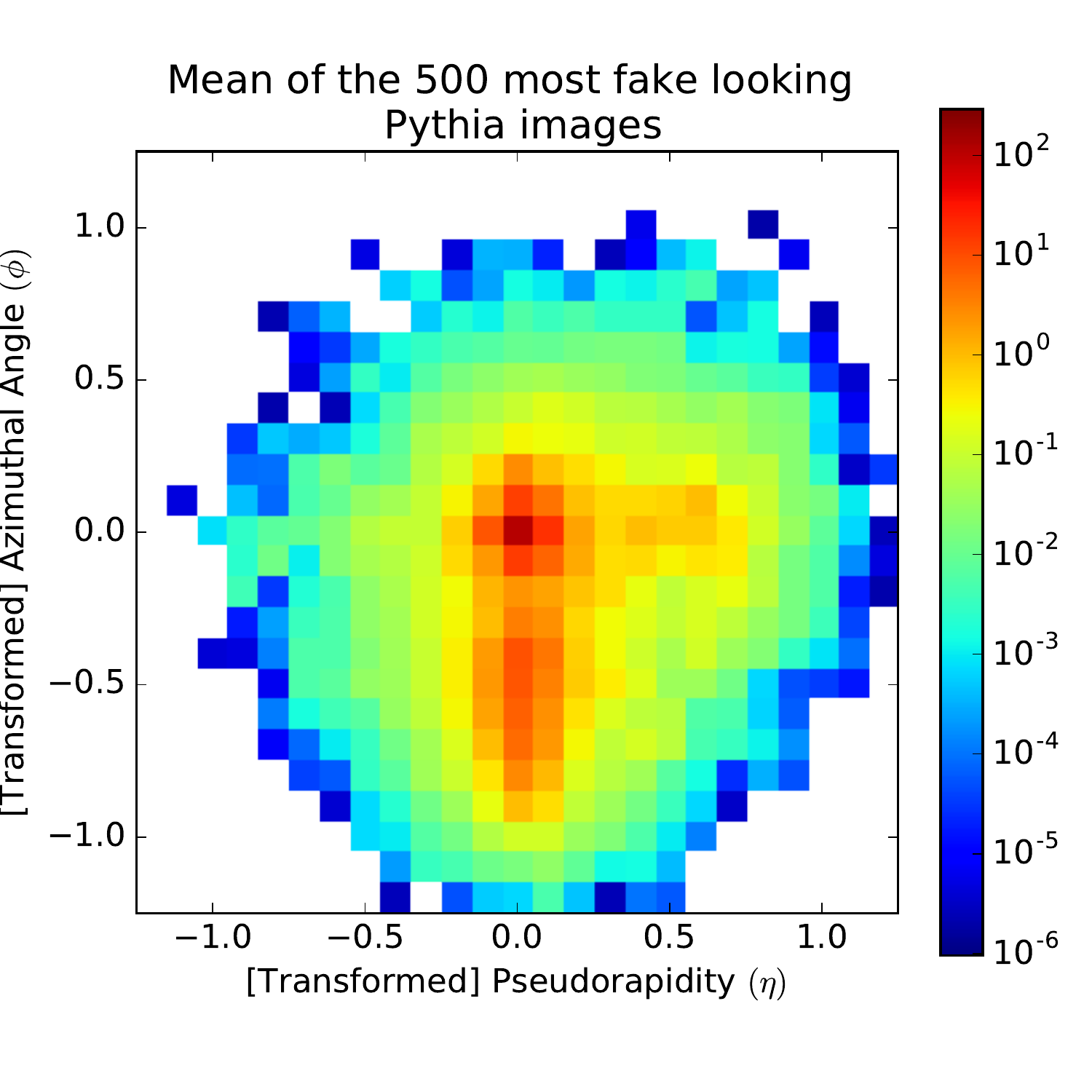}\\
\includegraphics[width=0.3\textwidth]{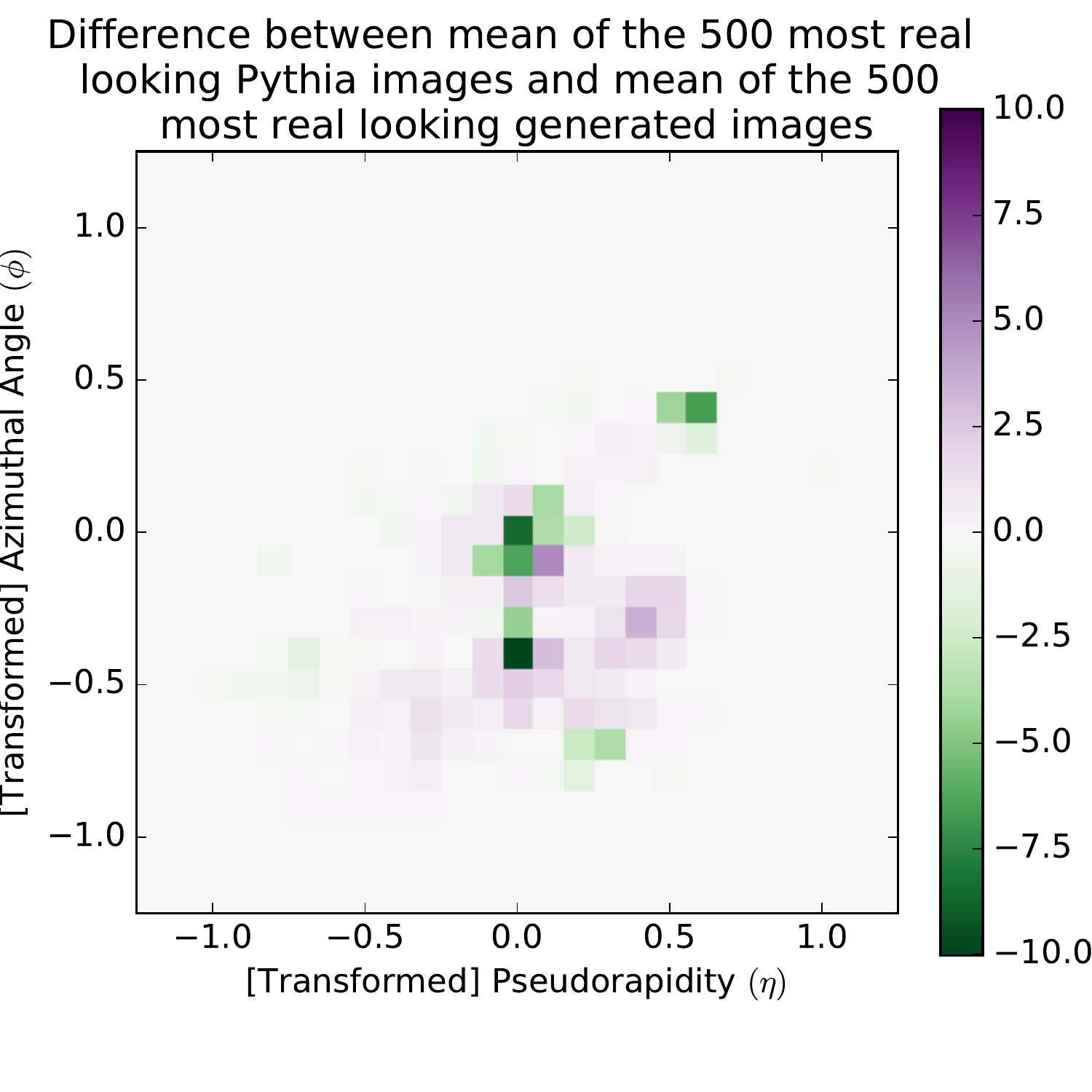}\hspace{50mm}\includegraphics[width=0.3\textwidth]{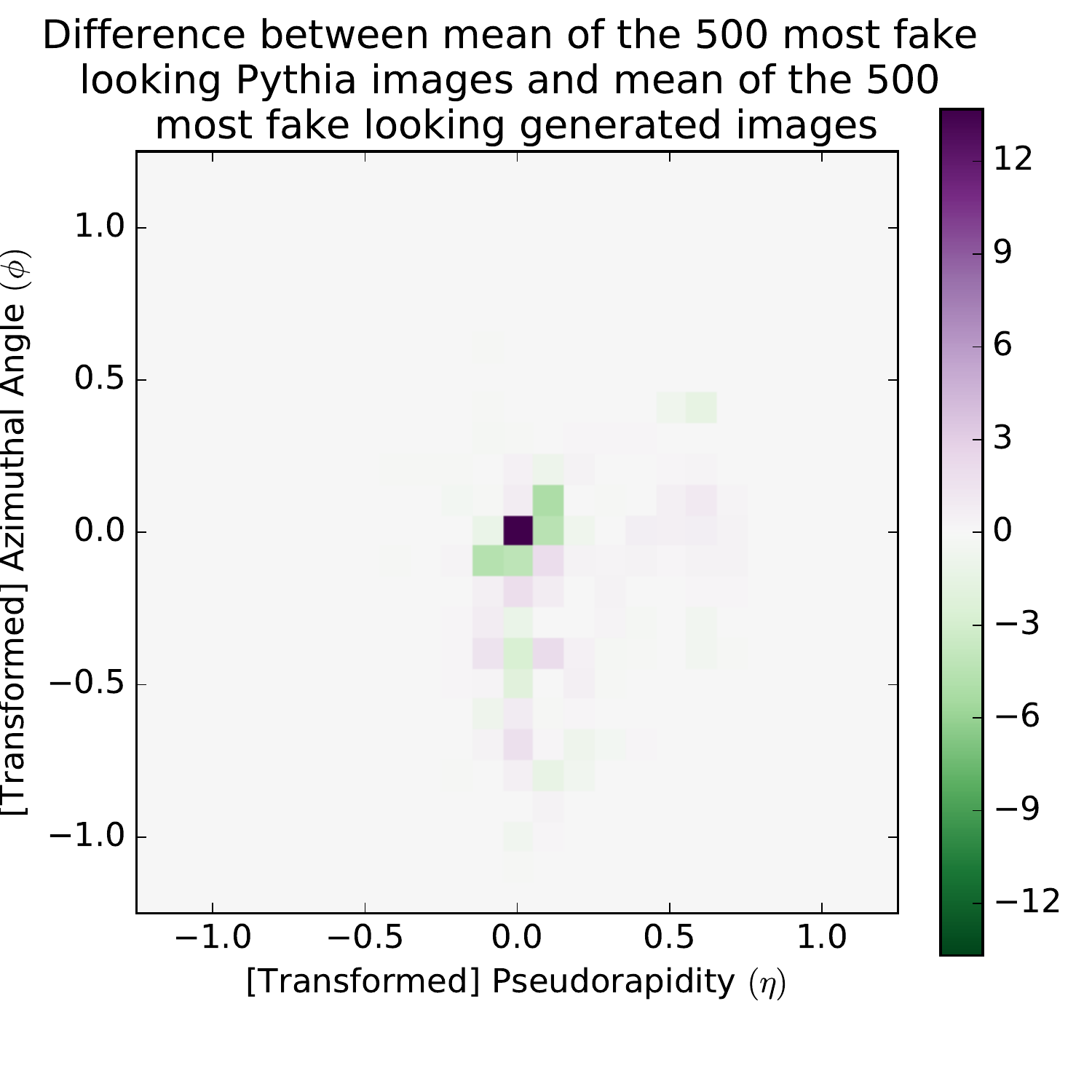}\\
\includegraphics[width=0.3\textwidth]{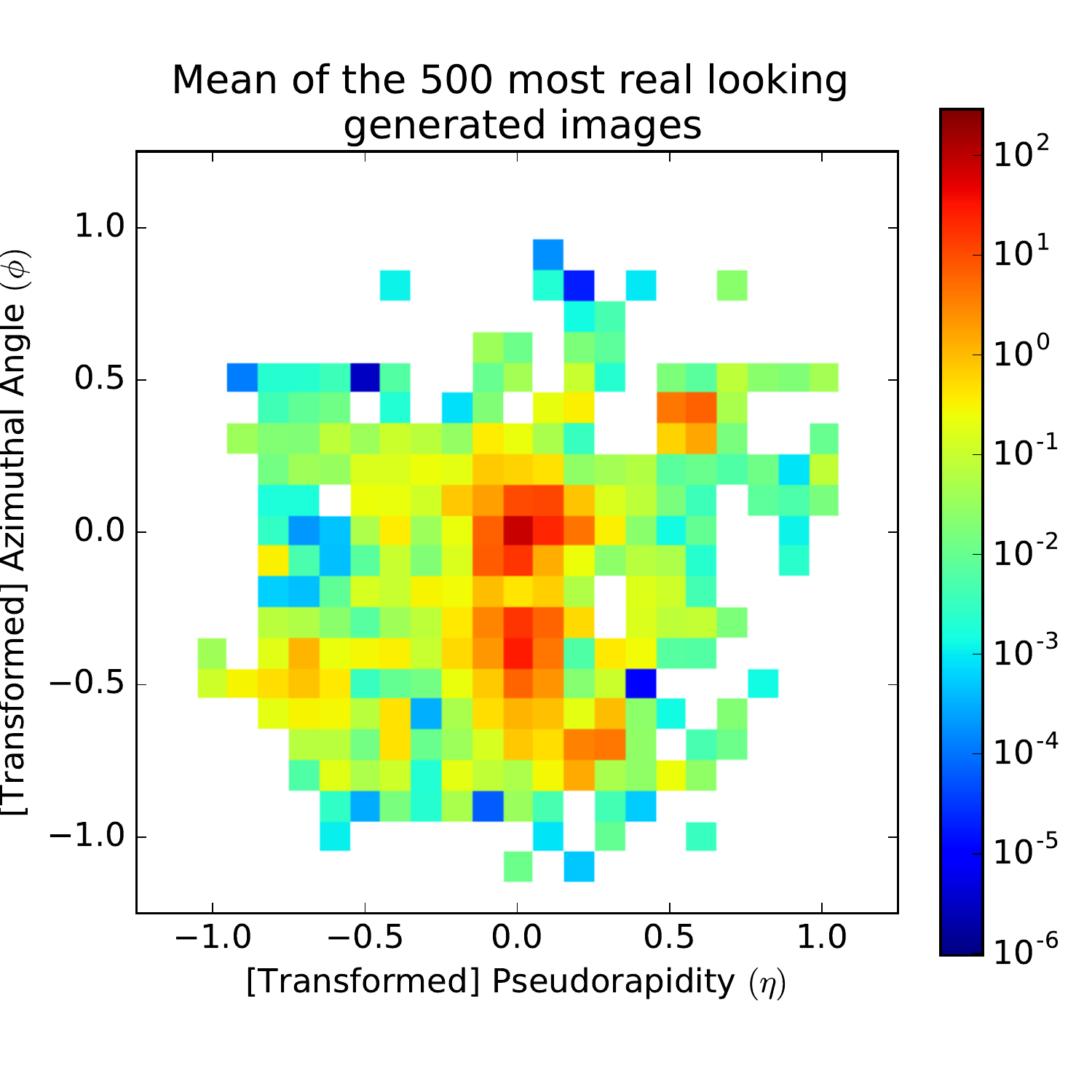}
\includegraphics[width=0.3\textwidth]{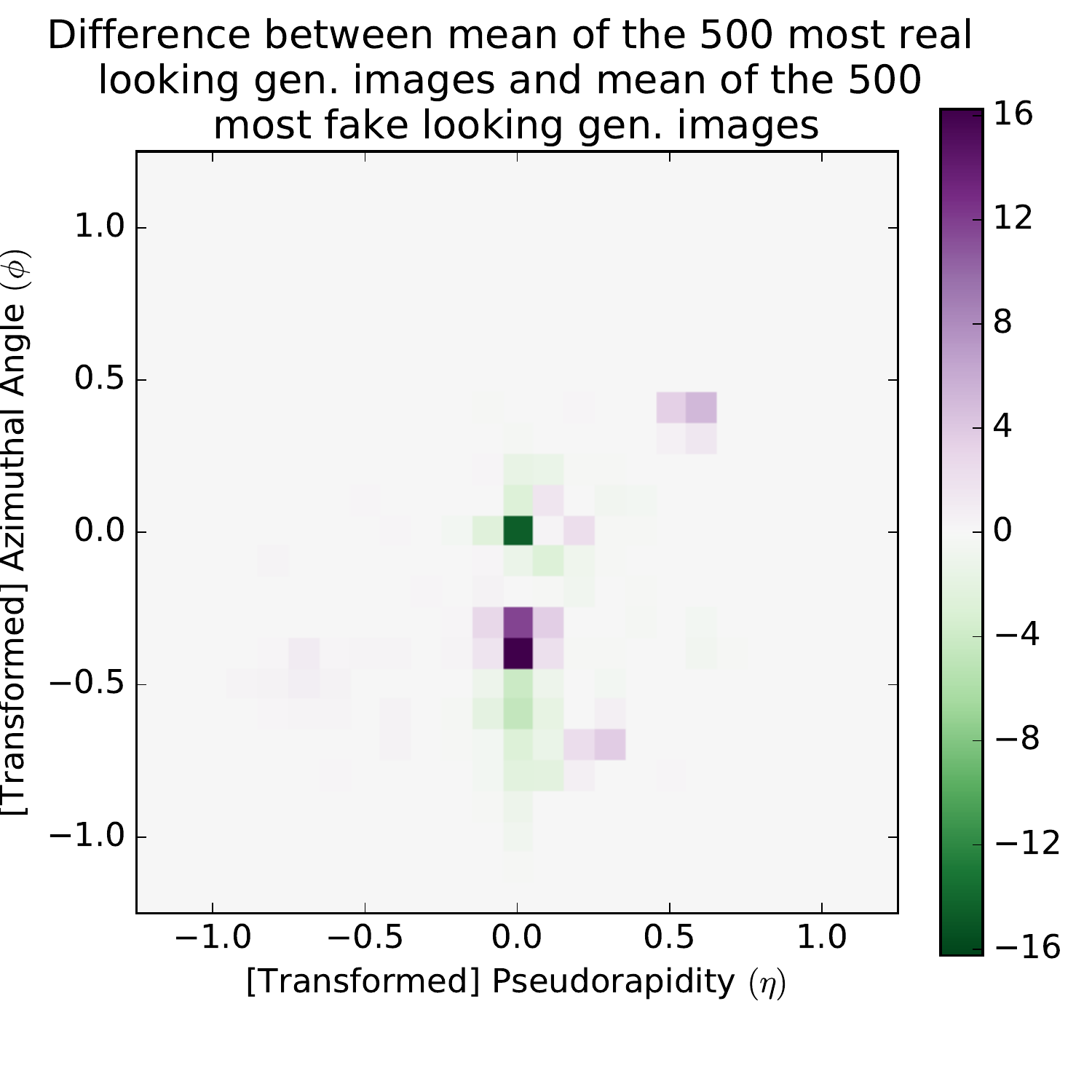}
\includegraphics[width=0.3\textwidth]{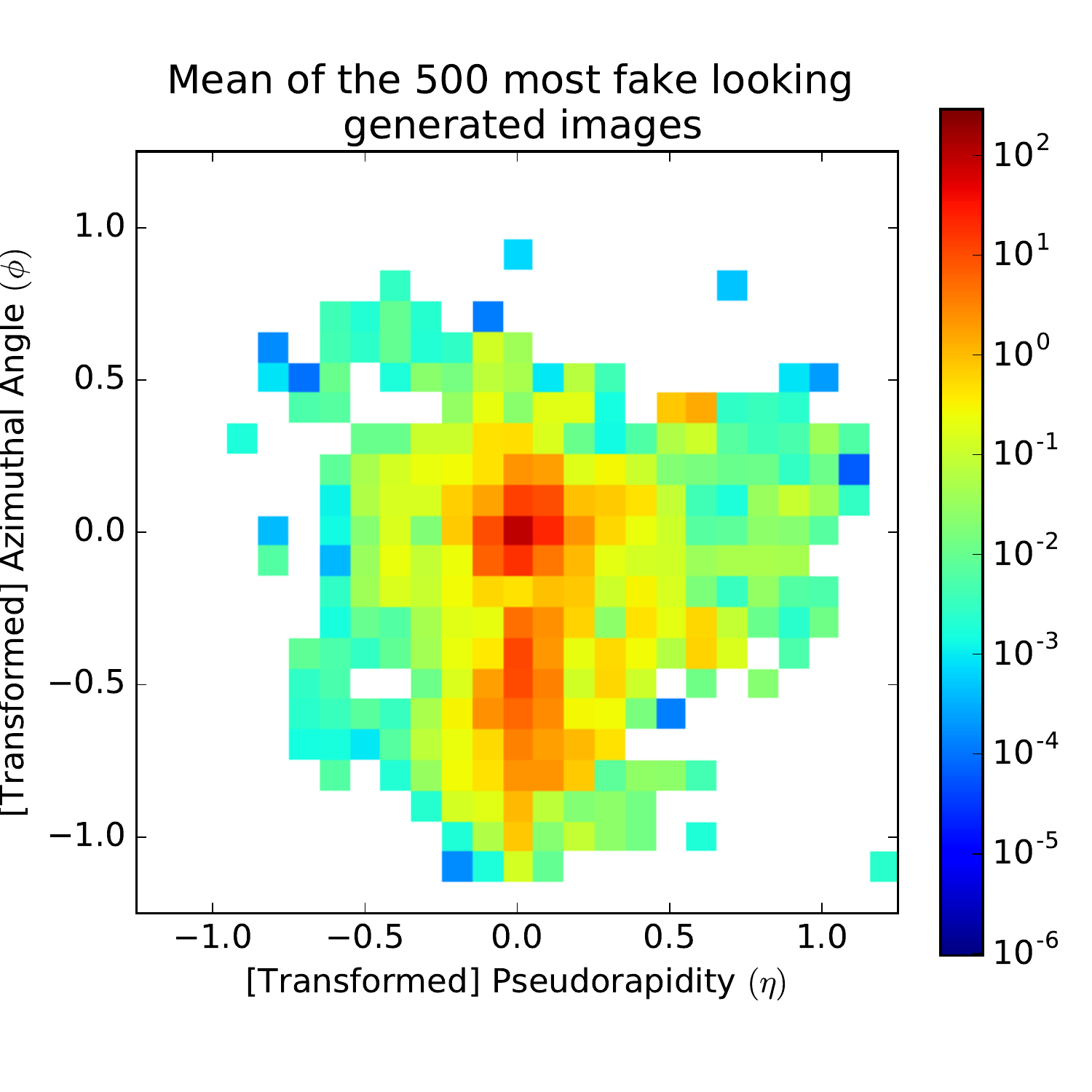}
\caption{Comparison between the 500 most real and most fake looking images, generated by Pythia, on the left, and by the GAN, on the right.}
\label{fig:500_real_fake}
\end{figure} 
} 

The plots shown in Fig.~\ref{fig:500_real_fake} provide unique insight into what features the discriminator is learning and using in its adversarial task of distinguishing generated images from real images. Looking at this figure from top to bottom, we see that, although Pythia images qualitatively differ from generated images, the metric that $D$ applies to discriminate between real and fake images is consistent among the two. The images with the highest $P(\mathrm{real})$ are largely asymmetric, with strong energy deposits in the bottom corners and a much closer location of the subleading subjet to the leading one. The net learns that more symmetric images, with a more uniform activation just to the right side of the leading subjet, stronger intensity in the central pixel, and more distant location of the subleading subjet, appear to be easier to produce for $G$ and are therefore more easily identifiable as fake.   

\begin{figure}[h!]
\centering
\includegraphics[width=0.3\textwidth]{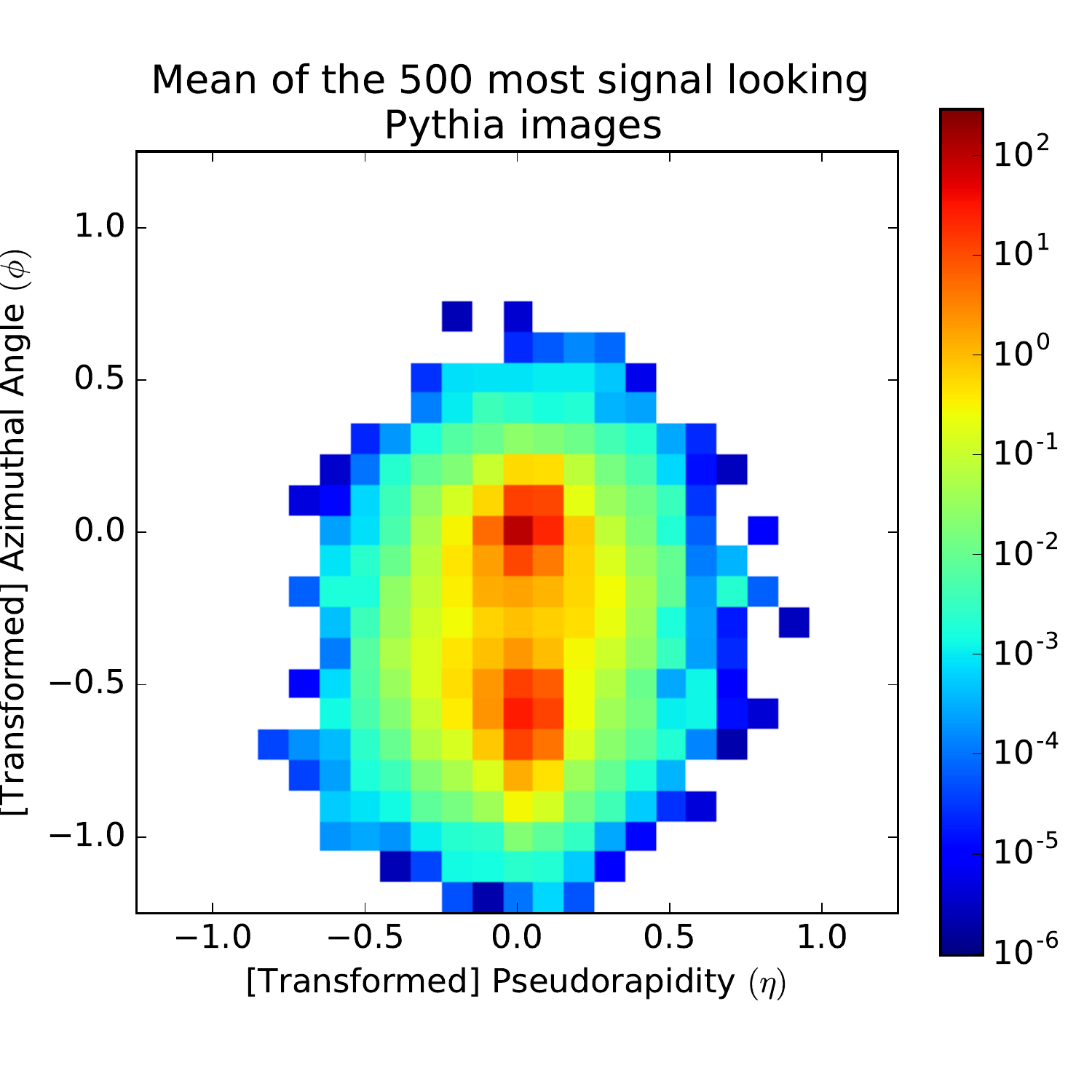}\includegraphics[width=0.3\textwidth]{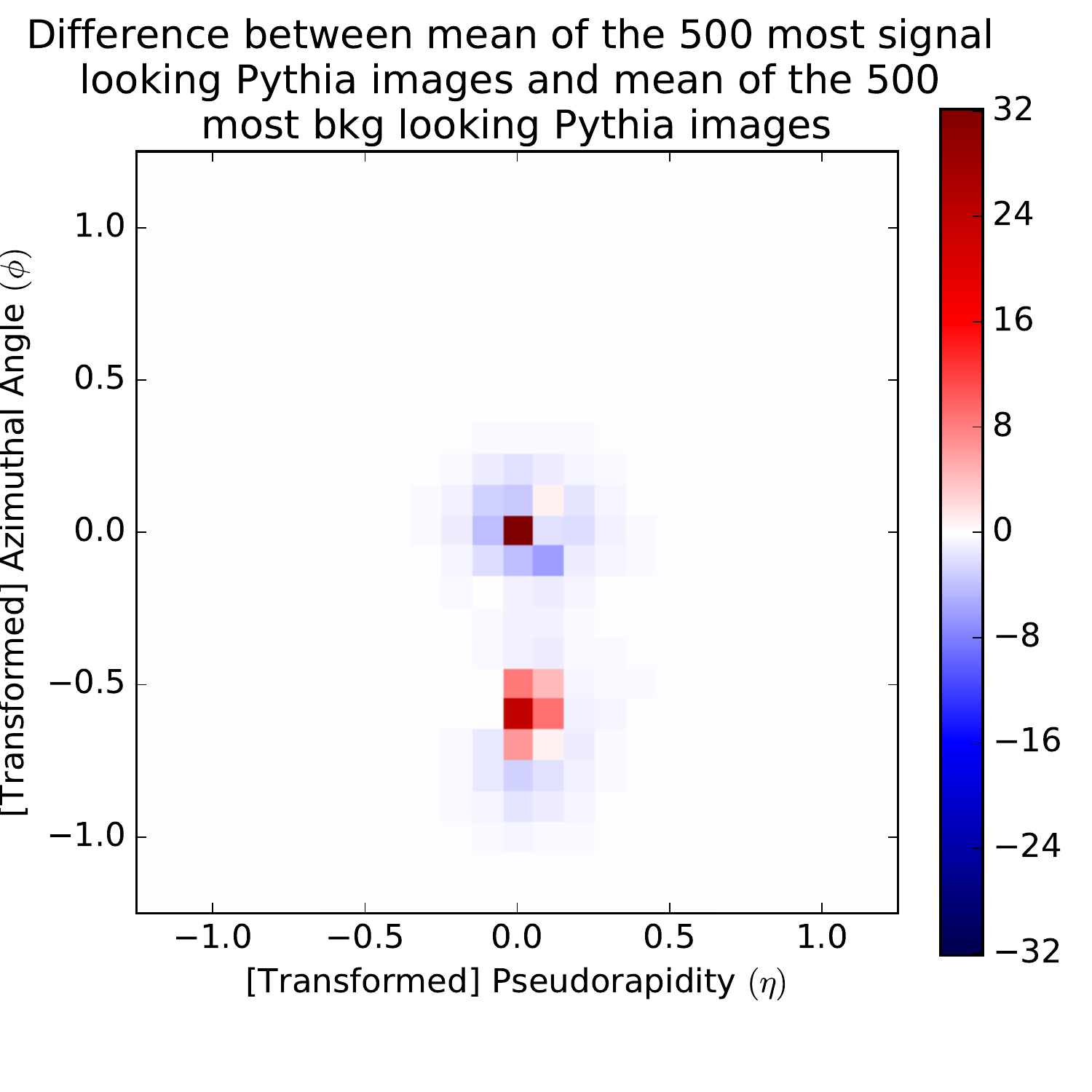}
\includegraphics[width=0.3\textwidth]{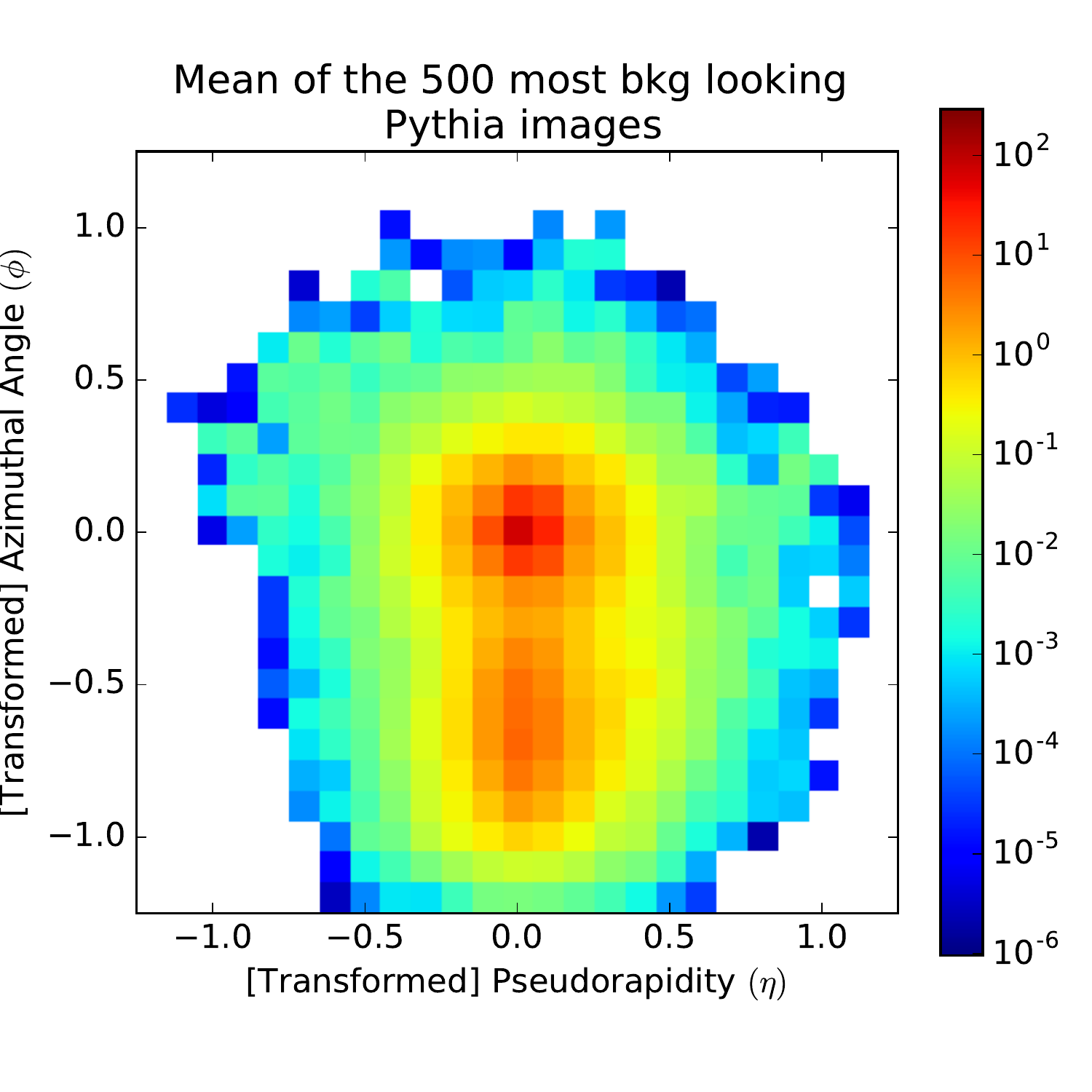}\\
\includegraphics[width=0.3\textwidth]{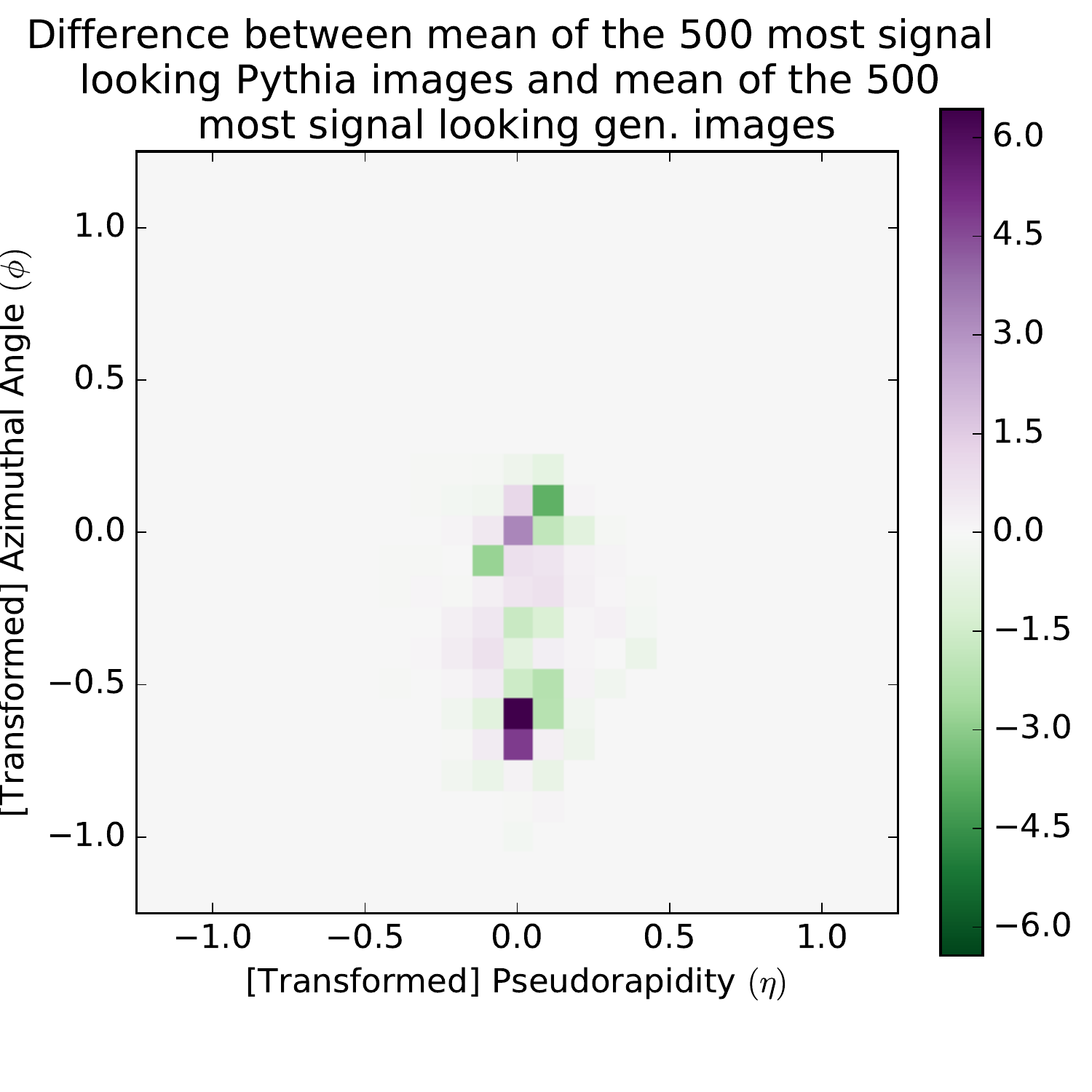}\hspace{50mm}\includegraphics[width=0.3\textwidth]{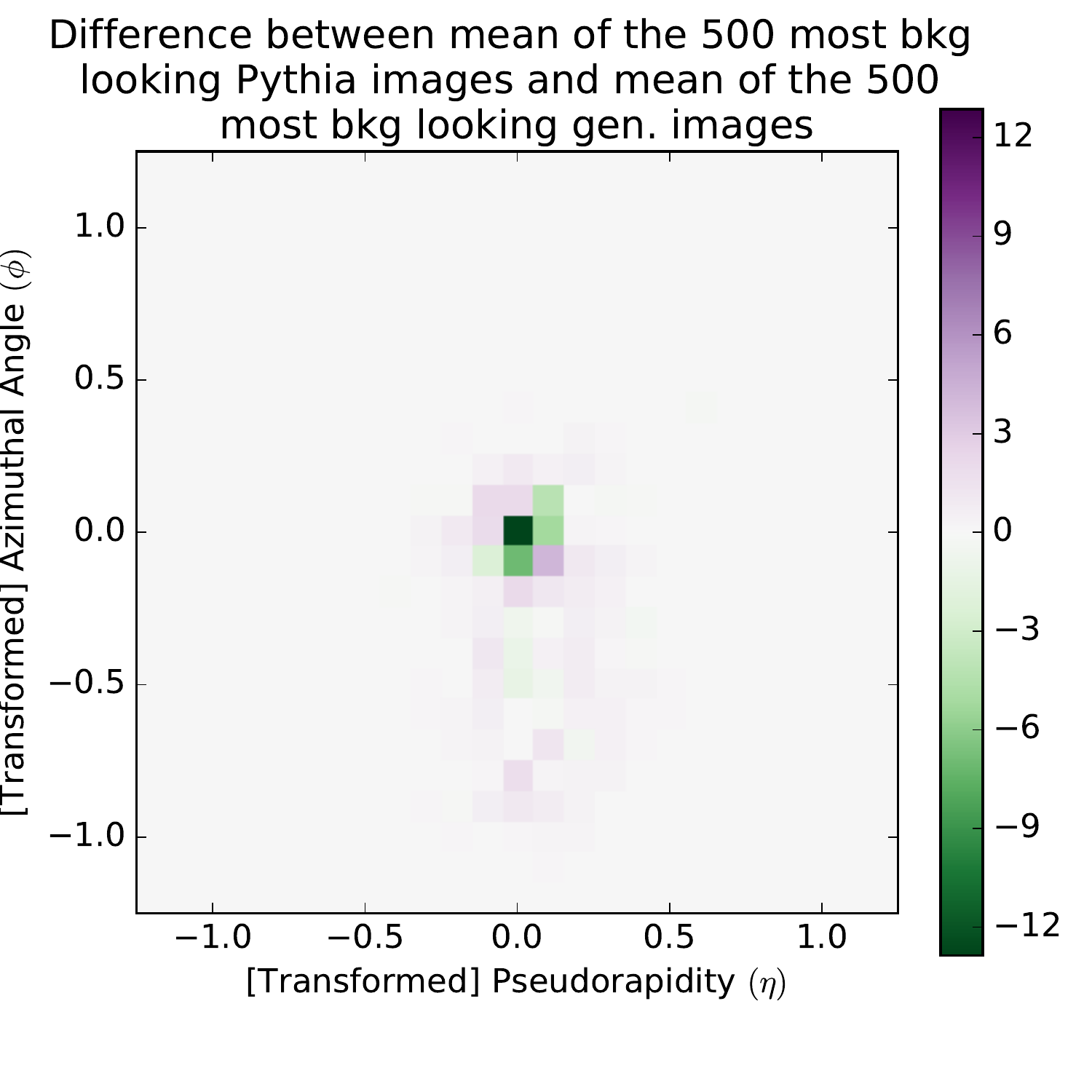}\\
\includegraphics[width=0.3\textwidth]{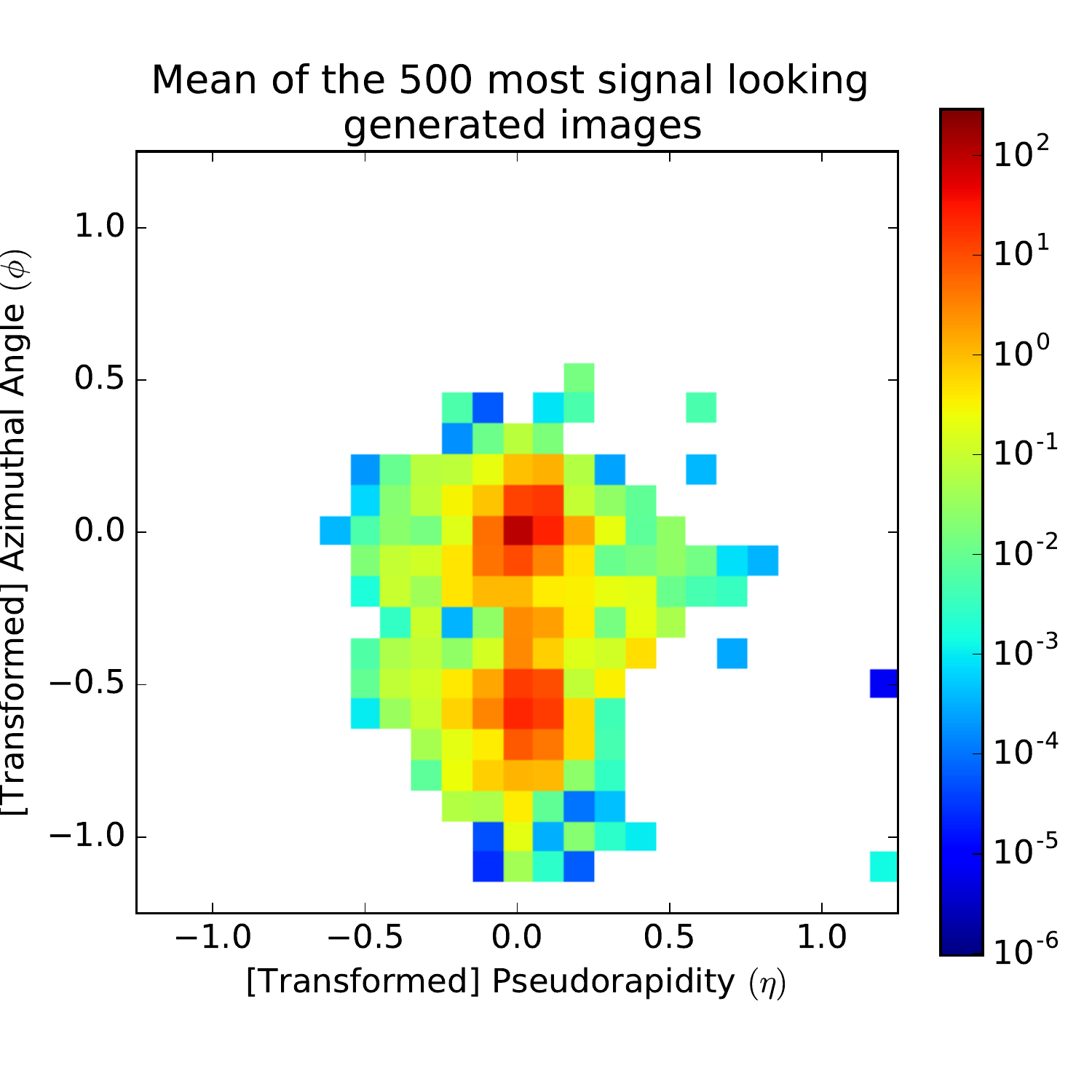}
\includegraphics[width=0.3\textwidth]{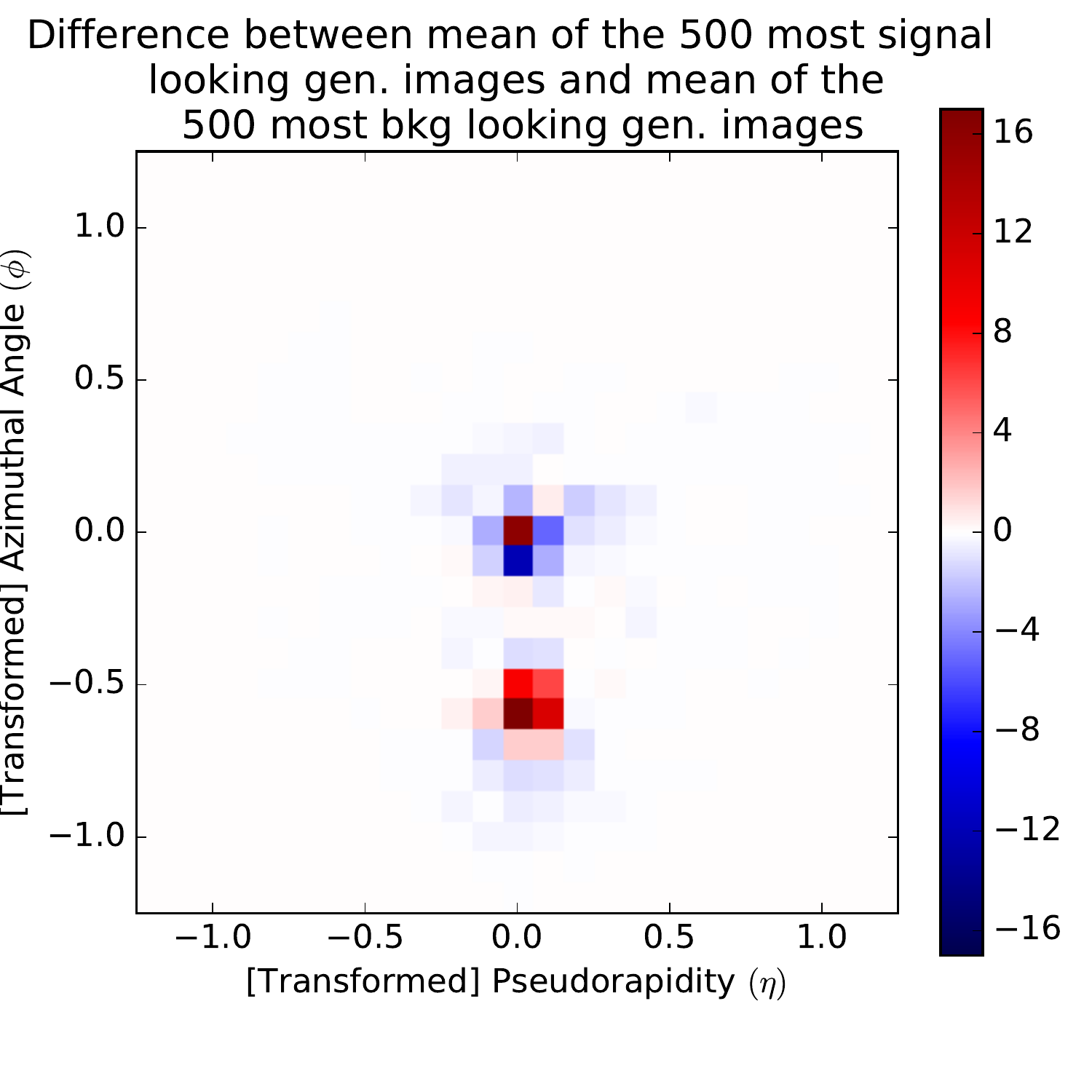}
\includegraphics[width=0.3\textwidth]{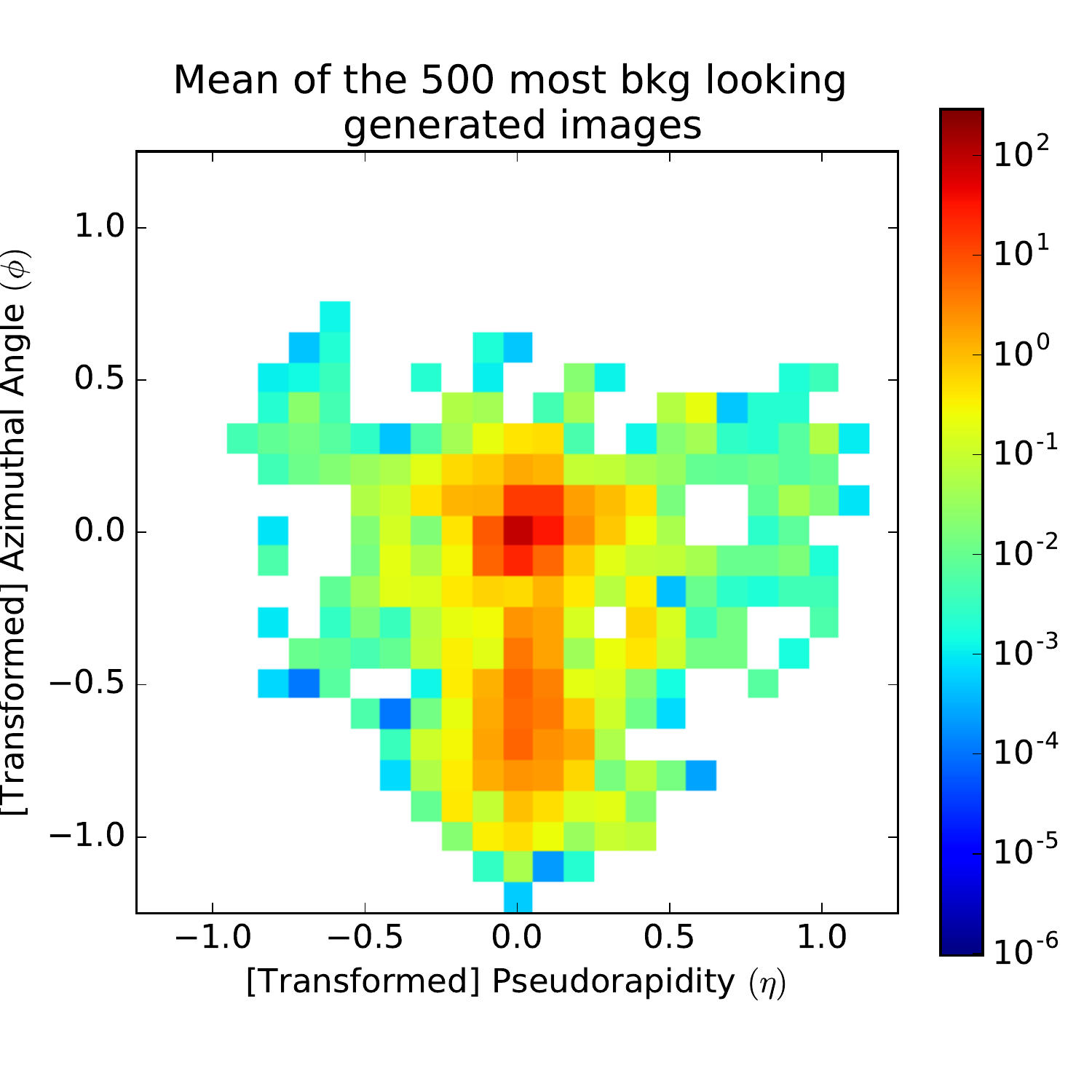}
\caption{Comparison between the 500 most signal and most background looking images, generated by Pythia, on the left, and by the GAN, on the right.}
\label{fig:500_signal_bkg}
\end{figure}

In addition, we can isolate the primary information that $D$ pays attention to when classifying signal and background images. The learned metric is consistently applied to both Pythia and GAN images, as shown in Fig.~\ref{fig:500_signal_bkg}. When identifying signal images, $D$ learns to looks for more concentrated images, with well-defined two prong structure. On the other hand, the network has learned that background images have a wider radiation pattern and a more fuzzy structure around the location of the second subjet.

\subsection{Generator}
\label{ssec:gen}

To further understand the generation process of jet images using GANs, we explore the inner workings of the generator network. As outlined in Sec.~\ref{ssub:implementation_and_training}, $G$ consists of a 2D convolution followed by 3 consecutive locally-connected layers, the last of which yields the final generated images. 
By peeking through the layers of the generator and following the path of a jet image in the making, we can visually explore the steps that lead to the production of jet images with their desired physical characteristics.   

\begin{figure}[h!]
    \centering
    \includegraphics[width=\textwidth]{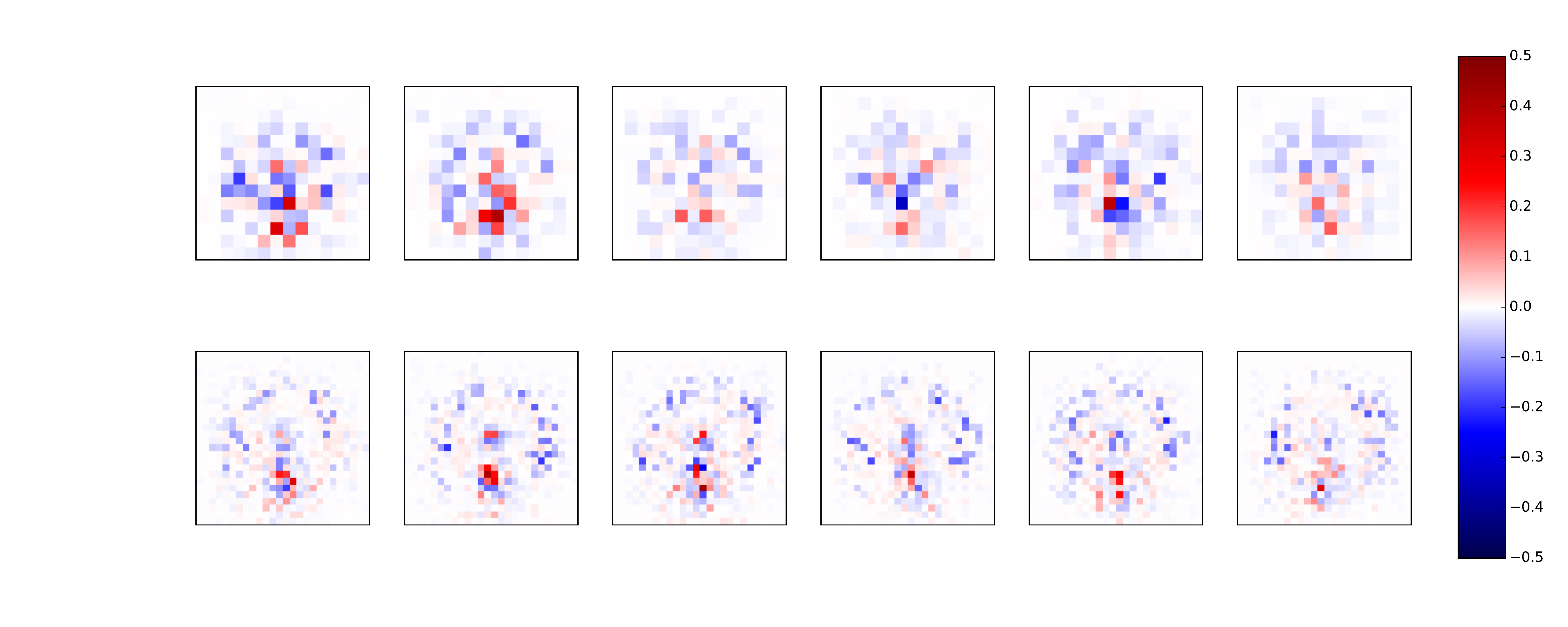}
    \caption{Pixel activations of images representing the various channels in the outputs of the two locally-connected hidden layers that form the generator, highlighting the difference in activation between the production of the average signal and average background samples. The first row represents the six $14\times14$ channels of the first LC layer's output; the second row represents the six $26 \times26$ channels of the second LC layer's output.}
    \label{fig:g_progress}
\end{figure}

We probe the network after each locally-connected hidden layer; we investigate how the average signal and average background images develop into physically distinct classes as a function of depth of the generator network, by plotting each channel separately (Fig.~\ref{fig:g_progress}). The output from the first locally-connected layer consists of six $14\times14$ images (\textit{i.e.}, an $14\times14\times6$ volume), and the second consists of six $26\times26$ images. The red pixels are more strongly activated for signal images, while blue pixels activate more strongly in the presence of background images, which proves the fact that the generator is learning, from its very early stages that spread out energy depositions are needed to produce more typical background images.

\subsection{Discriminator}
\label{ssec:disc}

The discriminator network undertakes the task of hierarchical feature extraction and learning for classification. Its role as an adversary to the generator is augmented by an auxiliary classification task. With interpretability being paramount in high energy physics, we explore techniques to visualize and understand the representations learned by the discriminator and their correlation with the known physical processes driving the production of jet images. 

\begin{figure}[H]
\centering
 \includegraphics[width=0.8\textwidth]{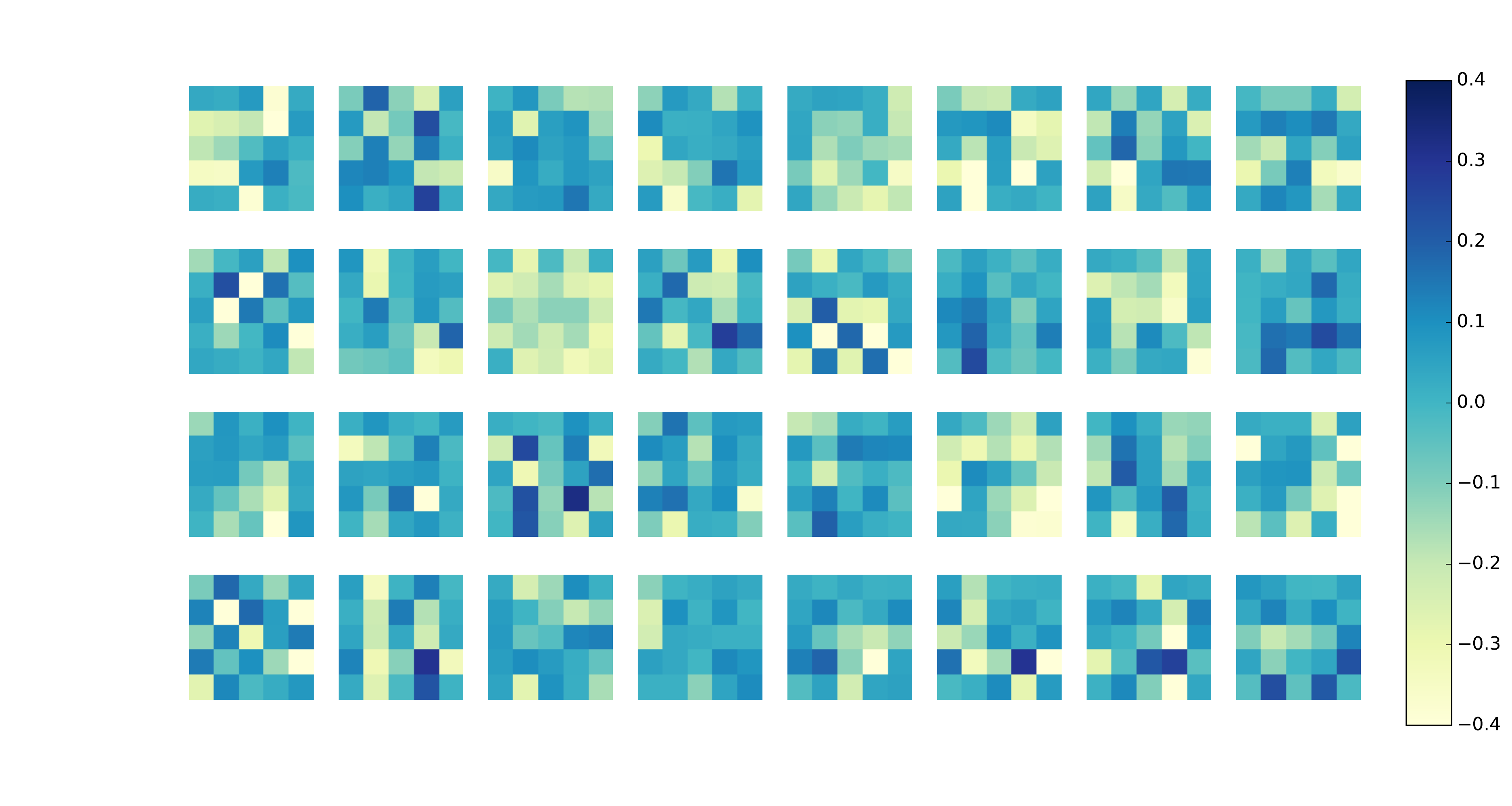}\\
\includegraphics[width=0.8\textwidth]{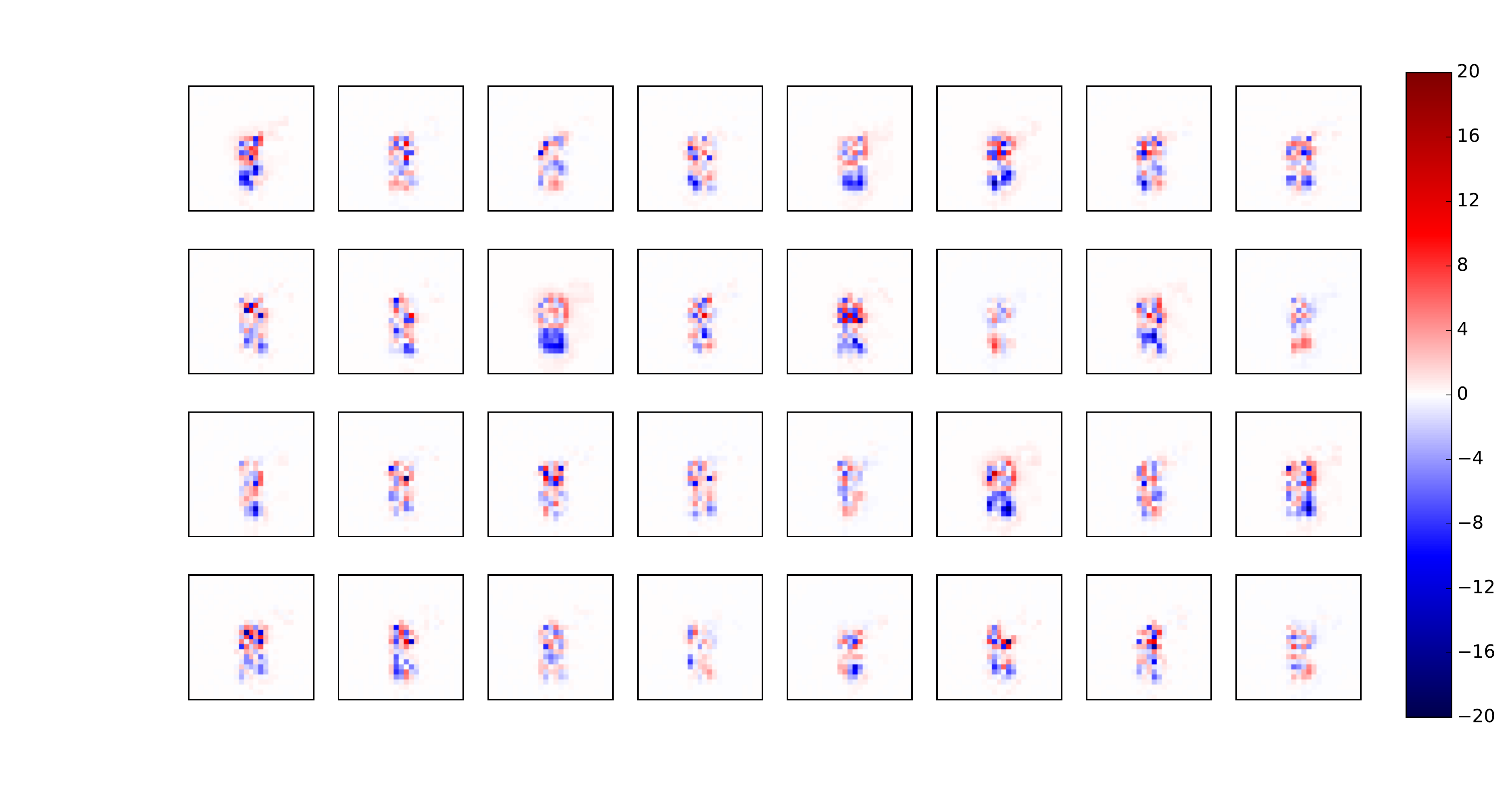}\\
\includegraphics[width=0.8\textwidth]{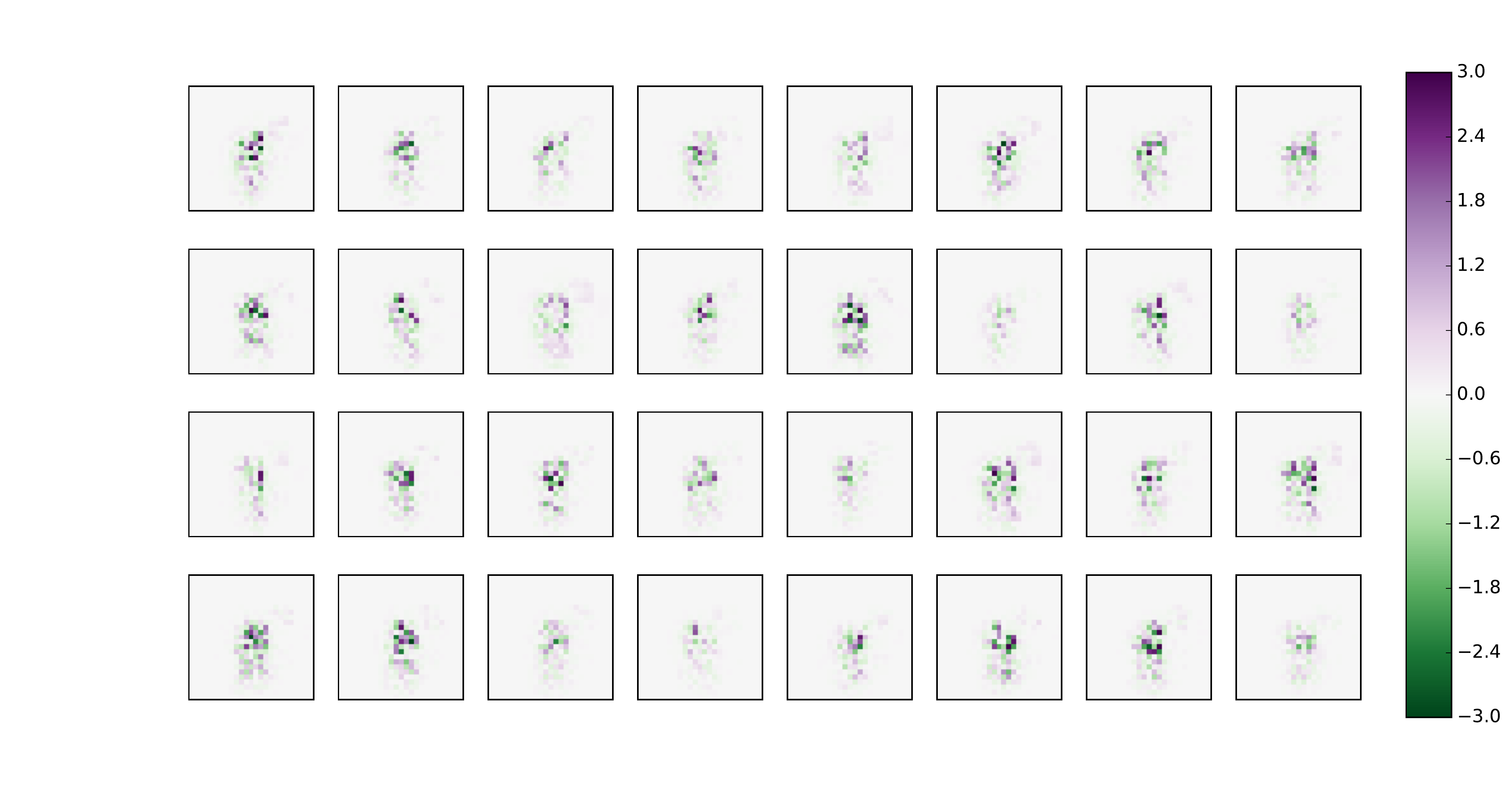}
\caption{Convolutional filters in the first layer of the discriminator (top), their convolved version with the difference between the average signal and background generated image (center), and their convolved version with the difference between average Pythia and average generate image.}
\label{fig:filters}
\end{figure}

The convolutional layer at the beginning of the network provides the most significant visual aid to guide the interpretation process. In this layer, we convolve 32 $5\times5$ learned convolutional filters with equi-sized patches of pixels by sliding them across the image. 
The weights of the $5\times5$ matrices that compose the convolutional kernels of the first hidden layer are displayed in a pictorial way in the top panel of Fig.~\ref{fig:filters}.

Each filter is then convolved with two physically relevant images: in the middle panel, the convolution is performed with the difference between the average GAN-generated signal image and the average GAN-generated background image; in the lower panel, the convolution is performed with the difference between the average Pythia image drawn from the data distribution and the average GAN-generated image. 
By highlighting pixel regions of interest, these plots shed light on the highest level representation learned by the discriminator, and how this representation translates into the auxiliary and adversarial classification outputs.

\begin{figure}[h!]
\centering
\includegraphics[width=0.3\textwidth]{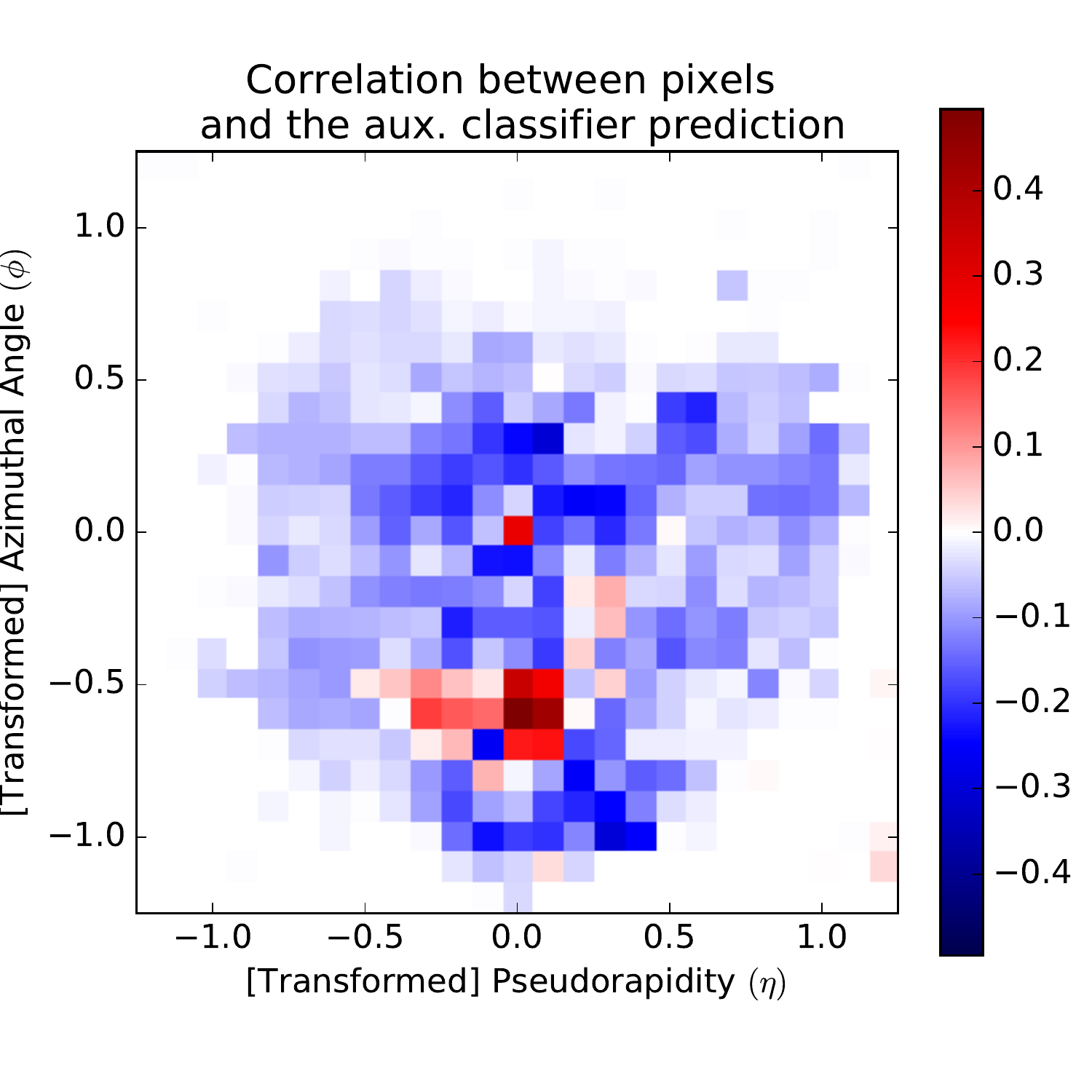}
\includegraphics[width=0.3\textwidth]{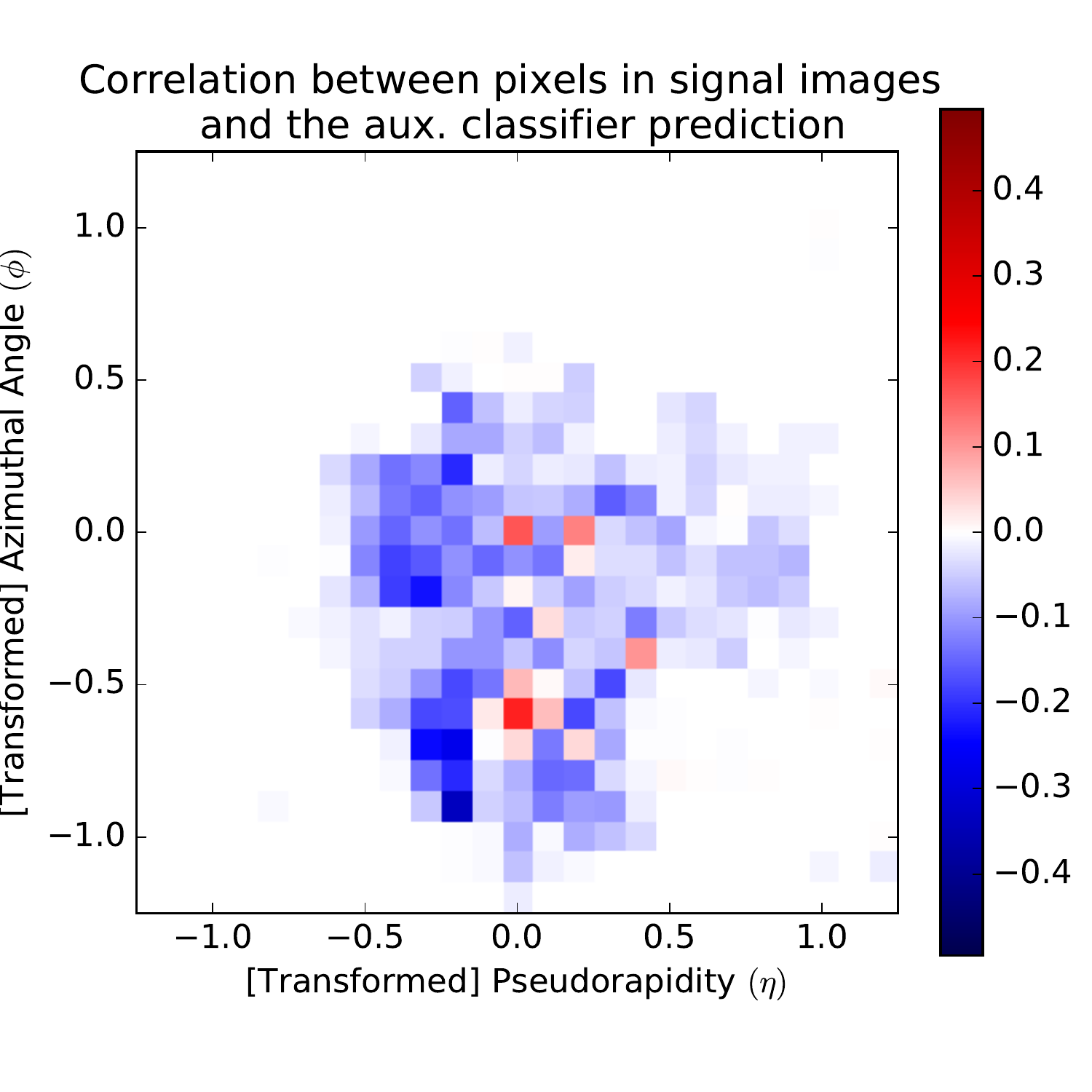}
\includegraphics[width=0.3\textwidth]{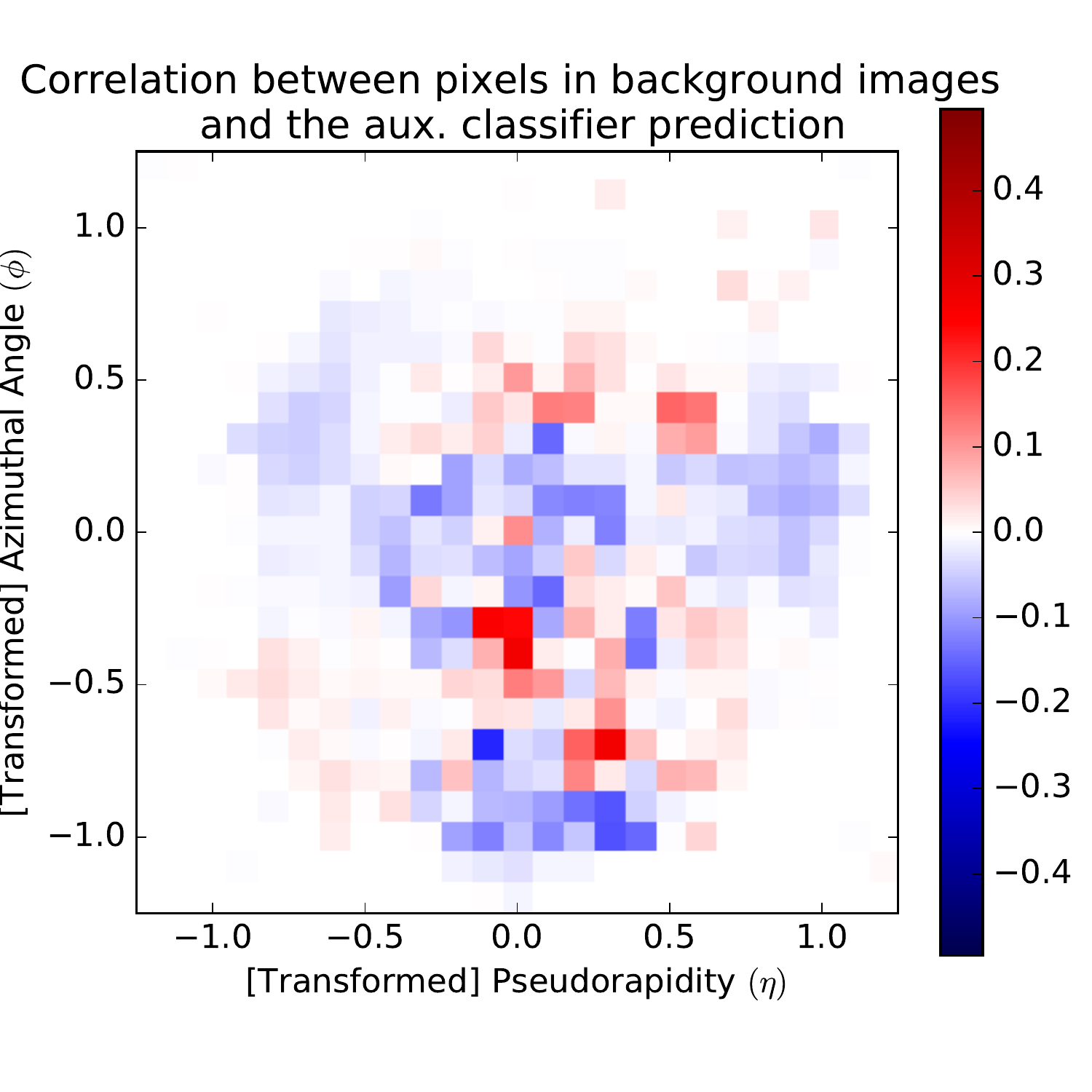}\\
\includegraphics[width=0.3\textwidth]{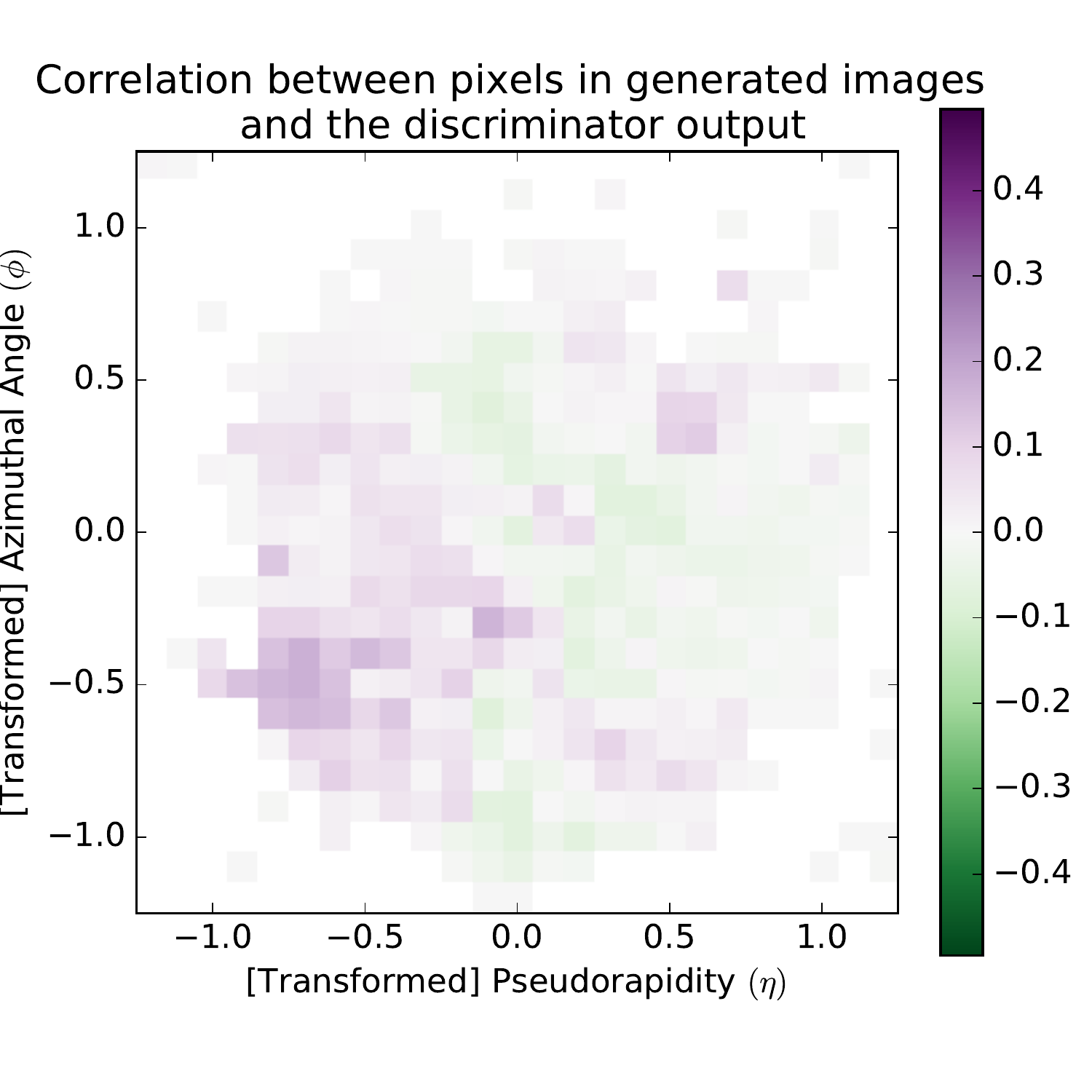}
\includegraphics[width=0.3\textwidth]{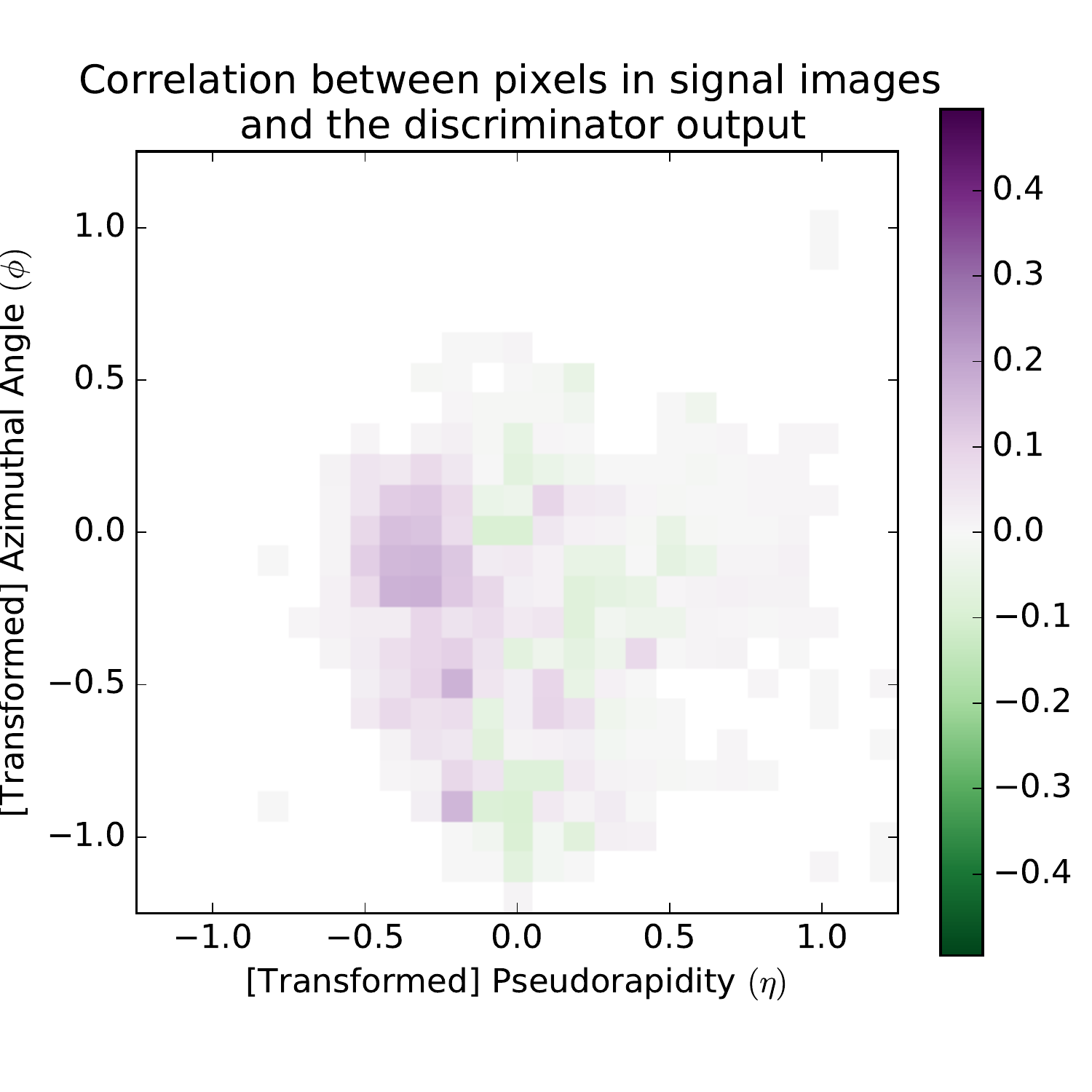}
\includegraphics[width=0.3\textwidth]{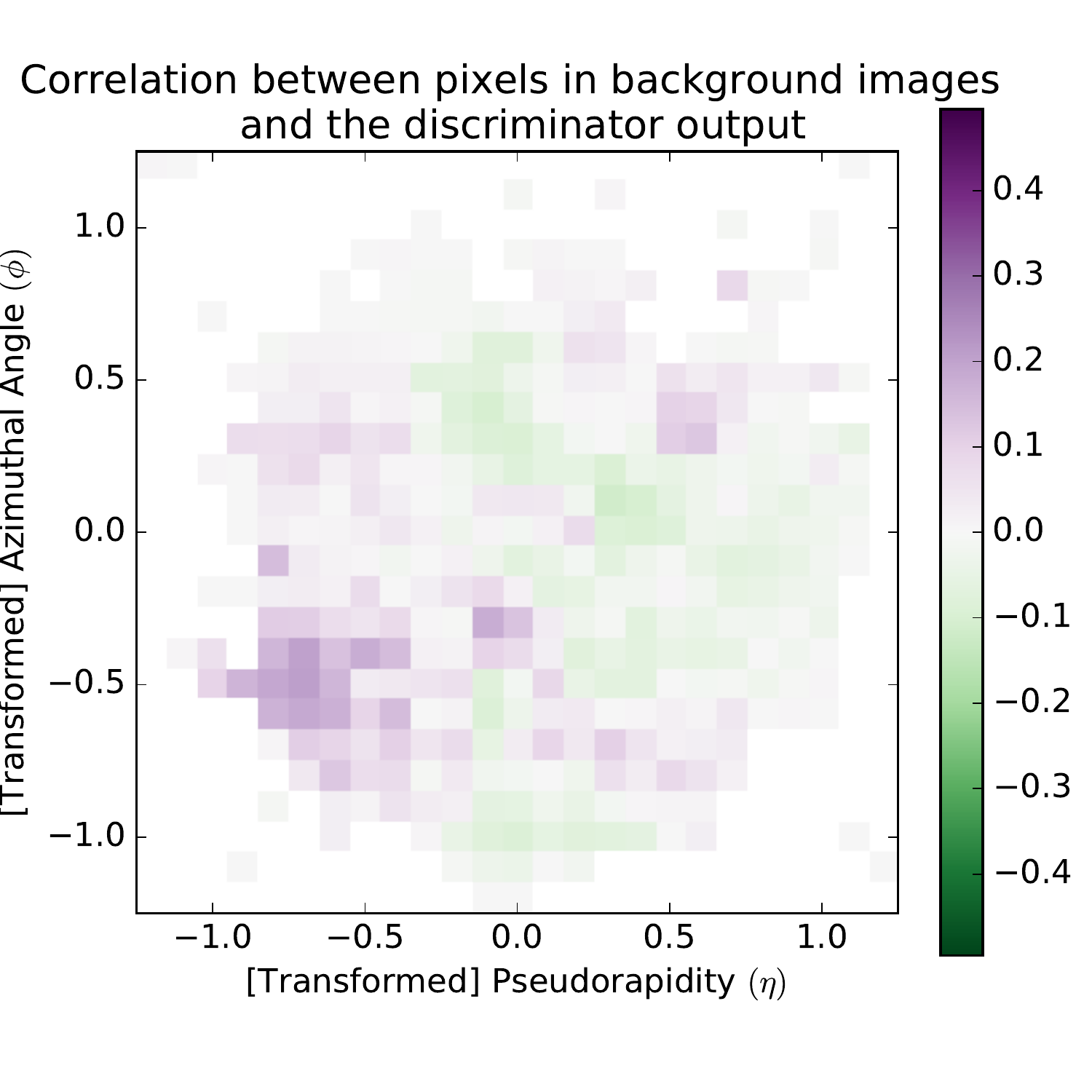}
\caption{Per-pixel linear correlation with the discriminator auxiliary (top) and adversarial (bottom) classification output for combined signal and background images (left), signal only images (center), and background only images (right).}
\label{fig:correlations_D}
\end{figure}

Finally, we compute linear correlations between the average image's pixel intensities and each of the discriminator's outputs, and visualize the learned location-dependent discriminating information in Fig.~\ref{fig:correlations_D}. These plots accentuate the distribution of distinctive local features in jet images. 
Additional material in Appendix~\ref{app:auxmaterial} is available to further explore the effects of conditioning on the primary and auxiliary $D$ outputs on the visualization techniques used in this paper, as well as the the correlation between $D$'s output and the physical process that originated the jet (Fig.~\ref{fig:confusion_fake_real}).

\subsection{Training Procedure Observations}
\label{ssec:train}

\begin{figure}[h!]
\centering
\includegraphics[width=0.5\textwidth]{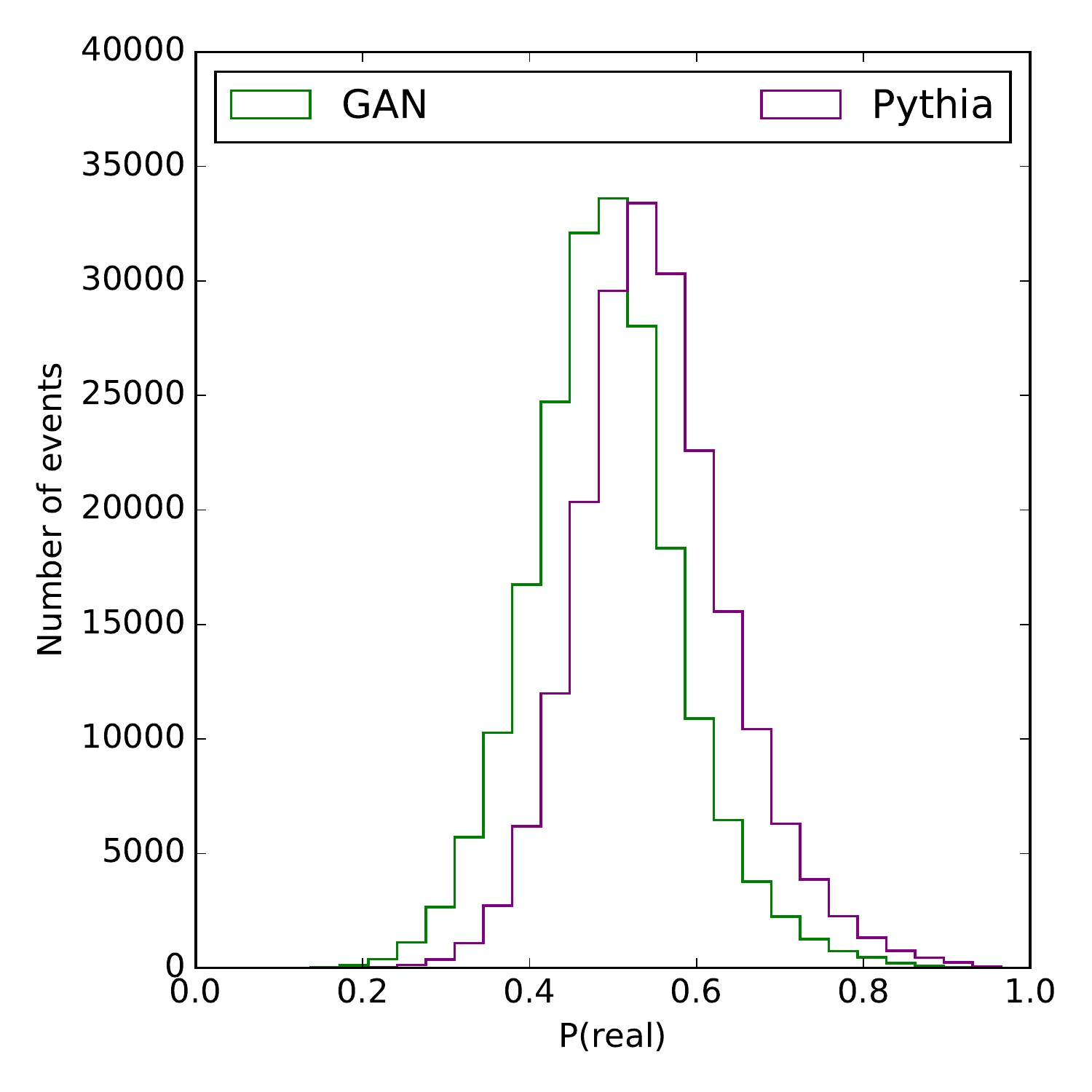}
\caption{Histograms of the discriminator's adversarial output for generated and real images at the epoch chosen to generate the images used in this paper (34th epoch). }
\label{fig:D_gan_pythia}
\end{figure}

During training, a handle on the performance of $D$ is always directly available through the evaluation of its ability to correctly classify real versus GAN-generated images. We visualize the discriminator's performance in Fig.~\ref{fig:D_gan_pythia} by constructing histograms of the output of the discriminator evaluated on GAN and Pythia generated images. Perfect confusion would be achieved if the discriminator output were equal to 1/2 for every image. However, this visualization fails at providing quantifiable information on the performance achieved by $G$. In fact, a high degree of confusion can also be symptomatic of undertraining in both $G$ and $D$ or mode-collapse in $G$, and is observed in the first few epochs if $D$ is not pre-trained. Similarly, a slow divergence towards $D$ winning the game against $G$ cannot lead to the conclusion that the performance of $G$ is not improving; it might simply be doing so at a slower pace than $D$, and the type of representation shown in Fig.~\ref{fig:D_gan_pythia} does not help us identify when the peak performance of $G$ is obtained.

Instead, we use convergence of generated mass and $p_T$ to the corresponding real distributions to select a stopping point for the training (Fig.~\ref{fig:m_pt_tau21}). We select the 34th epoch of training to generate the images shown in this paper, but qualitatively similar results can be observed after only a handful of epochs. At this time, the training seems to have almost reached the equilibrium point in which $D$ is unsure of how to label most samples, regardless of their true origin, without any strong dependence on physical process that generated the image (Fig.~\ref{fig:confusion_fake_real}).

\subsection{Architecture Comparisons}

\label{ssec:arch}

With a toolkit of methods for validating the effectiveness of a GAN for jet images, we can now compare the LAGAN performance against other architectures. To this end, we focus on the jet image observables introduced in Sec.~\ref{ssec:images}.  To quantify the preservation of these physically meaningful manifolds under the data distribution and learned distribution, and to standardize generative model performance comparisons, we design a simple, universal scoring function.

Consider the true data distribution of jet images $\mathcal{D}$, and a generated distribution $\mathcal{S}$ from a particular generative model. Define $\mathcal{M}_{\mathcal{D}}(x)$ and $\mathcal{M}_{\mathcal{S}}(x)$ to be the distributions of any number of physically meaningful manifolds under the data and generated distributions respectively. 
We require that both conditional distributions be well modeled, or more formally that, for a sensibly chosen distance metric $d(\cdot,\cdot)$, $d\left(\mathcal{M}_{\mathcal{S}}(x\vert c), \mathcal{M}_{\mathcal{D}}(x\vert c)\right)\rightarrow 0\ \forall c\in \mathcal{C}$, where $\mathcal{C}$ is the set of classes. To ensure good convergence for all classes, the performance scoring function is designed to minimize the maximum distance between the real and generated class-conditional distributions. Formally, such a minimax scoring function is defined as
\begin{equation}
\label{eqn:metric}
    \sigma(\mathcal{S}, \mathcal{D}) = \max_{c\in \mathcal{C}} d(\mathcal{M}_{\mathcal{D}}(x\vert c), \mathcal{M}_{\mathcal{S}}(x\vert c))
\end{equation}

Note that Eq.~\ref{eqn:metric} is minimized when the generative model exactly recovers the data distribution. For this application, the similarity metric $d$ has been chosen to be the Earth Mover's Distance~\cite{earthmover}.

Previous work on the classification of jet images from boosted $W$ bosons and QCD jets found the combination of jet mass and $\tau_{21}$ to be the most discriminating high-level variable combination~\cite{deOliveira:2015xxd}. Therefore, these two features are chosen to form the reduced manifold space used for performance evaluation. The empirical 2D Probability Mass Function over $(m, \tau_{21})$ is discretized into a $40\times 40$ equispaced grid containing the full support, and the distance between any two points in the PMF is defined to be the Euclidean distance of the respective coordinate indices. This distance is used in the internal flow optimization procedure when calculating the Earth Mover's Distance. The two classes are jets produced from boosted $W$ boson decays and generic quark and gluon jets.

\begin{figure}[h!]
\centering
\includegraphics[width=0.5\textwidth]{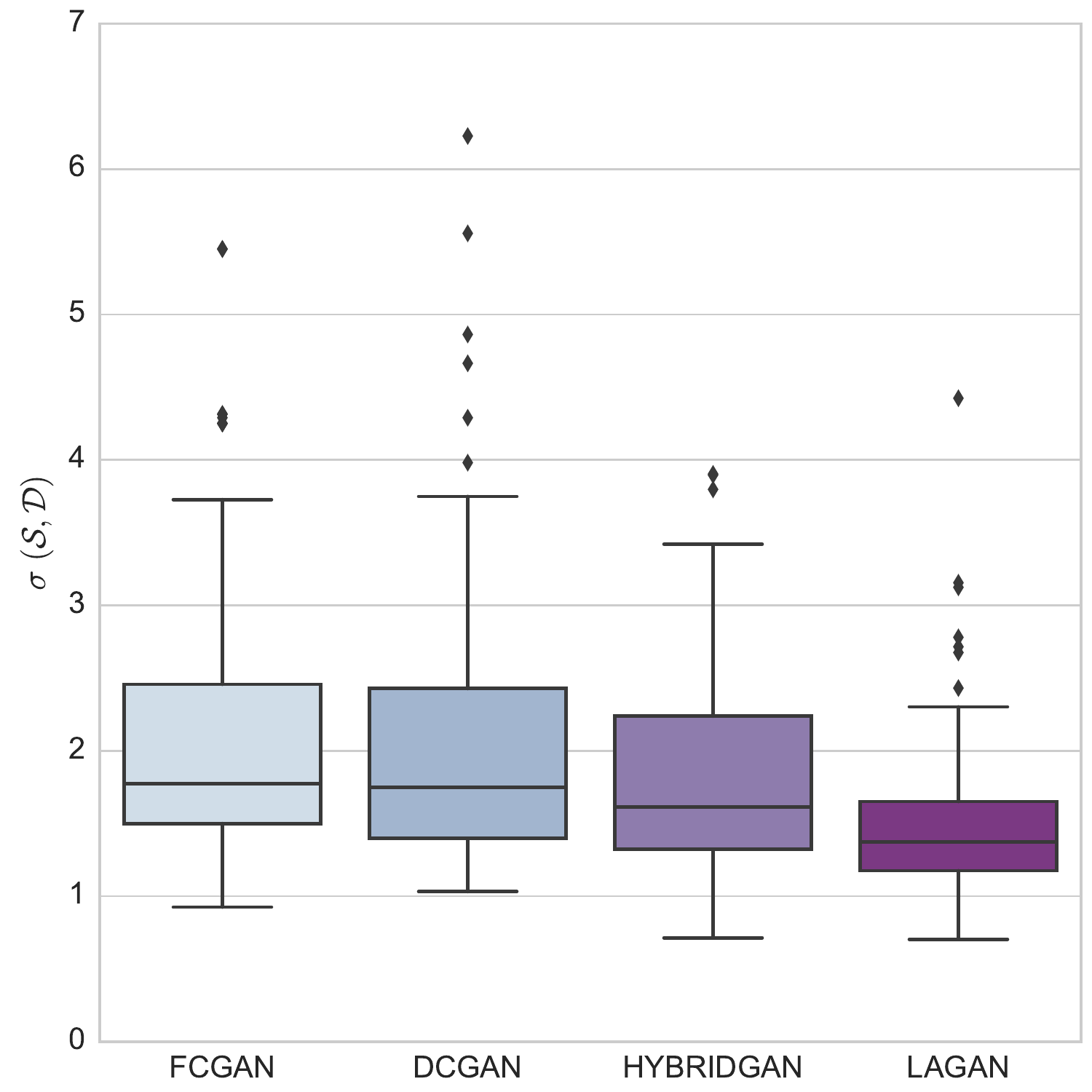}
\caption{Boxplots quantifying the performance of the 4 bench-marked  network  architectures  in  terms  of  the  scoring  function $\sigma$ defined in Eq. 4.  The boxes represent the interquartile range (IQR),  and  the  line  inside  the  box  represents  the  media. The whiskers extend to include all points that fall withing $1.5\times\text{IQR}$ of the upper and lower quartiles.}
\label{fig:boxplot}
\end{figure}

The LAGAN architecture performance is compared with other GAN architectures using Eq.~\ref{eqn:metric} in Fig.~\ref{fig:boxplot}.  In addition to the DCGAN, we compare to two other networks introduced in Sec.~\ref{sec:arch}: an architecture that uses fully connected instead of locally connected layers (FCGAN) and an architecture that combines a fully connected stream and a convolutional stream (HYBRIDGAN).  Figure~\ref{fig:boxplot} shows  that  the  LAGAN  architecture  achieves  the  lowest median distance score, as well as the lowest spread between the first and third quartiles, confirming it as both a powerful and robust architecture for generating jet images.

\subsection{Inference Time Observations}

\label{ssec:inference}

\begin{figure}[h!]
\centering
\includegraphics[width=0.5\textwidth]{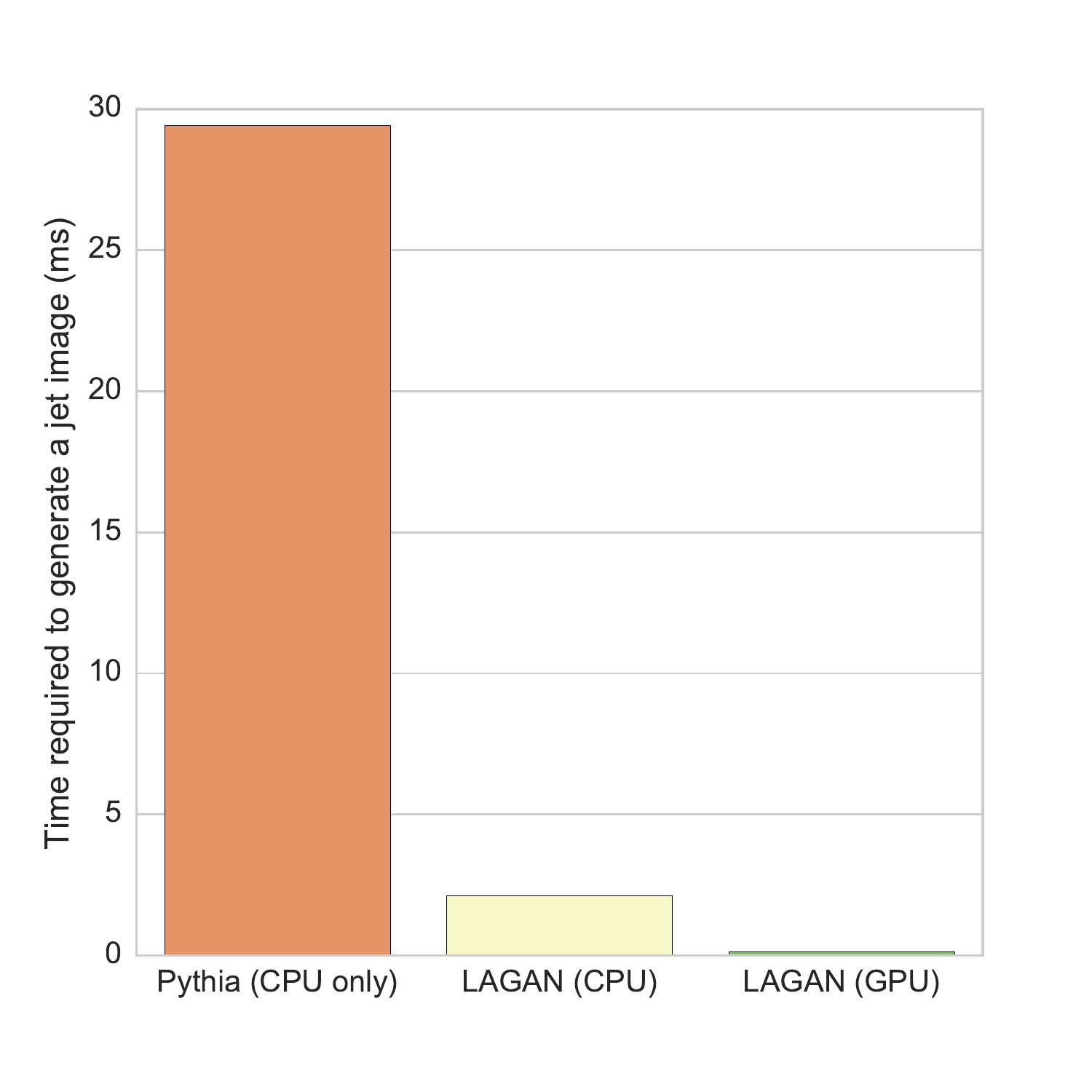}
\caption{Time required to produce a jet image in PYTHIA-basedgeneration, LAGAN-based generation on the CPU, and LAGAN-based generation on the GPU.}
\label{fig:timing}
\end{figure}

To evaluate the relative benefit of a GAN-powered fast simulation regime, we benchmark the speed of event generation using Pythia on a CPU versus our finalized GAN model on both CPU and GPU, as many applications of such technology in High Energy Physics will not have access to inference-time graphics processors. We benchmark on a \texttt{p2.xlarge} Elastic Compute Cloud (EC2) instance on Amazon Web Services (AWS) for control and reproducibility, using an Intel\textsuperscript{\textregistered} Xeon\textsuperscript{\textregistered} \texttt{E5-2686 v4 @ 2.30GHz} for all CPU tests, and an NVIDIA\textsuperscript{\textregistered} Tesla K80 for GPU tests. We use TensorFlow v0.12, CUDA 8.0, cuDNN v5.1 for all GPU-driven generation. GAN-based approaches can offer up to two orders of magnitude improvement over traditional event generation (see Table~\ref{speedup} and Fig.~\ref{fig:timing}), validating this approach as an avenue with pursuing.

\begin{table}[h!]
\centering
\caption{Performance Comparison for LAGAN and Pythia-driven event generation}
\label{speedup}
\begin{tabular}{l|lll}
\textbf{Method} & \textbf{Hardware} & \textbf{\# events / sec} & \textbf{Relative Speed-up} \\ \hline\hline
Pythia          & CPU               & 34                    & 1                        \\ \hline
LAGAN           & CPU               & 470                   & 14                      \\ \hline
LAGAN           & GPU               & 7200                  & 210                    
\end{tabular}
\end{table}

\section{Conclusions and Outlook}
\label{sec:conclusion}



We have introduced a new neural network architecture, the Location-Aware Generative Adversarial Network (LAGAN), to learn a rich generative model of jet images. Both qualitative and quantitative evaluations show that this model produces realistic looking images. To the best of our knowledge, this represents the one of the first successful applications of Generative Adversarial Networks to the physical sciences. Jet images differ significantly from natural images studied in the extensive GAN literature and the LAGAN represents a significant step forward in extending the GAN idea to other domains.  

The LAGAN marks the beginning of a new era of fast and powerful generative models for high energy physics. Jet images have provided a useful testing ground to deploy state-of-the-art generation methods. Preliminary work that incorporates the LAGAN in an architecture for higher dimensional images demonstrates the importance of this work for future GAN-based simulation ~\cite{Paganini:2017hrr}.
Neural network generation is a promising avenue for many aspects of the computationally heavy needs of high energy particle and nuclear physics.





\section*{Acknowledgements}
\label{sec:ack}

This work was supported in part by the Office of High Energy Physics of the U.S. Department of Energy under contracts DE-AC02-05CH11231 and DE-FG02-92ER40704. The authors would like to thank Ian Goodfellow for insightful deep learning related discussion, and would like to acknowledge Wahid Bhimji, Zach Marshall, Mustafa Mustafa, Chase Shimmin, and Paul Tipton, who helped refine our narrative.

\clearpage

\appendix

\section{Additional Material}
\label{app:auxmaterial}

\begin{figure}[h!]
\centering
\includegraphics[width=0.333\textwidth]{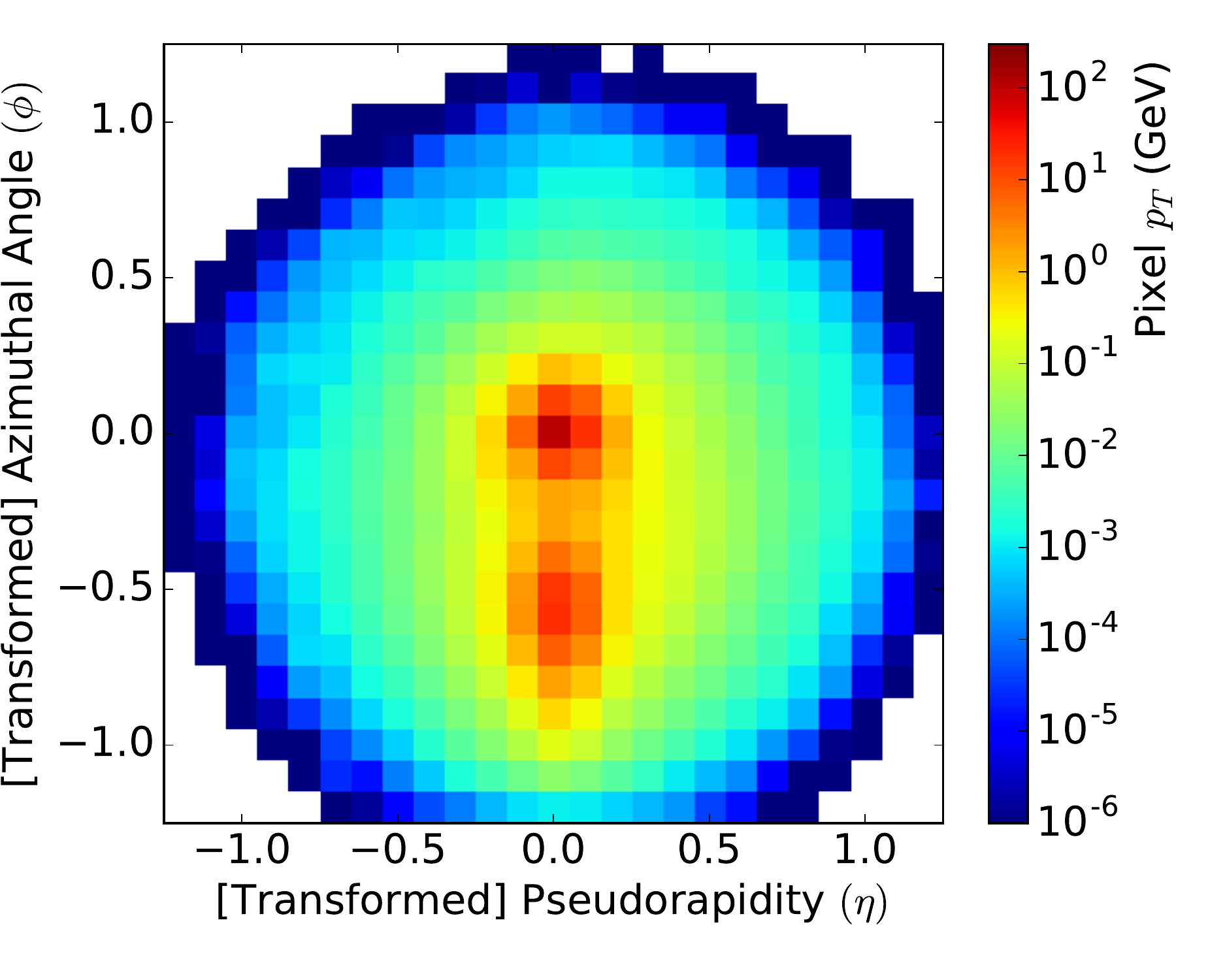}\includegraphics[width=0.333\textwidth]{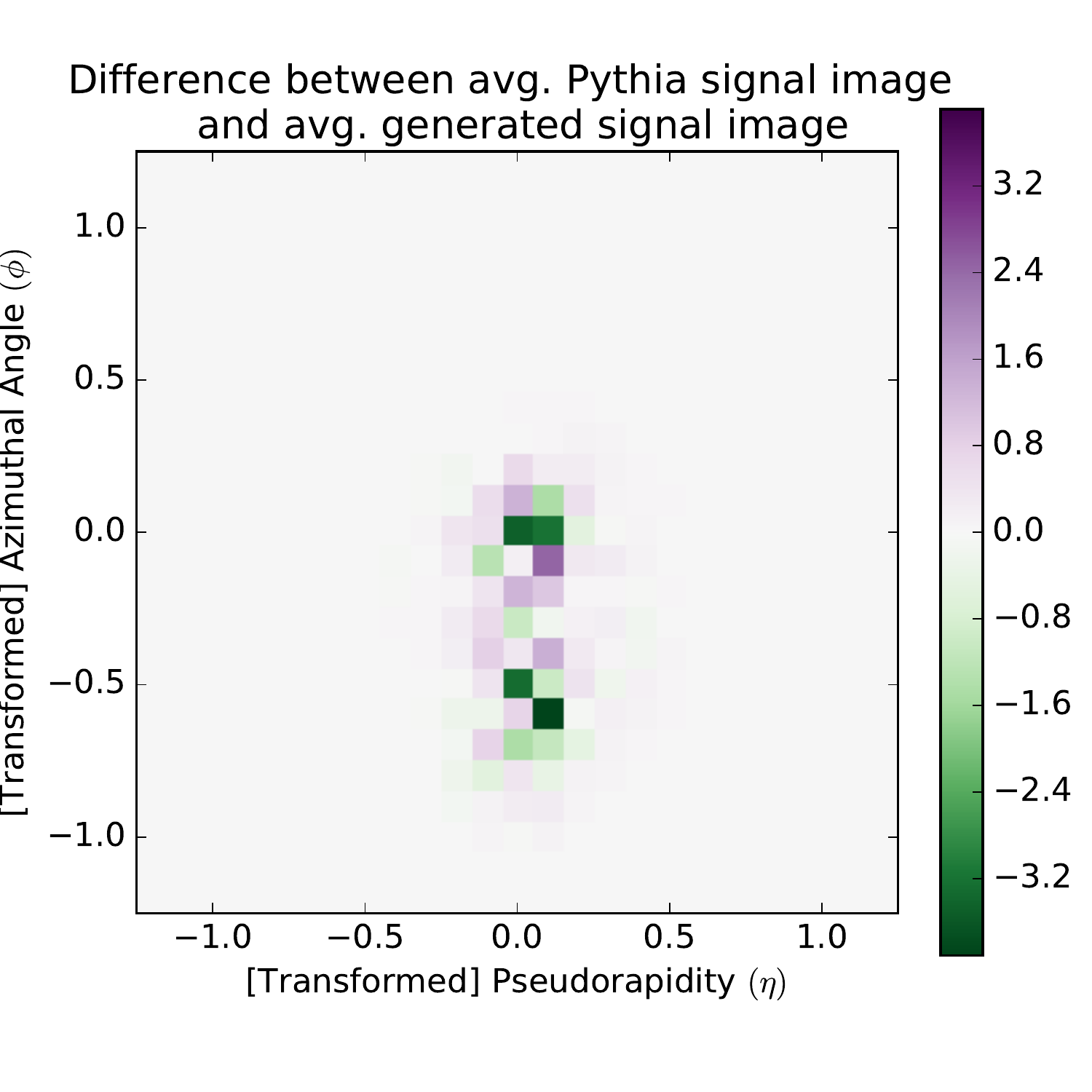}\includegraphics[width=0.333\textwidth]{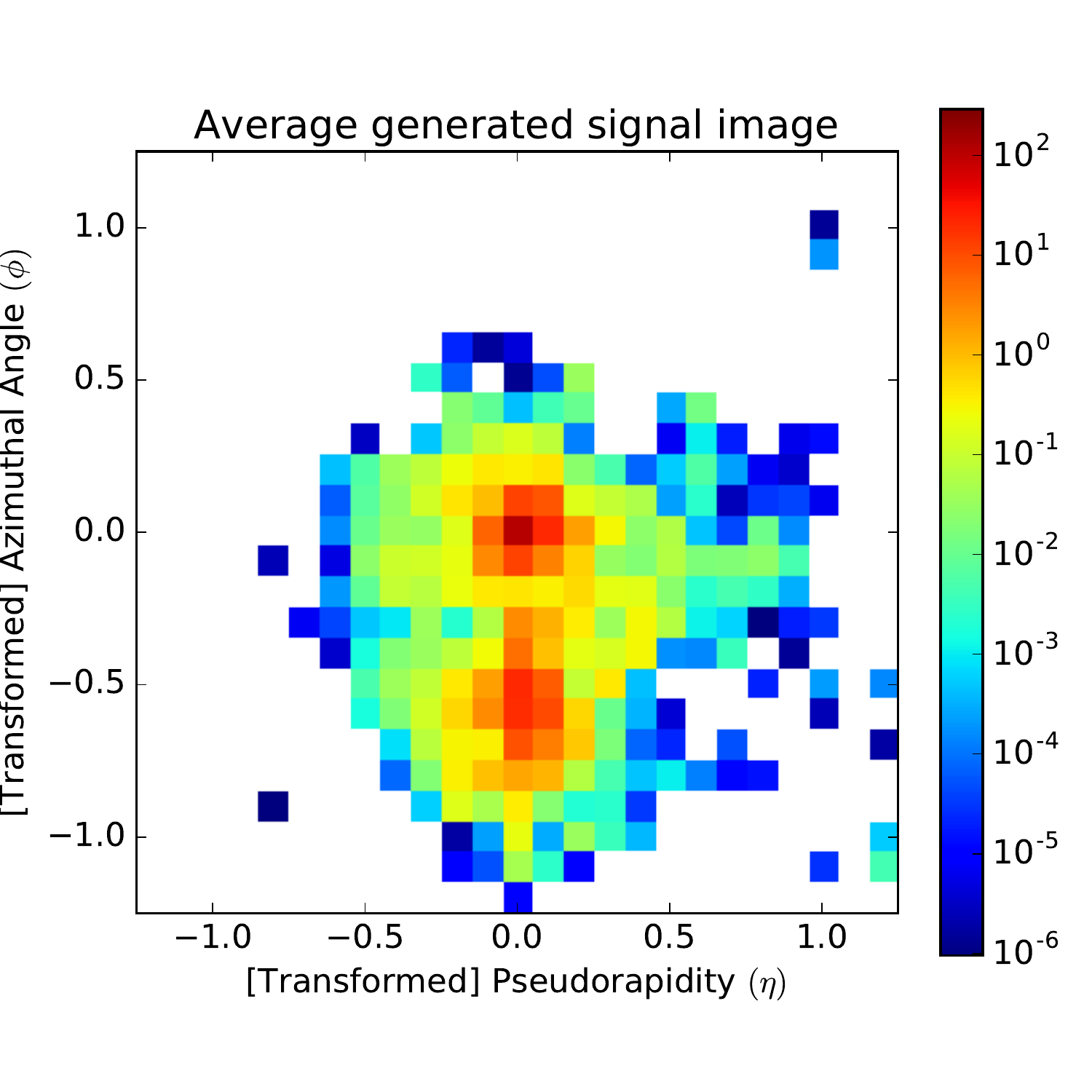}
\caption{Average signal image produced by Pythia (left) and by the GAN (right), displayed on log scale, with the difference between the two (center), displayed on linear scale.}
\label{fig:signal_pythia-gan}
\end{figure}

\begin{figure}[h!]
\centering
\includegraphics[width=0.333\textwidth]{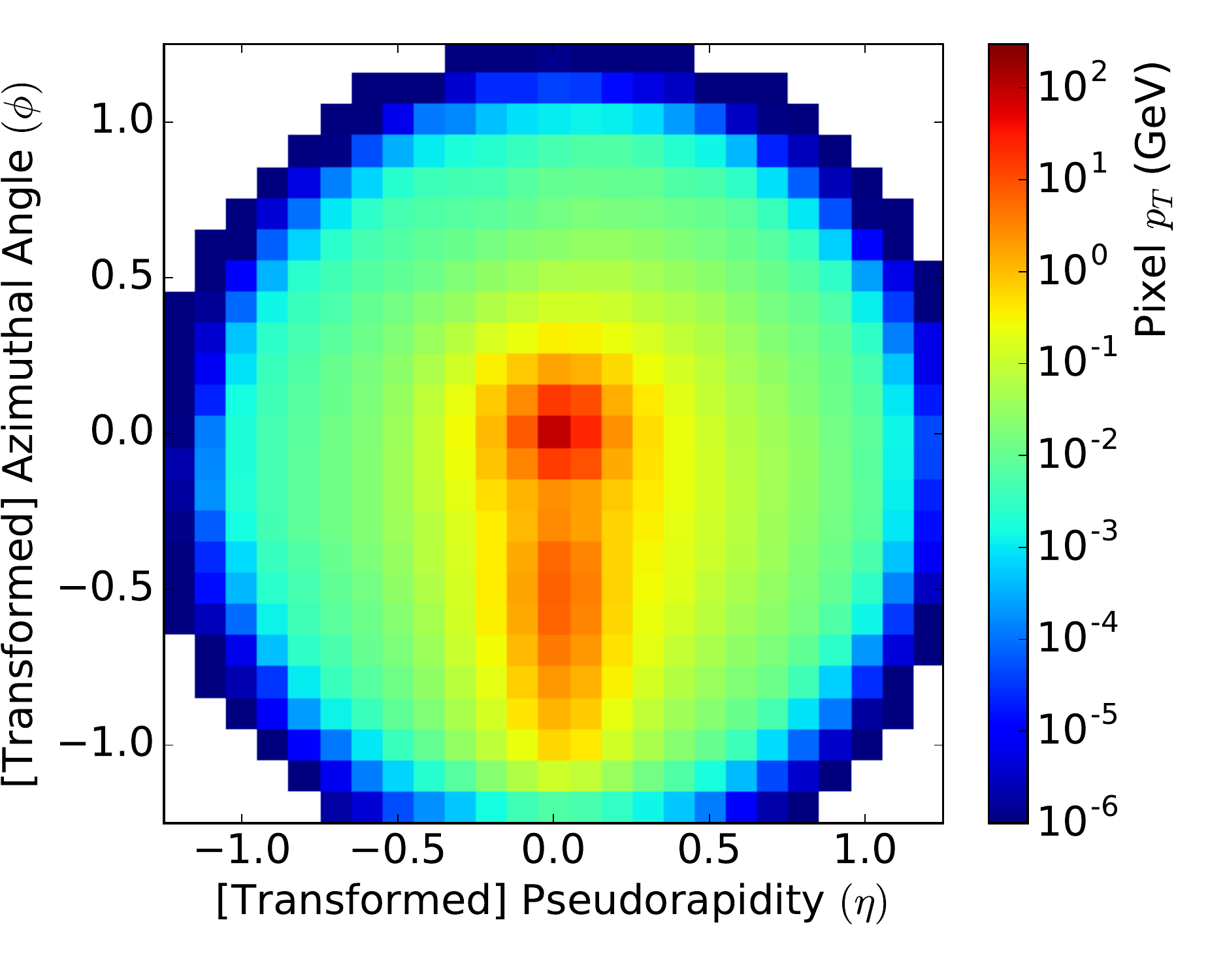}\includegraphics[width=0.333\textwidth]{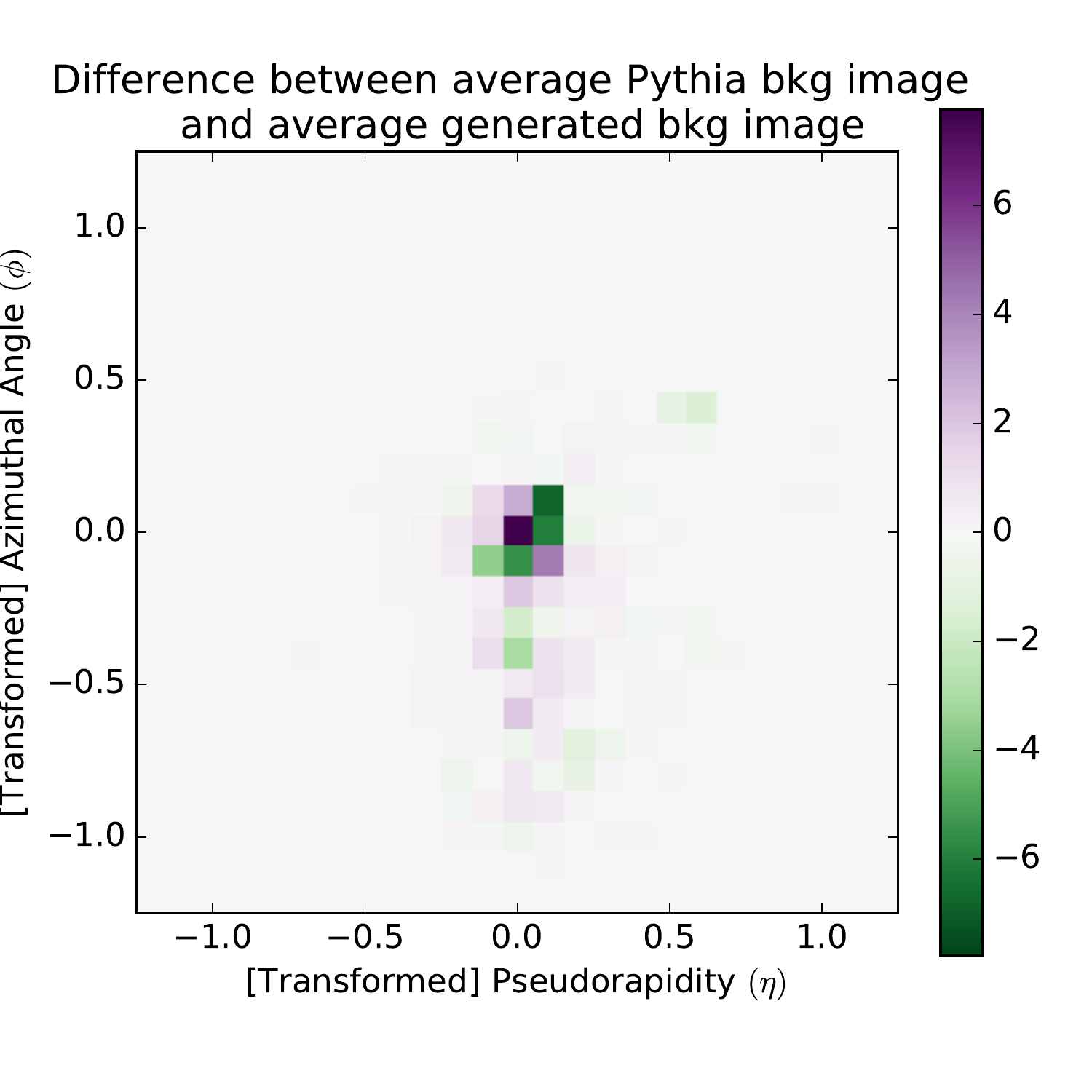}\includegraphics[width=0.333\textwidth]{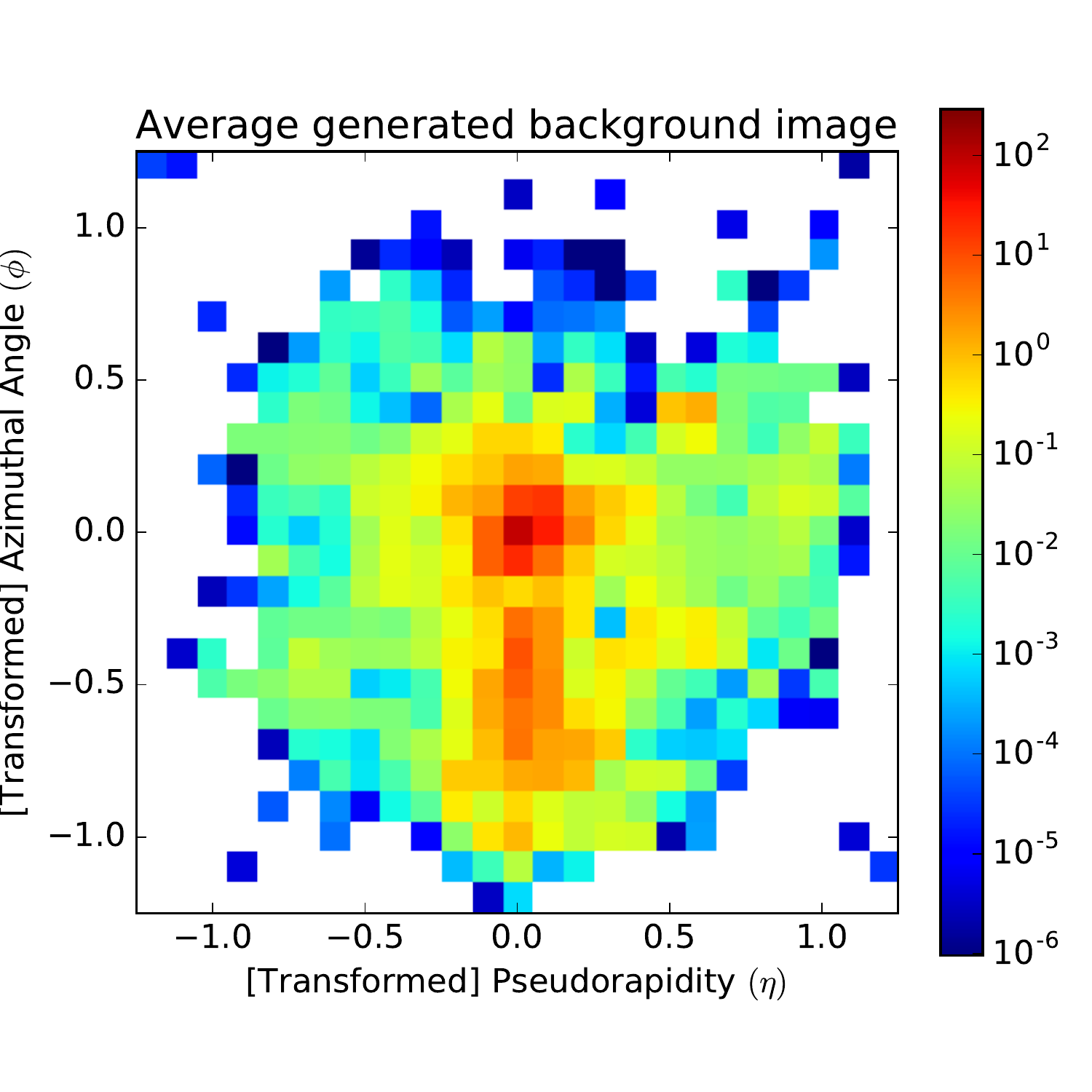}
\caption{Average background image produced by Pythia (left) and by the GAN (right), displayed on log scale, with the difference between the two (center), displayed on linear scale.}
\label{fig:bkg_pythia-gan}
\end{figure}

\begin{figure}[h!]
    \centering
    \includegraphics[width=0.333\textwidth]{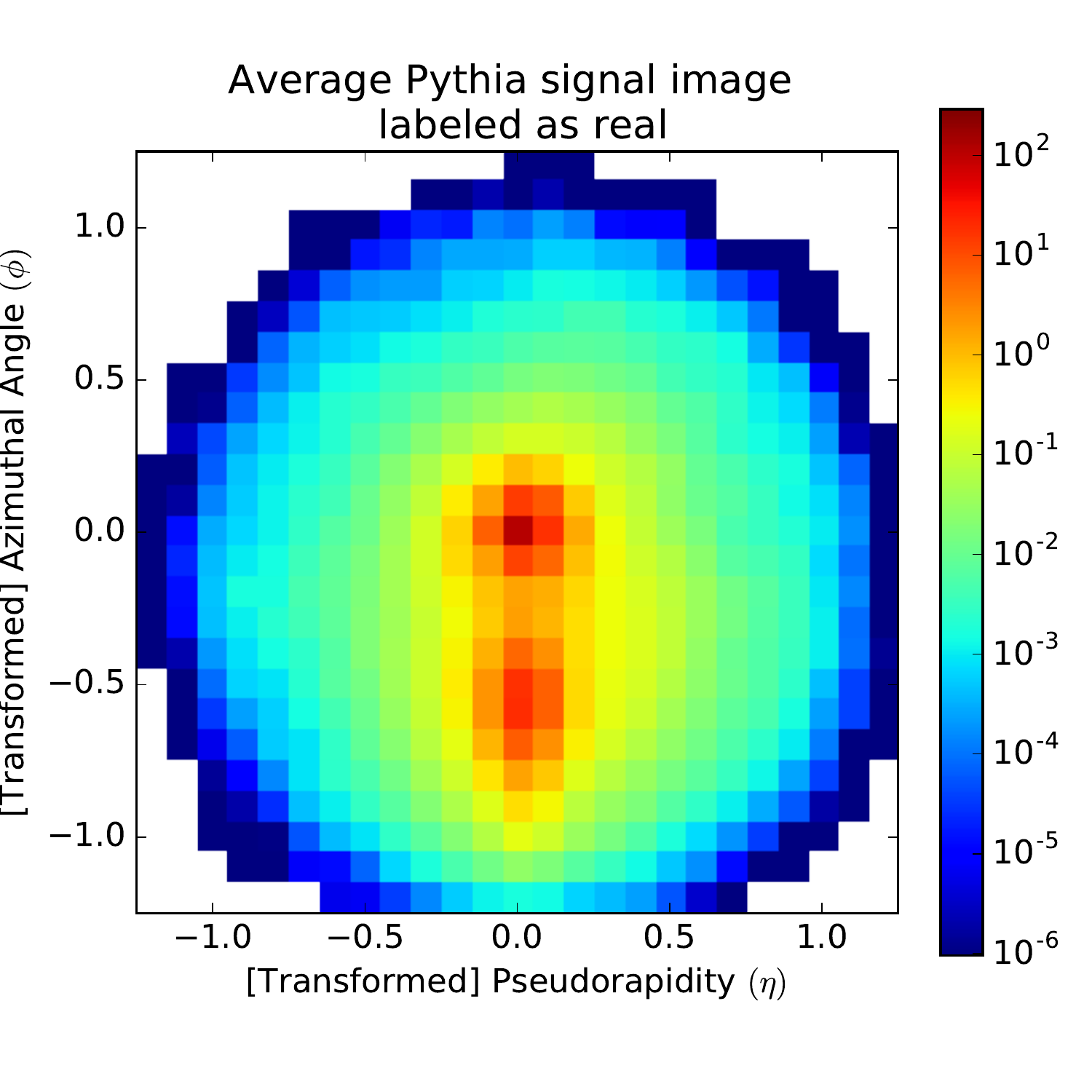}\includegraphics[width=0.333\textwidth]{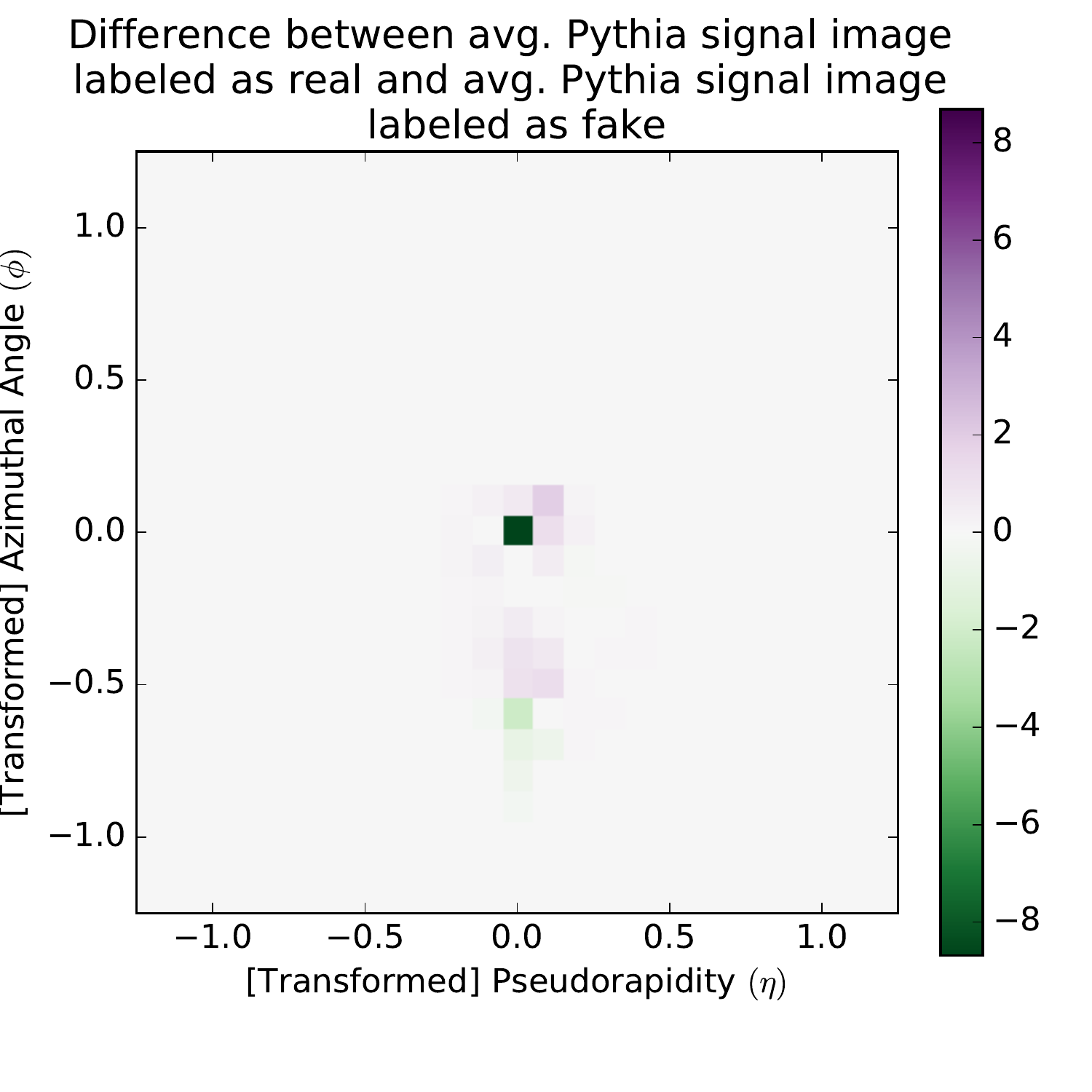}\includegraphics[width=0.333\textwidth]{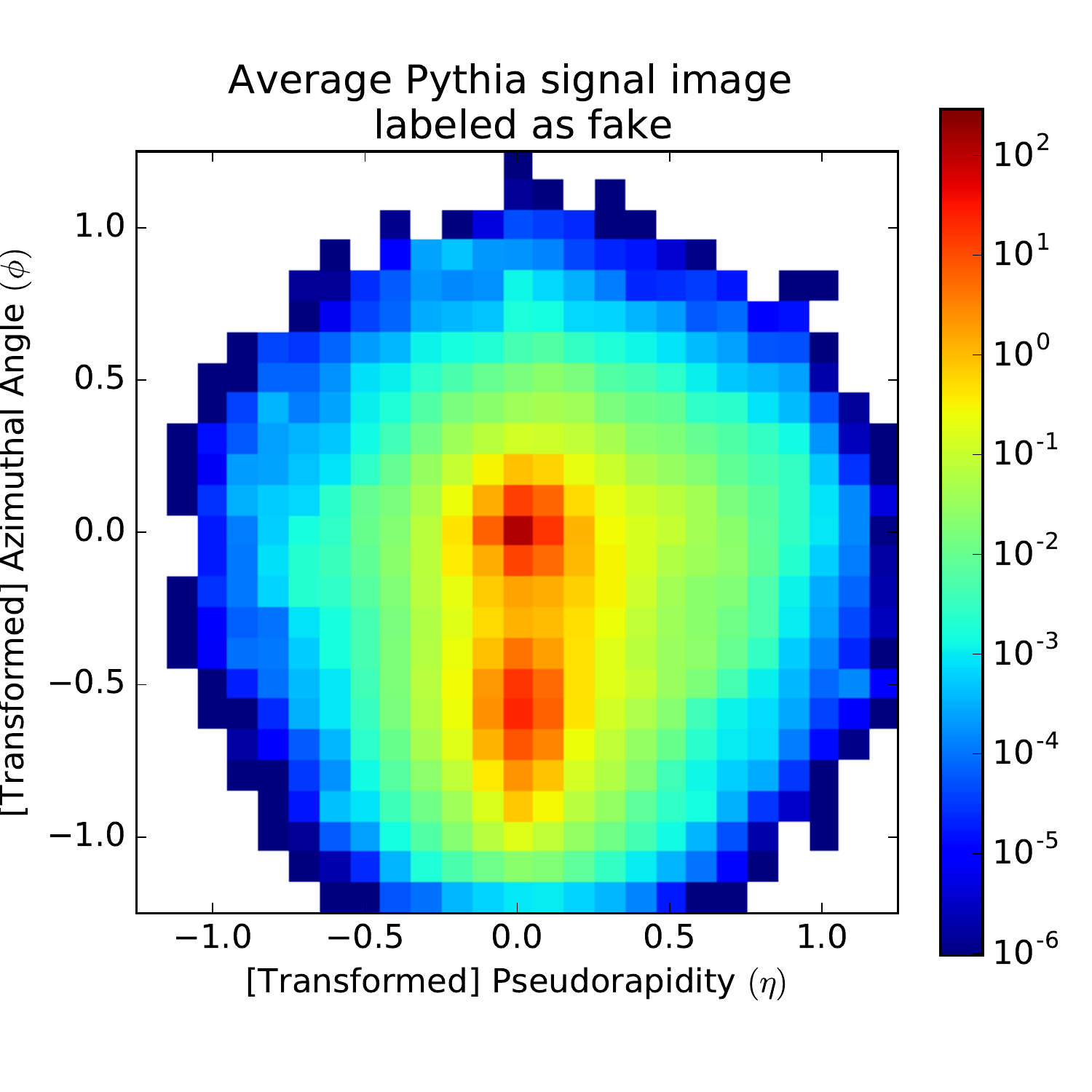}
    \caption{The average signal Pythia image labeled as real (left), as fake (right), and the difference between these two (middle) plotted on linear scale.}
    \label{fig:pythia_signal_real-fake}
\end{figure}

\begin{figure}[h!]
    \centering
    \includegraphics[width=0.333\textwidth]{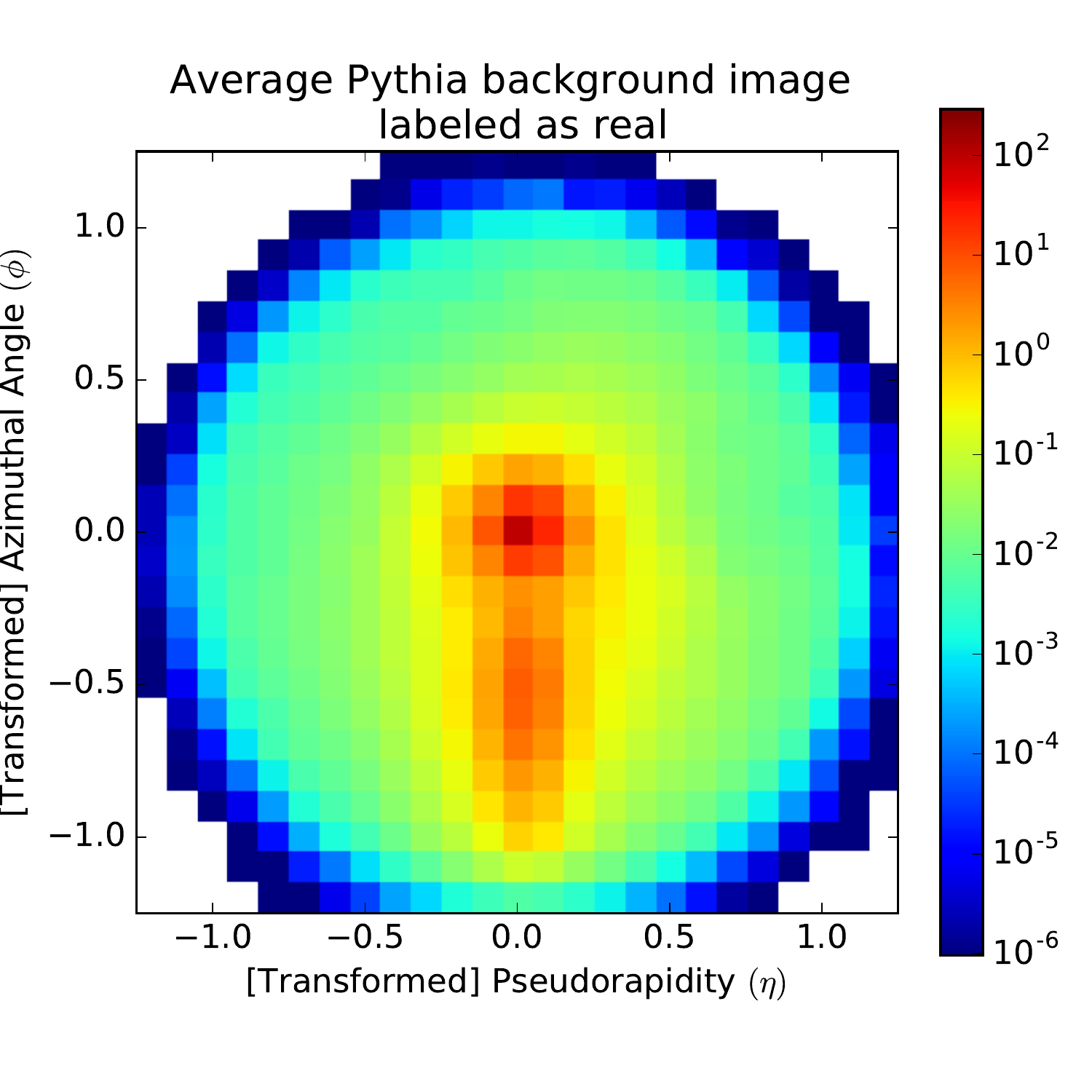}\includegraphics[width=0.333\textwidth]{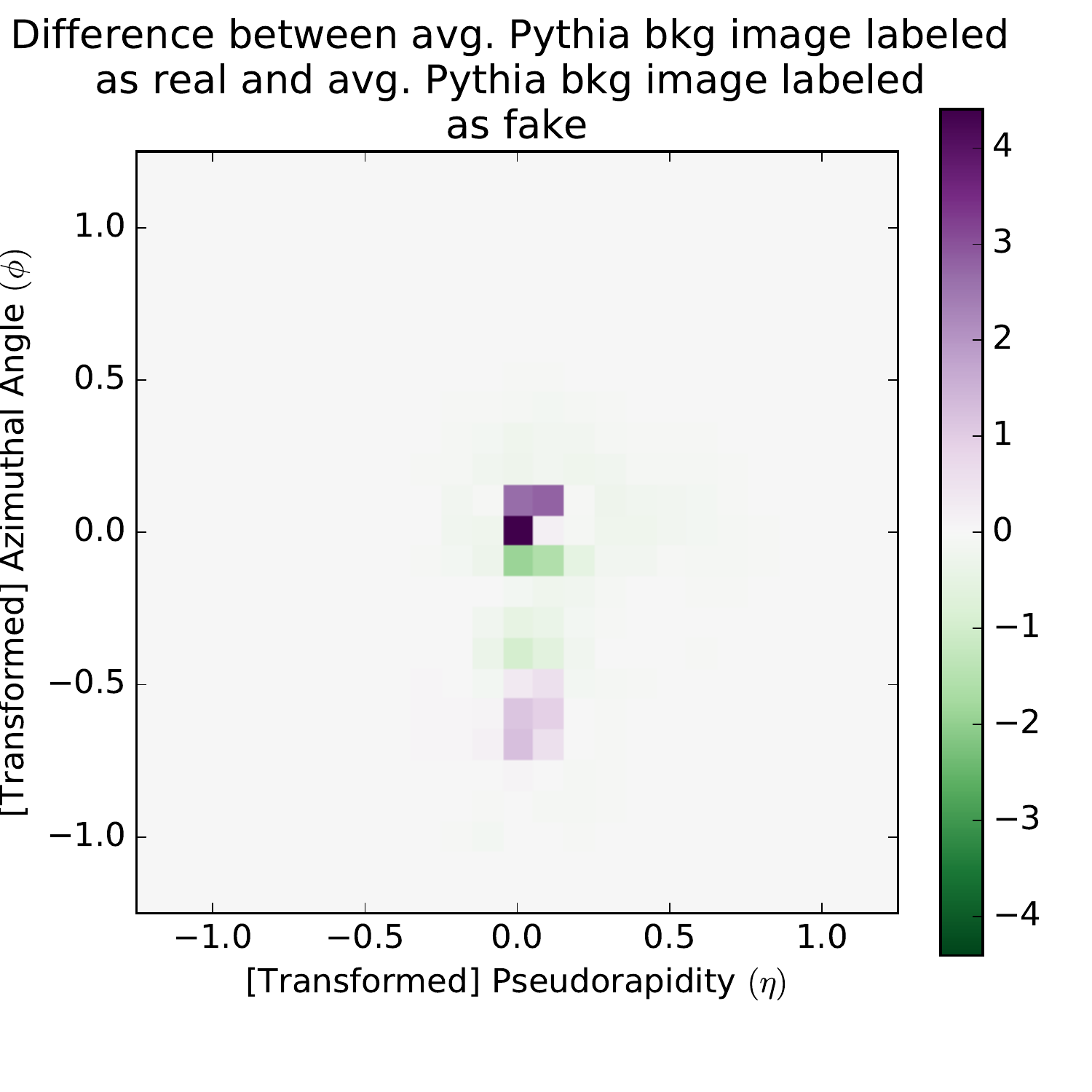}\includegraphics[width=0.333\textwidth]{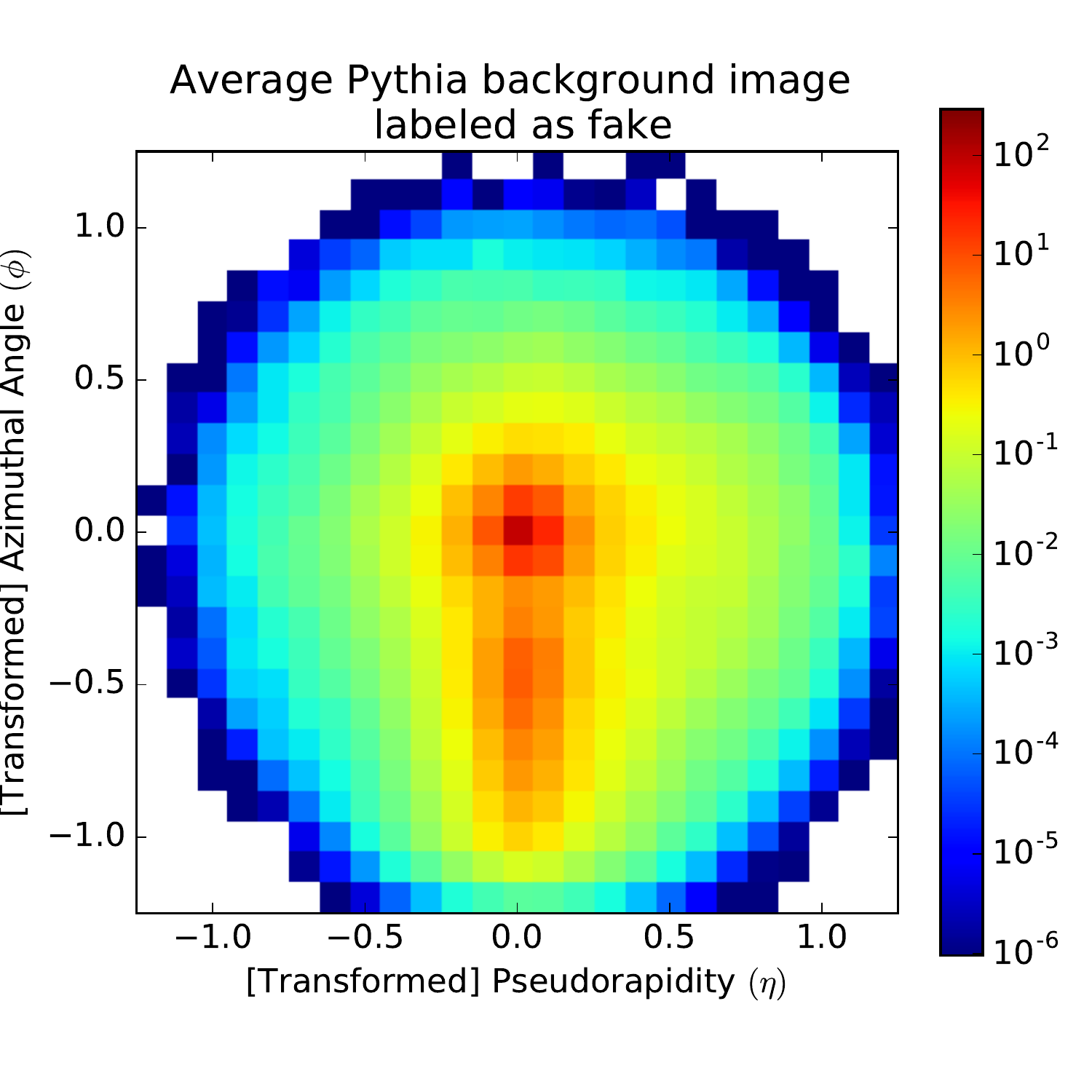}
    \caption{Average background Pythia image labeled as real (left), as fake (right), and the difference between these two (middle) plotted on linear scale.}
    \label{fig:pythia_bkg_real-fake}
\end{figure}

\begin{figure}[h!]
    \centering
    \includegraphics[width=0.333\textwidth]{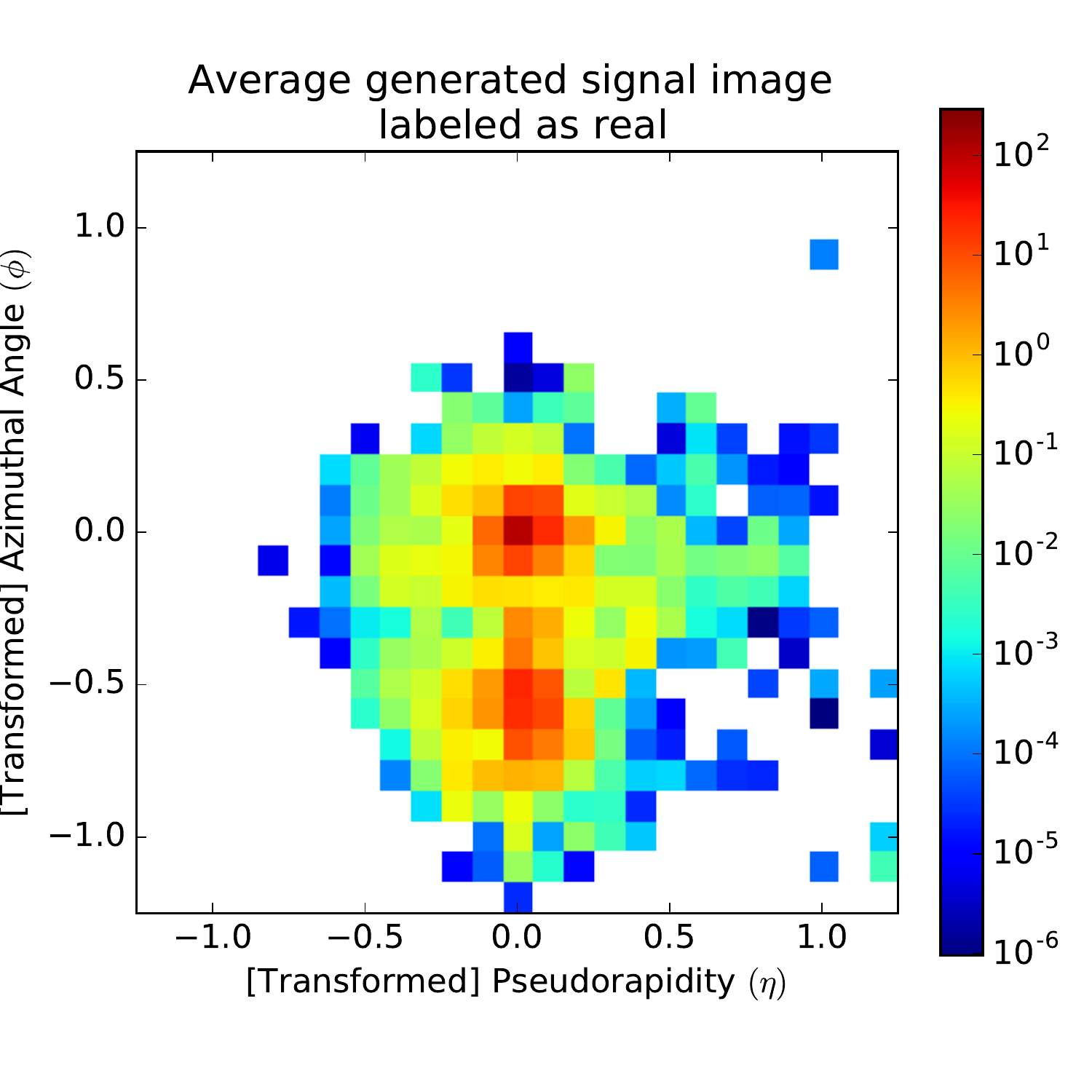}\includegraphics[width=0.333\textwidth]{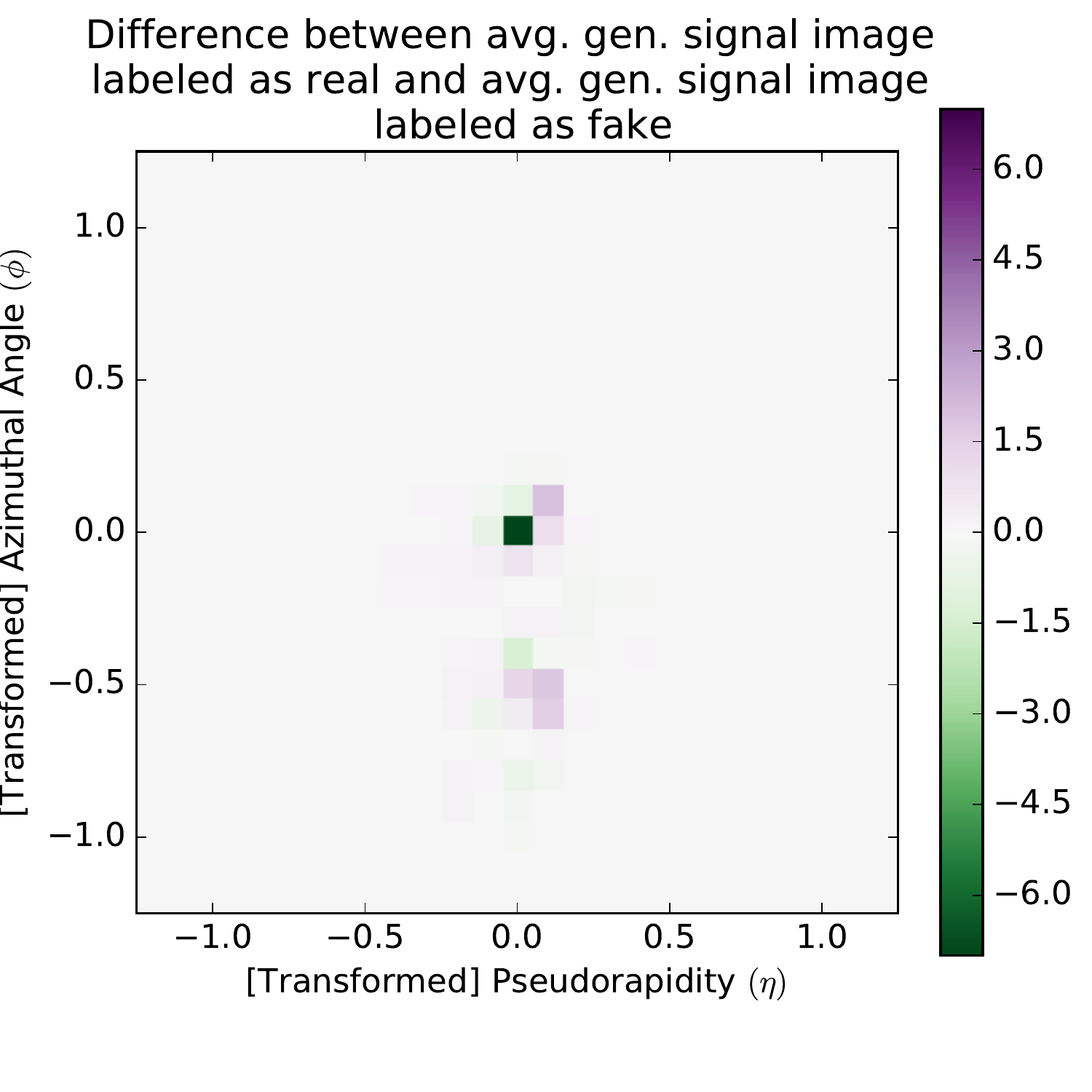}\includegraphics[width=0.333\textwidth]{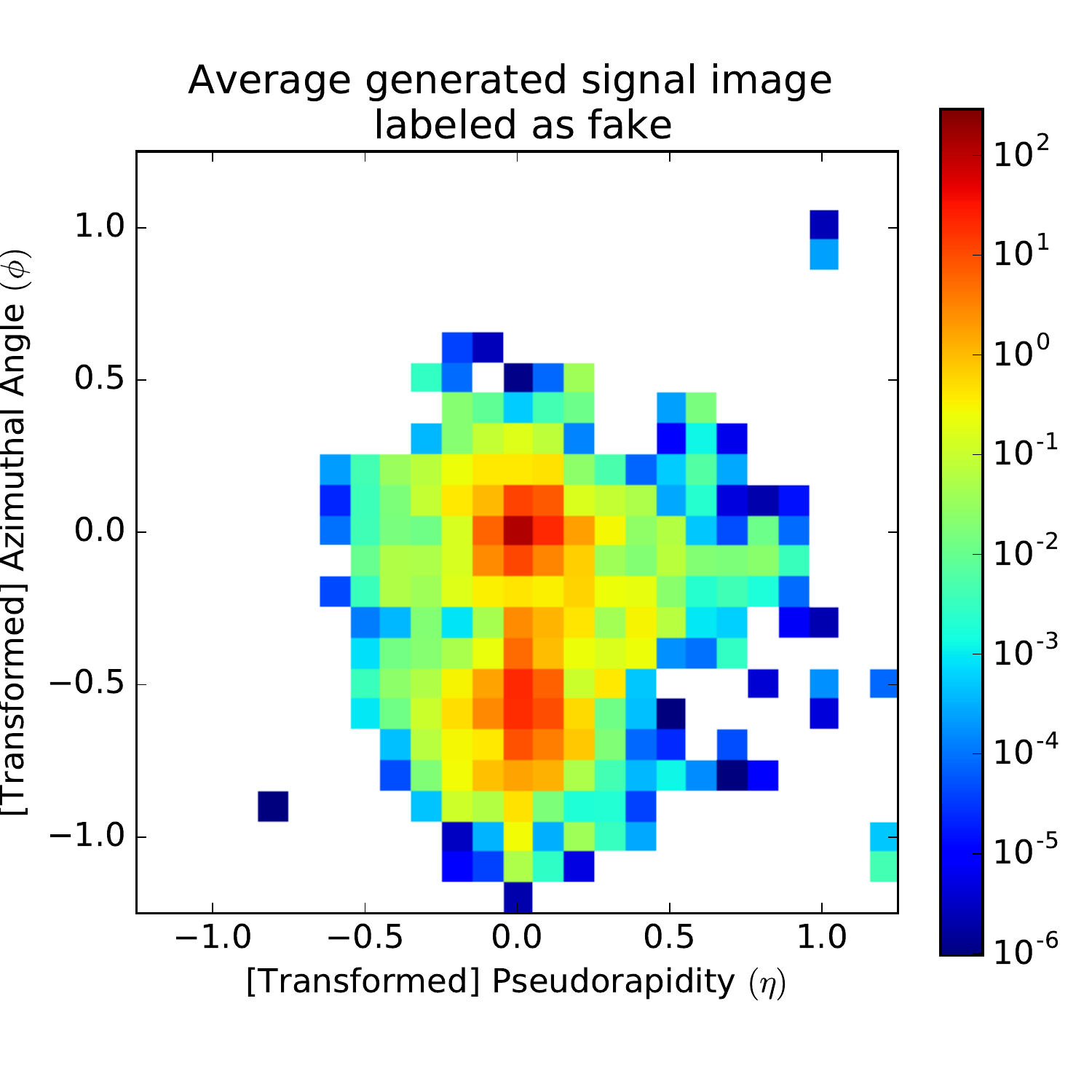}
    \caption{Average signal GAN-generated image labeled as real (left), as fake (right), and the difference between these two (middle) plotted on linear scale.}
    \label{fig:gan_signal_real-fake}
\end{figure}

\begin{figure}[h!]
    \centering
    \includegraphics[width=0.333\textwidth]{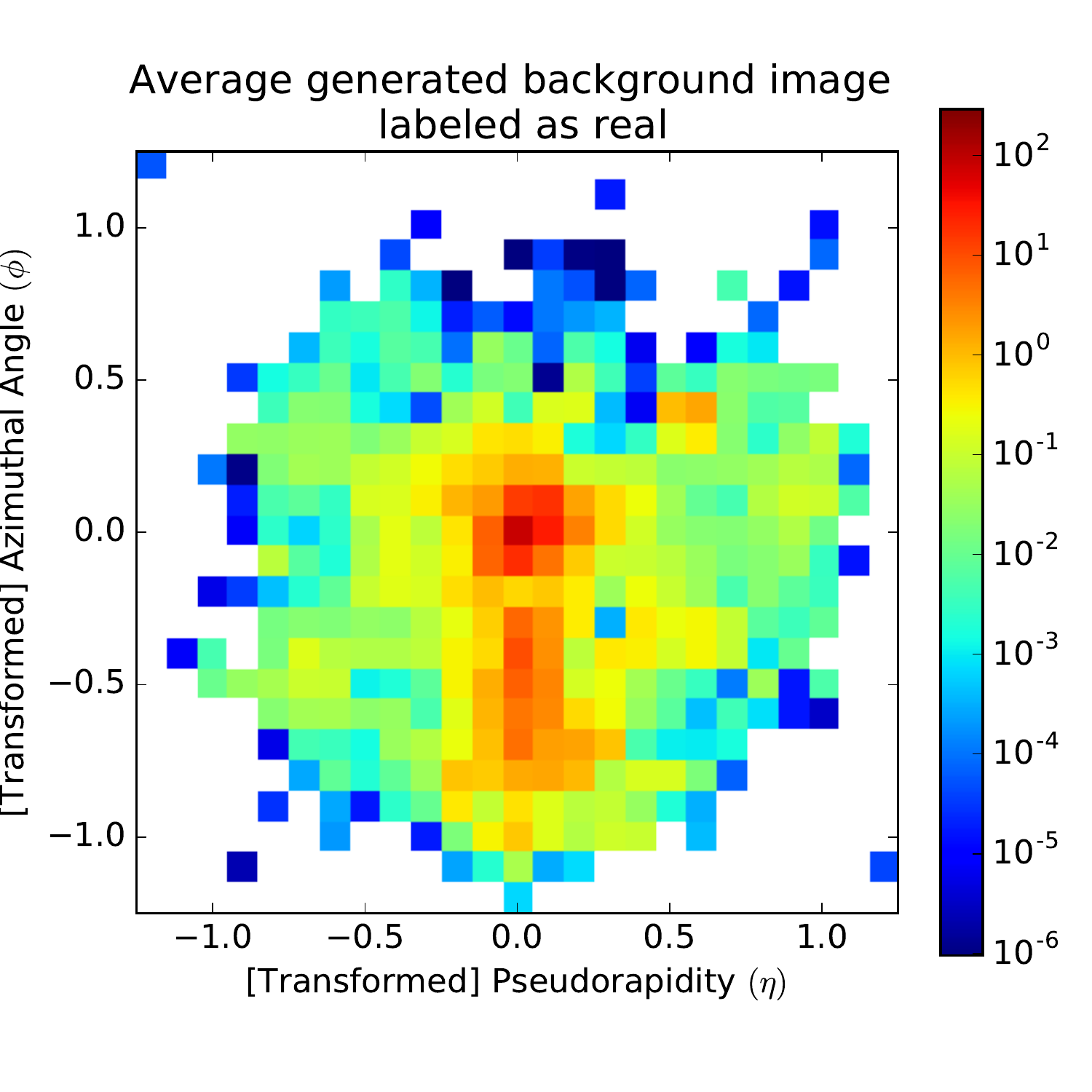}\includegraphics[width=0.333\textwidth]{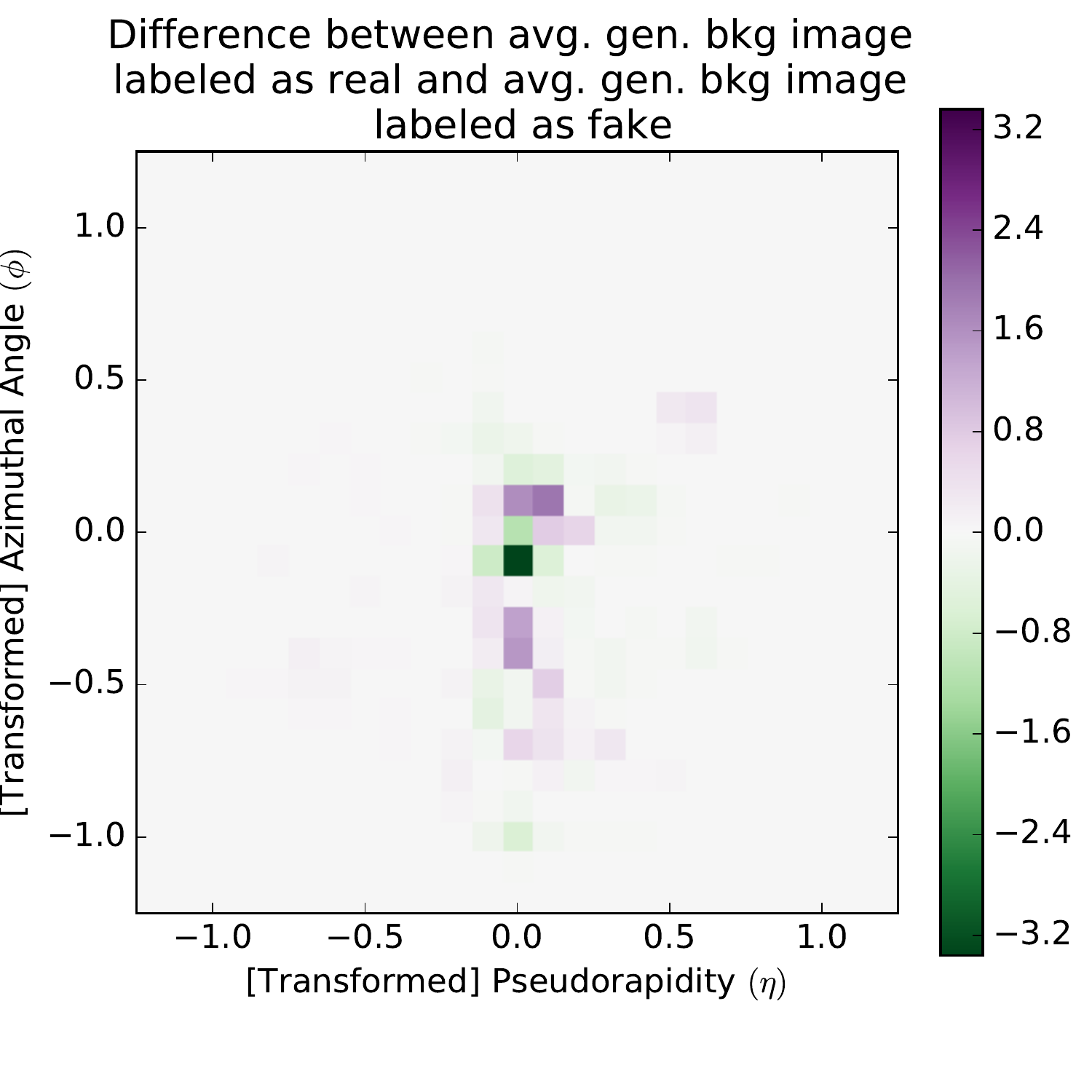}\includegraphics[width=0.333\textwidth]{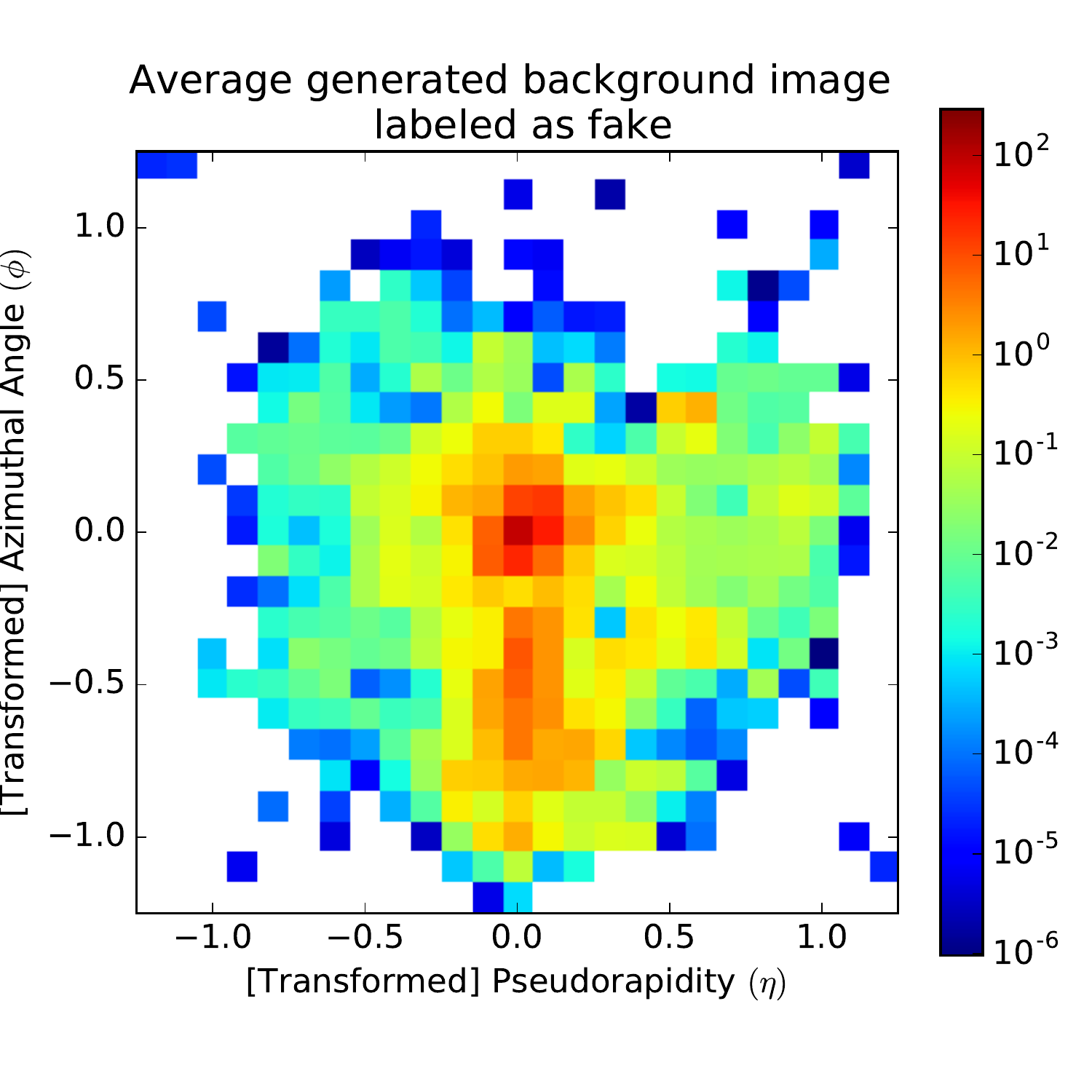}
    \caption{Average background GAN-generated image labeled as real (left), as fake (right), and the difference between these two (middle) plotted on linear scale.}
    \label{fig:gan_bkg_real-fake}
\end{figure}

\begin{figure}[h!]
\centering
\includegraphics[width=0.333\textwidth]{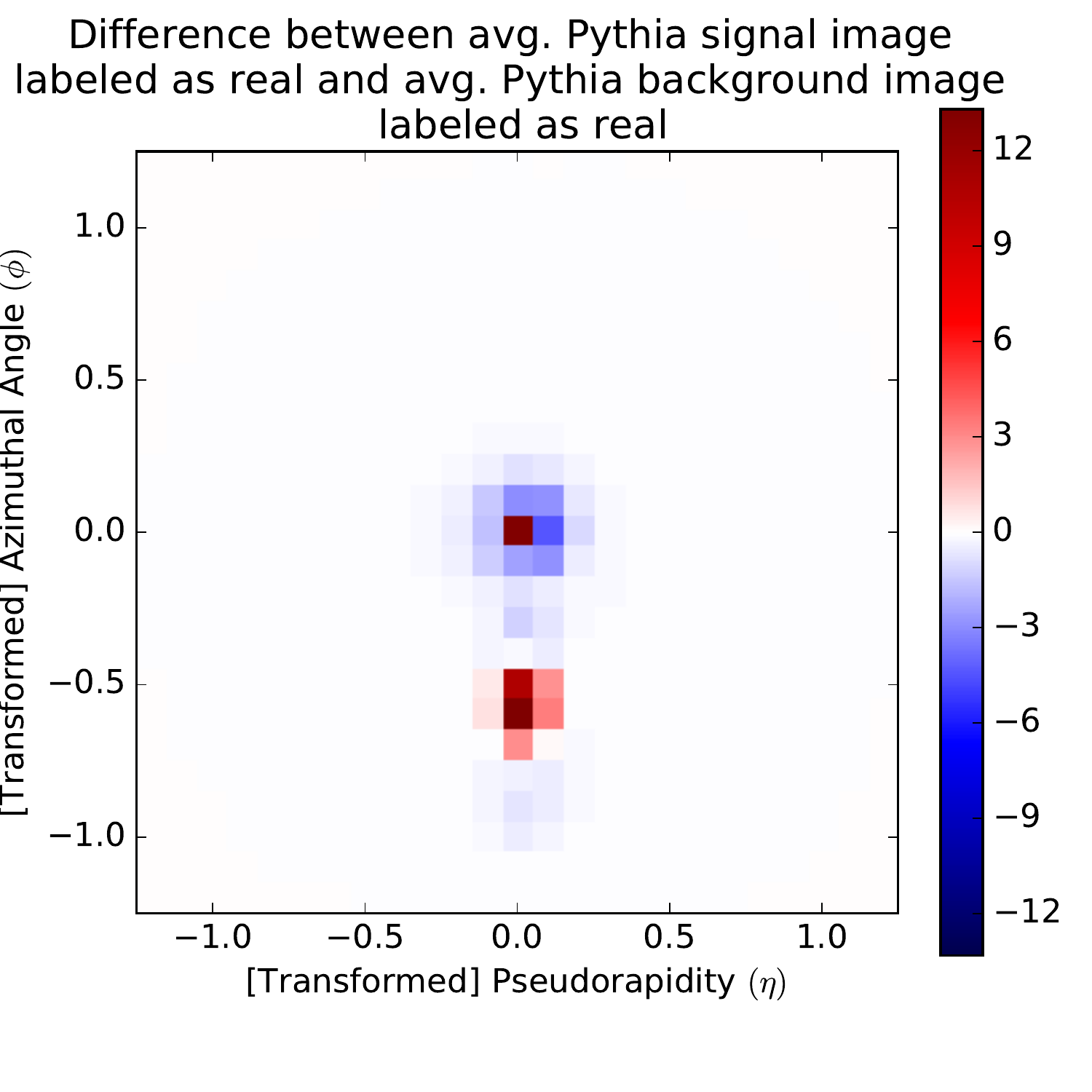}\hspace{5mm}\includegraphics[width=0.333\textwidth]{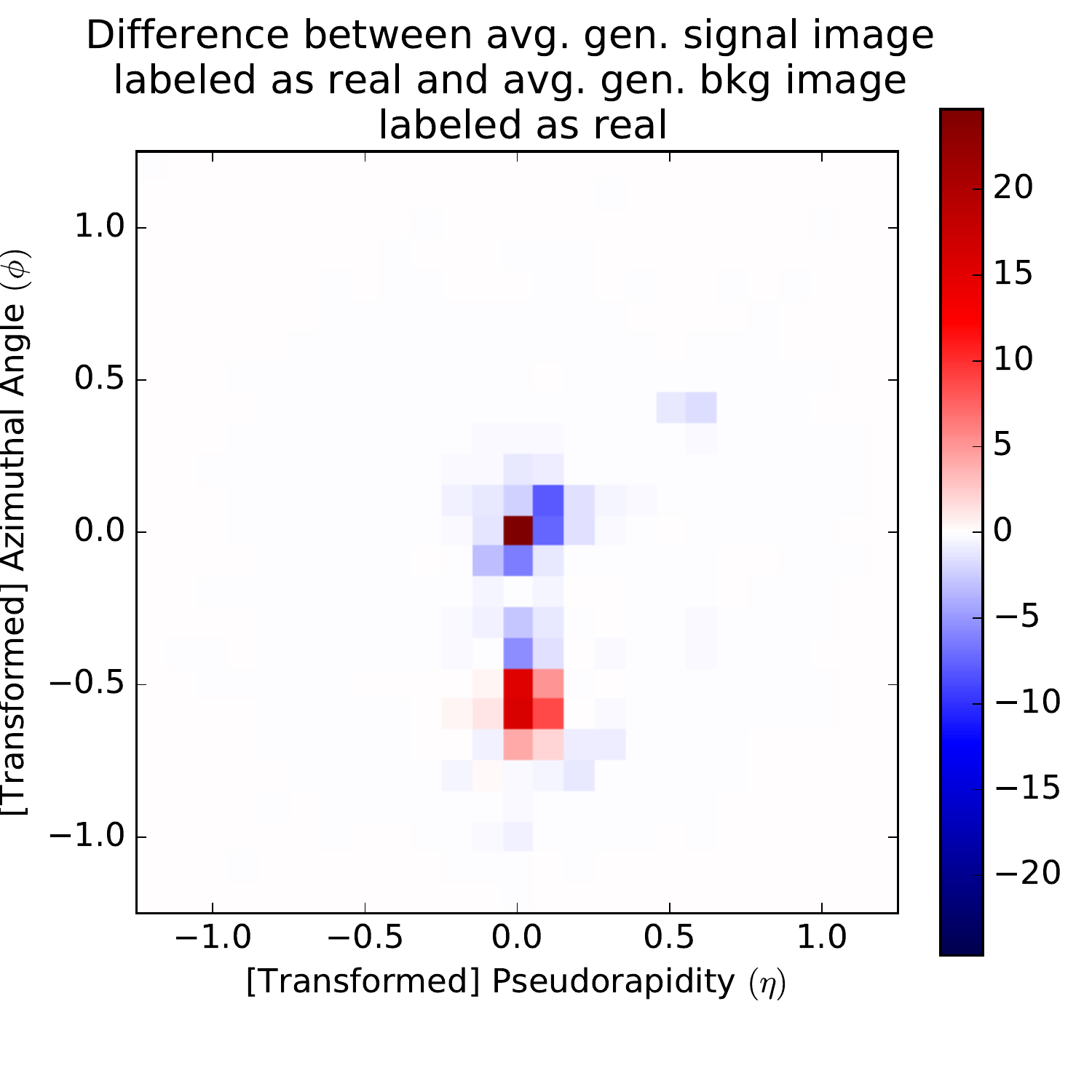}
\caption{Difference between the average signal and the average background images labeled as real, produced by Pythia (left) and by the GAN (right), both displayed on linear scale.}
\label{fig:real_signal-bkg}
\end{figure}

\begin{figure}[h!]
\centering
\includegraphics[width=0.333\textwidth]{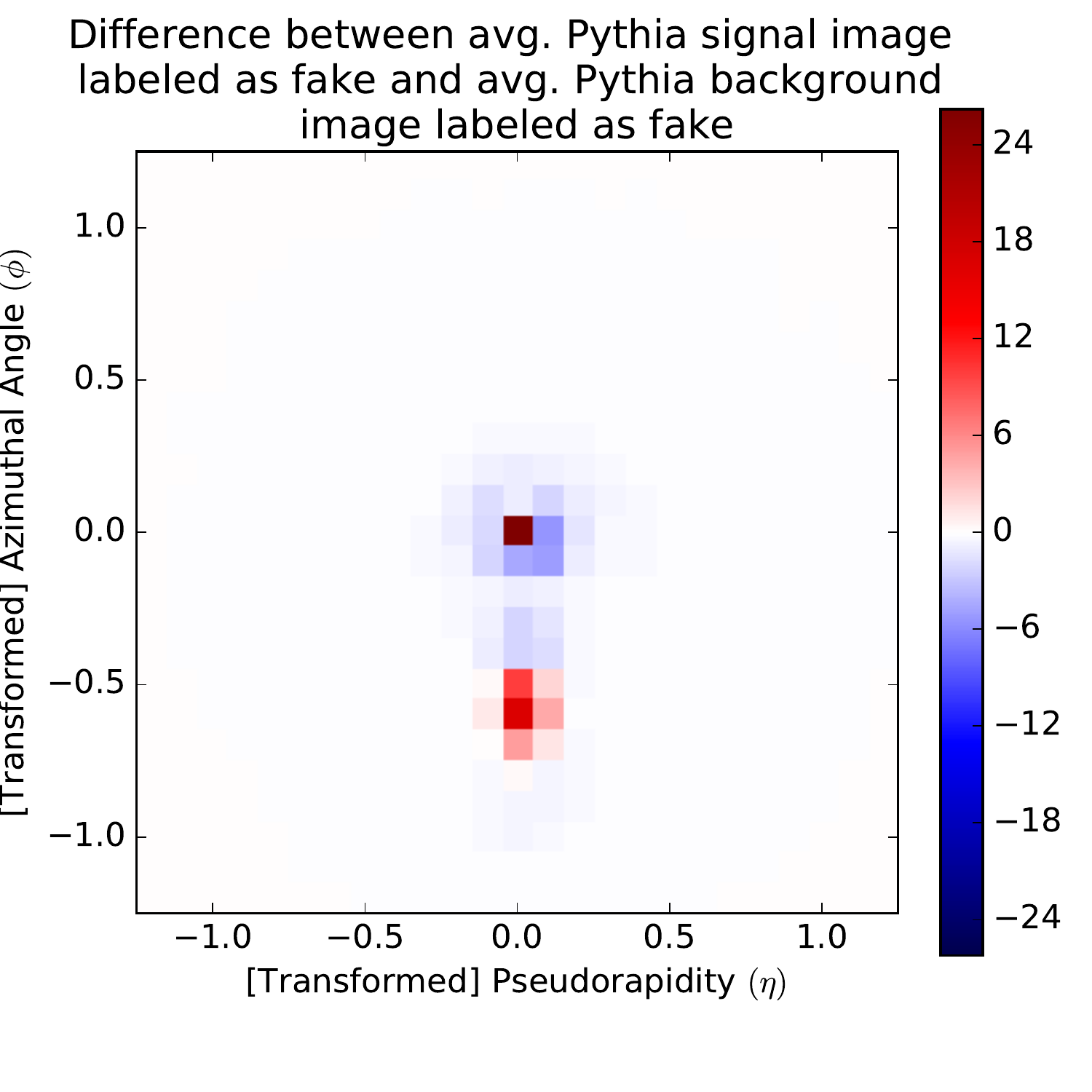}\hspace{5mm}\includegraphics[width=0.333\textwidth]{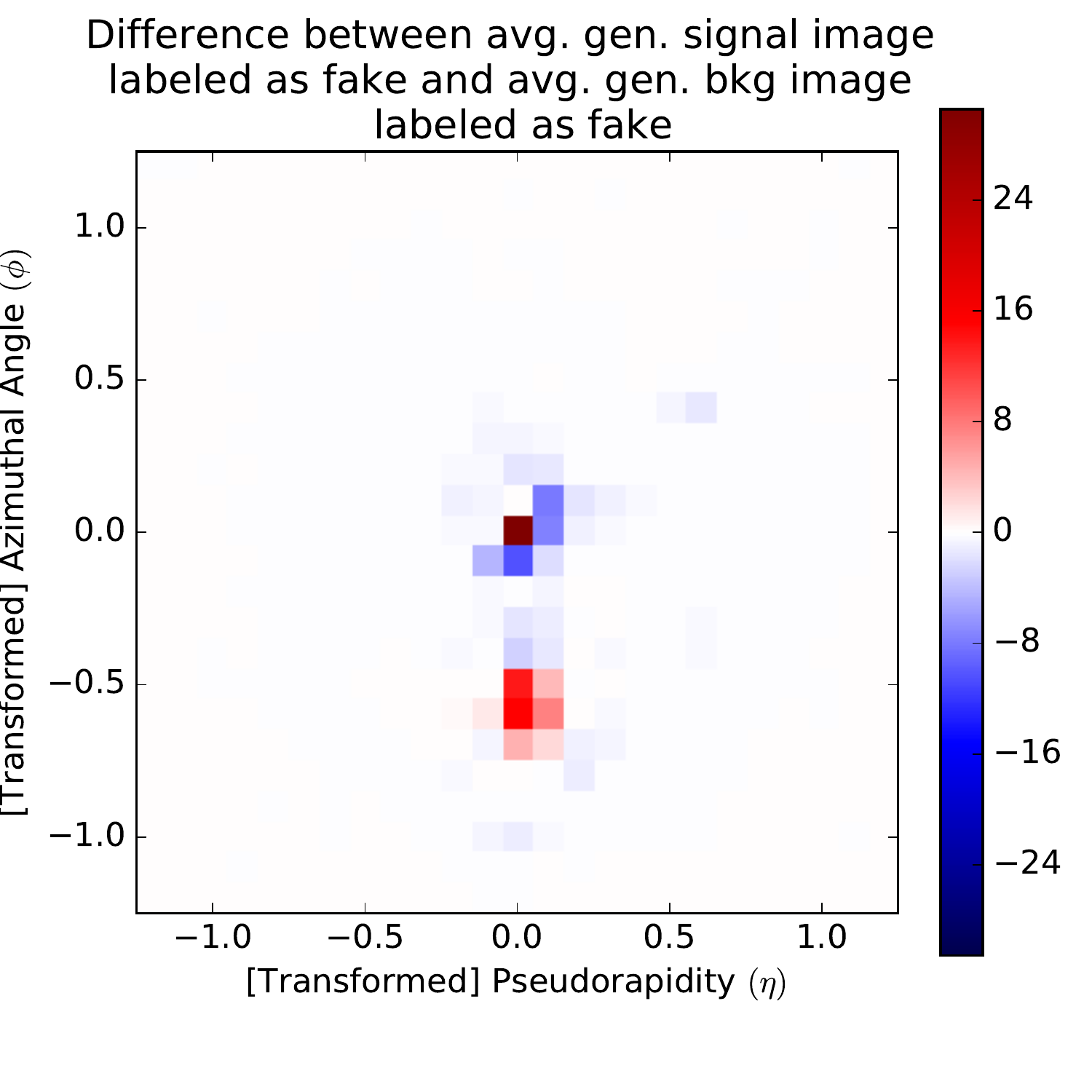}
\caption{Difference between the average signal and the average background images labeled as fake, produced by Pythia (left) and by the GAN (right), both displayed on linear scale.}
\label{fig:fake_signal-bkg}
\end{figure}

\begin{figure}[h!]
\centering
\includegraphics[width=0.3\textwidth]{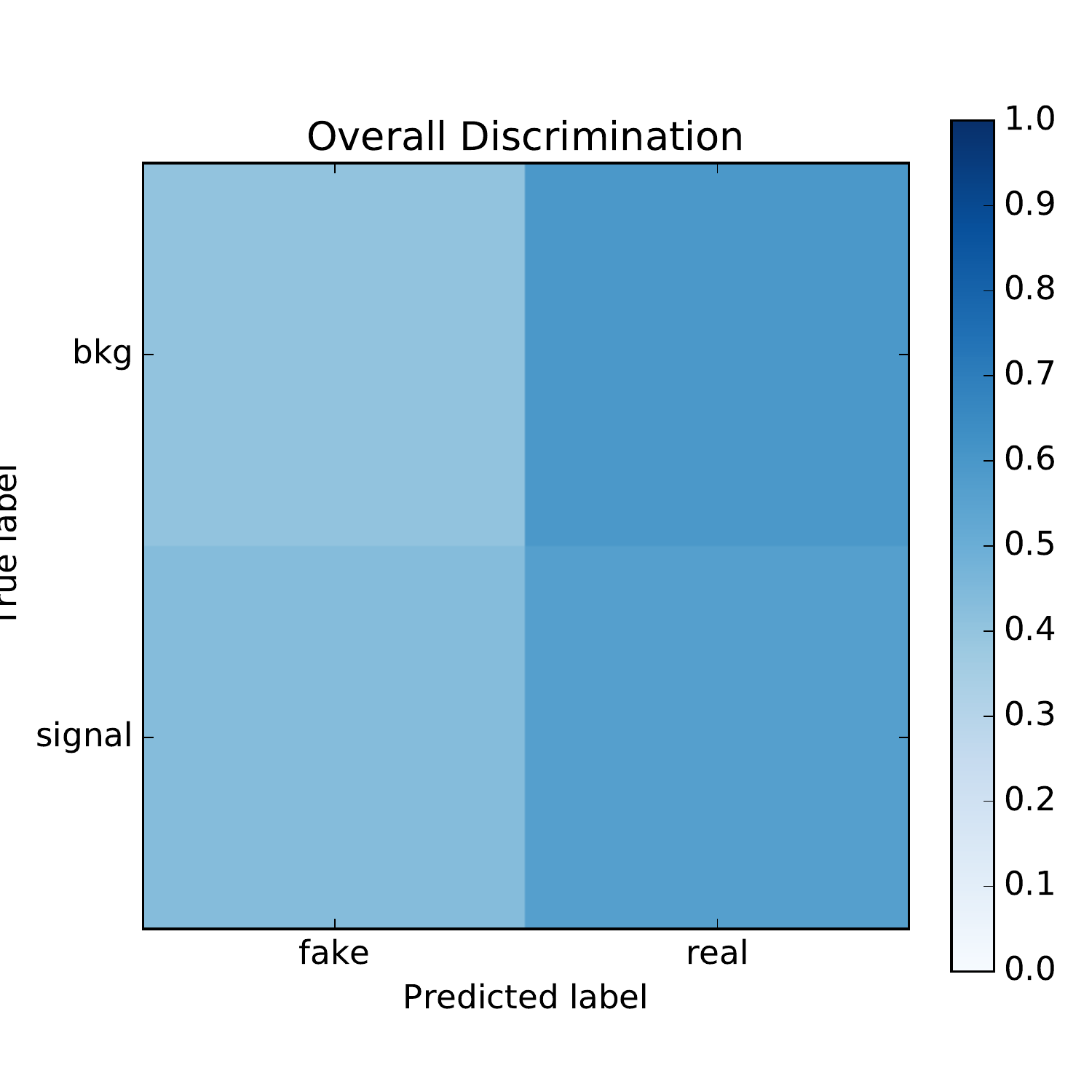}\hspace{2mm}\includegraphics[width=0.3\textwidth]{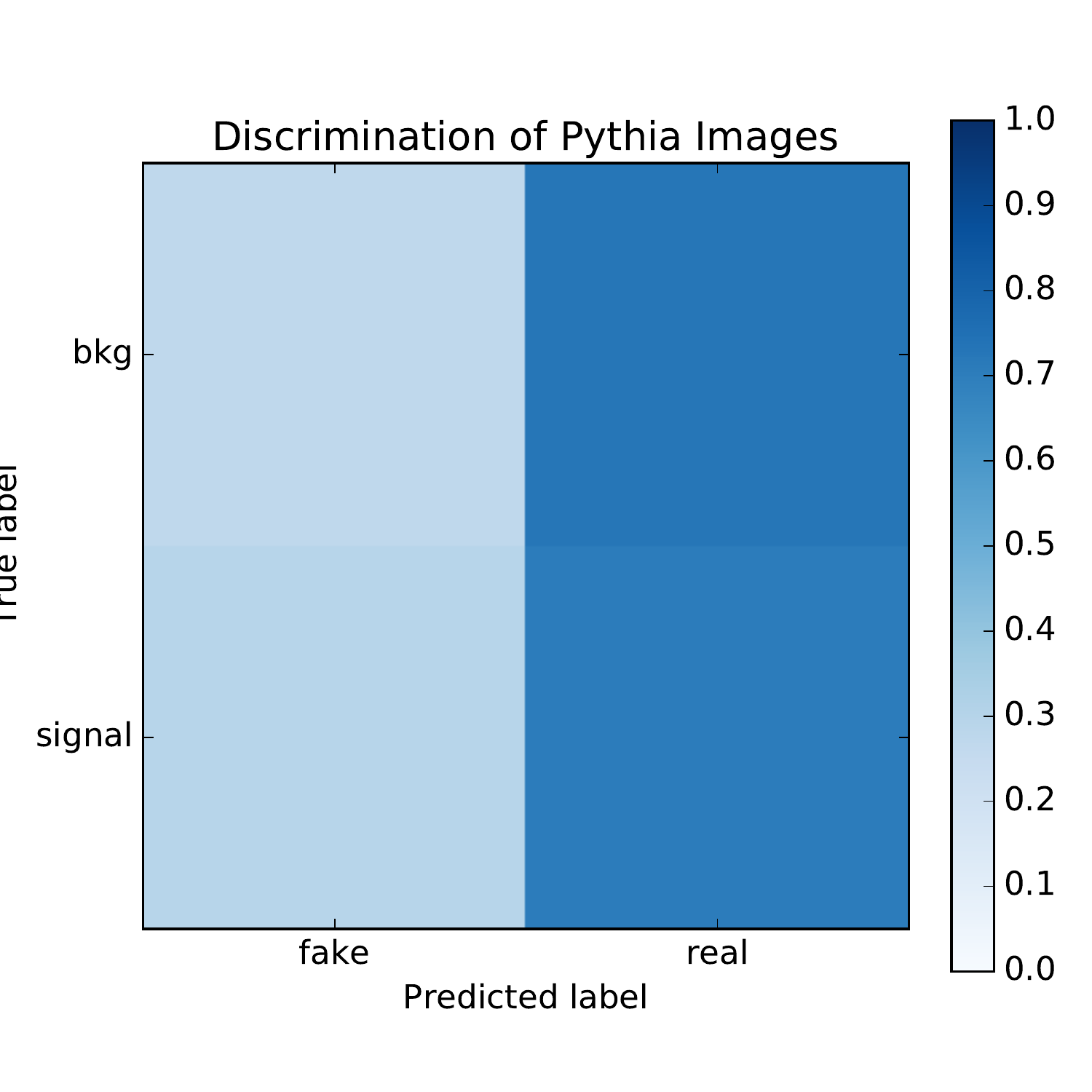}\hspace{2mm}\includegraphics[width=0.3\textwidth]{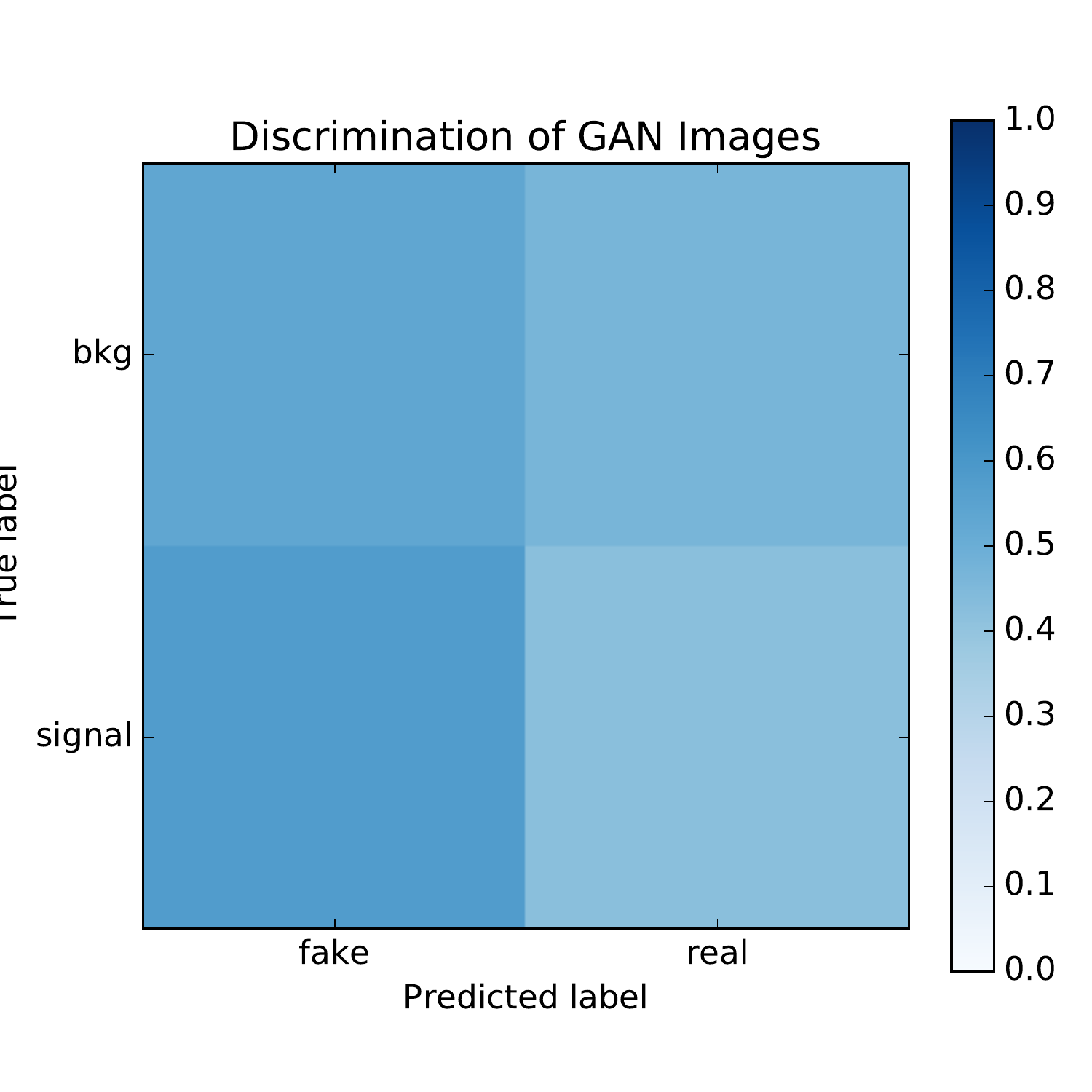}
\caption{We plot the normalized confusion matrices showing the correlation of the predicted $D$ output with the true physical process used to produce the images. The matrices are plotted for all images (left), Pythia images only (center) and GAN images only (right).}
\label{fig:confusion_fake_real}
\end{figure}

\newpage
\section{Image Pre-processing}
\label{app:image_process}
Reference~\cite{deOliveira:2015xxd} contains a detailed discussion on the the impact of image pre-processing and information content of the image. For example, it is shown that normalizing each image removes a significant amount of information about the jet mass.  One important step that was not fully discussed is the rotational symmetry about the jet axis. It was shown in Ref.~\cite{deOliveira:2015xxd} that a rotation about the jet axis in $\eta-\phi$ does not preserve the jet mass, i.e. $\eta_i\mapsto \cos(\alpha)\eta_i+\sin(\alpha)\phi_i,\phi_i\mapsto \cos(\alpha)\phi_i-\sin(\alpha)\eta_i$, where $\alpha$ is the rotation angle and $i$ runs over the constituents of the jet. One can perform a proper rotation about the $x$-axis (preserving the leading subjet at $\phi=0$) via 

\begin{align}
p_{x,i}&\mapsto p_{x,i}\\
p_{y,i}&\mapsto p_{y,i}\cos(\beta)+p_{z,i}\sin(\beta)\\
p_{z,i}&\mapsto p_{z,i}\cos(\beta)-p_{y,i}\sin(\beta)\\
E_i&\mapsto E_i,
\end{align}

\noindent where

\begin{align}
\beta = -\text{atan}(p_\text{y,translated subjet 2}/p_\text{z,translated subjet 2}) - \pi/2.
\end{align}

\begin{figure}[h!]
\centering
\includegraphics[width=0.45\textwidth]{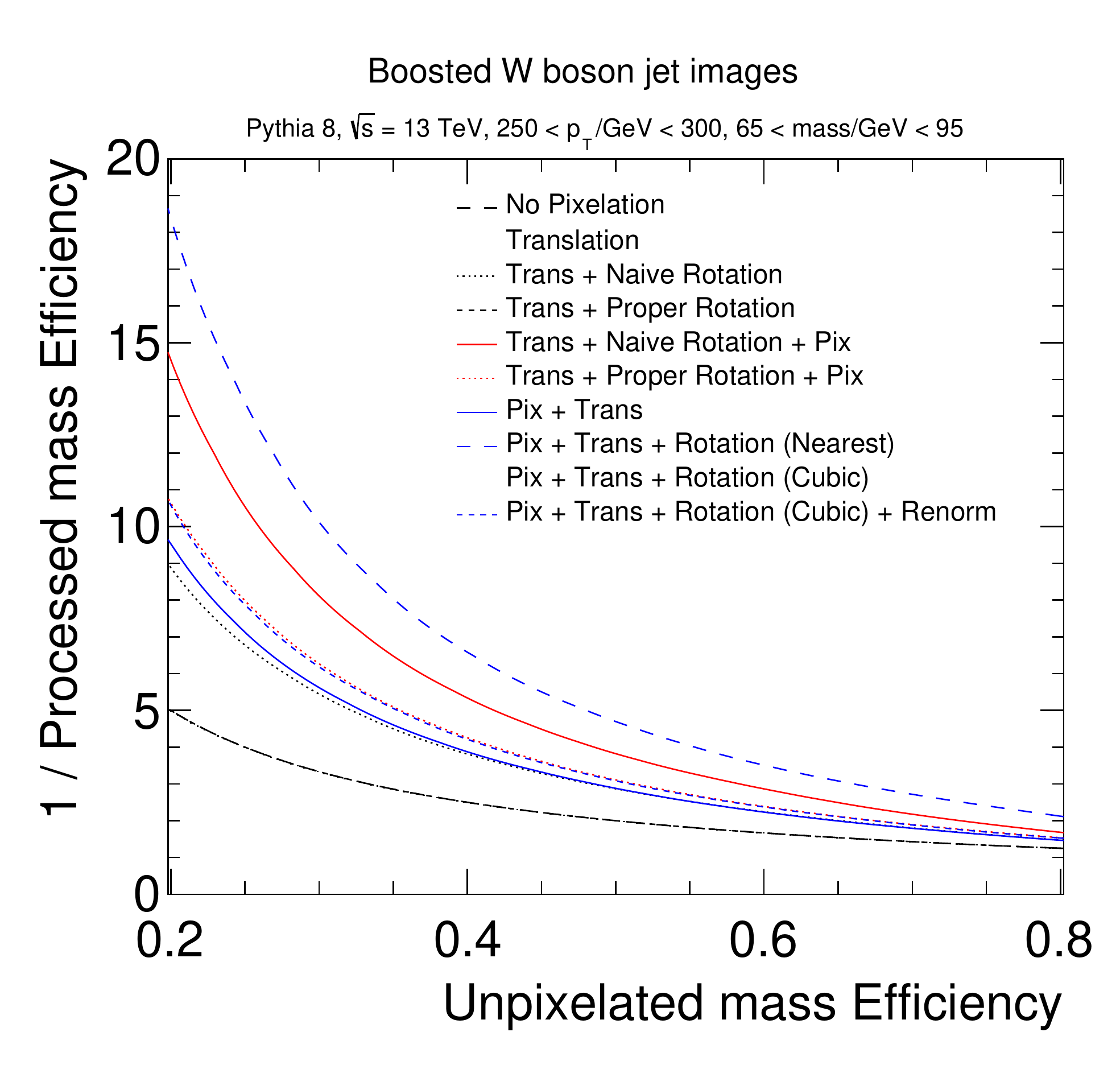}\includegraphics[width=0.45\textwidth]{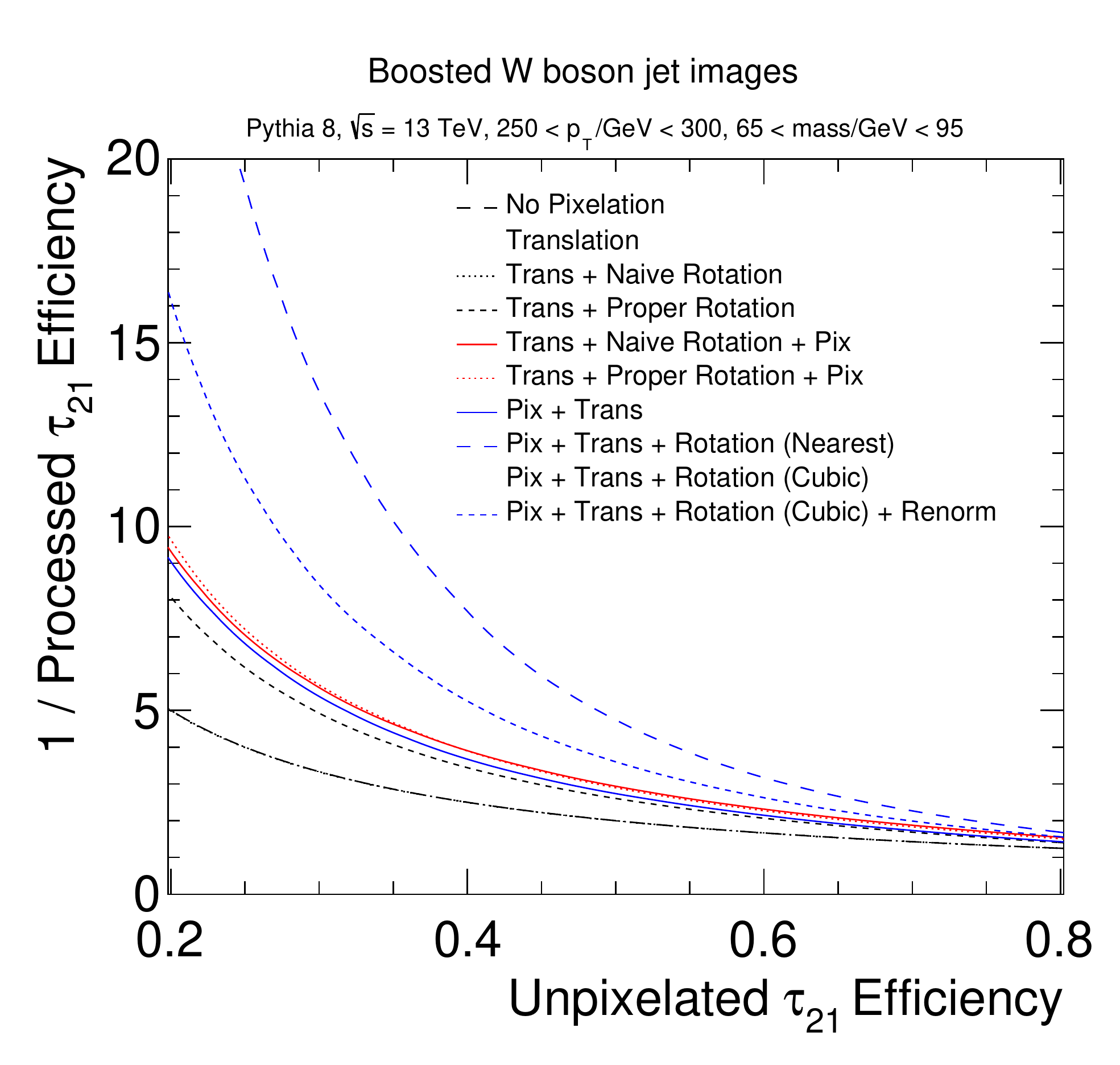}\\
\caption{These plots show ROC curves quantifying the information lost about the jet mass (left) or $n$-subjettiness (right) after pre-processing.  A preprocessing step that does not loose any information will be exactly at the random classifier line $f(x)=1/x$. Both plots use signal boosted $W$ boson jets for illustration.}
\label{fig:preprocessing:rocs}
\end{figure}

\begin{figure}[h!]
\centering
\includegraphics[width=0.45\textwidth]{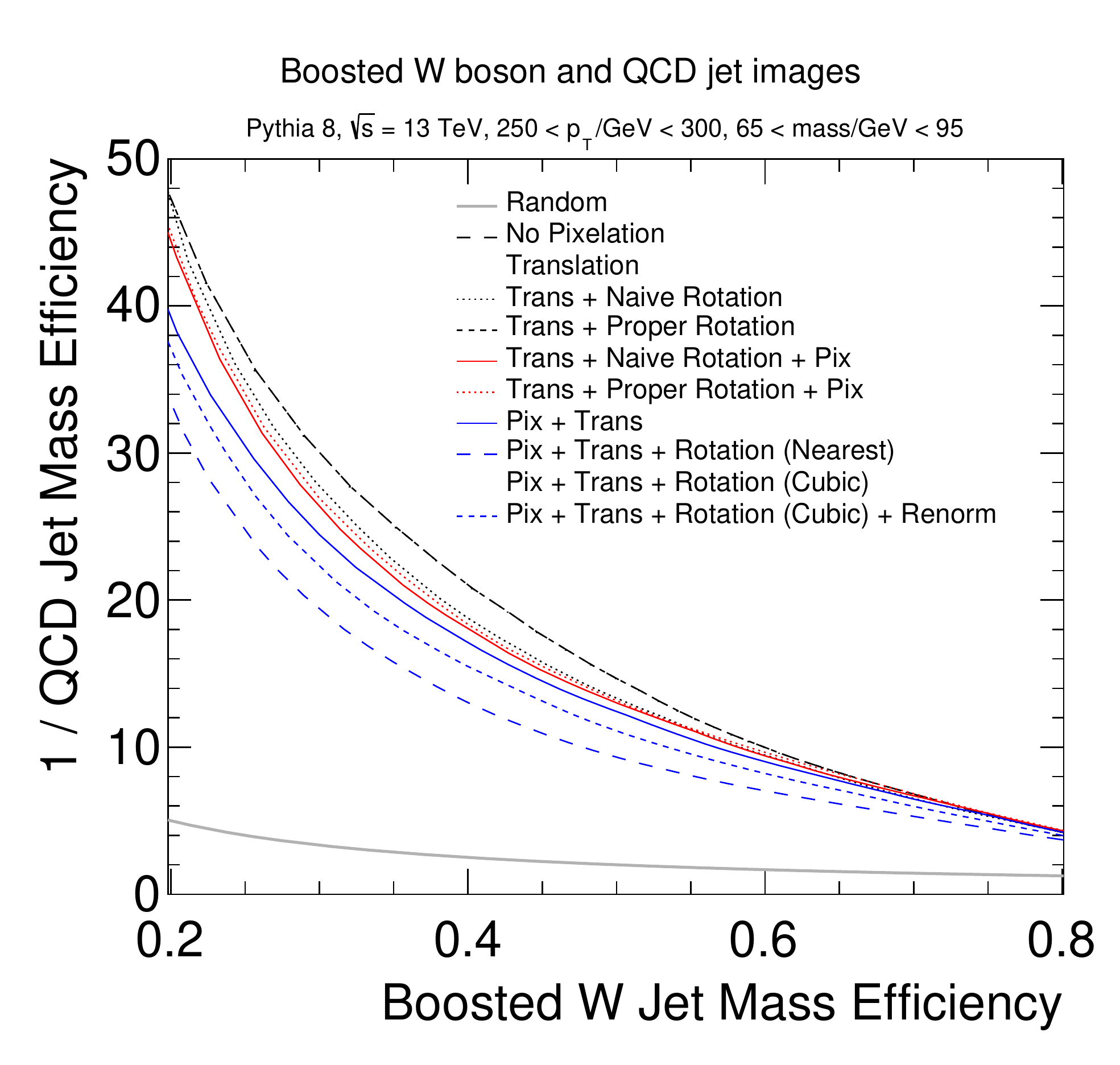}\includegraphics[width=0.45\textwidth]{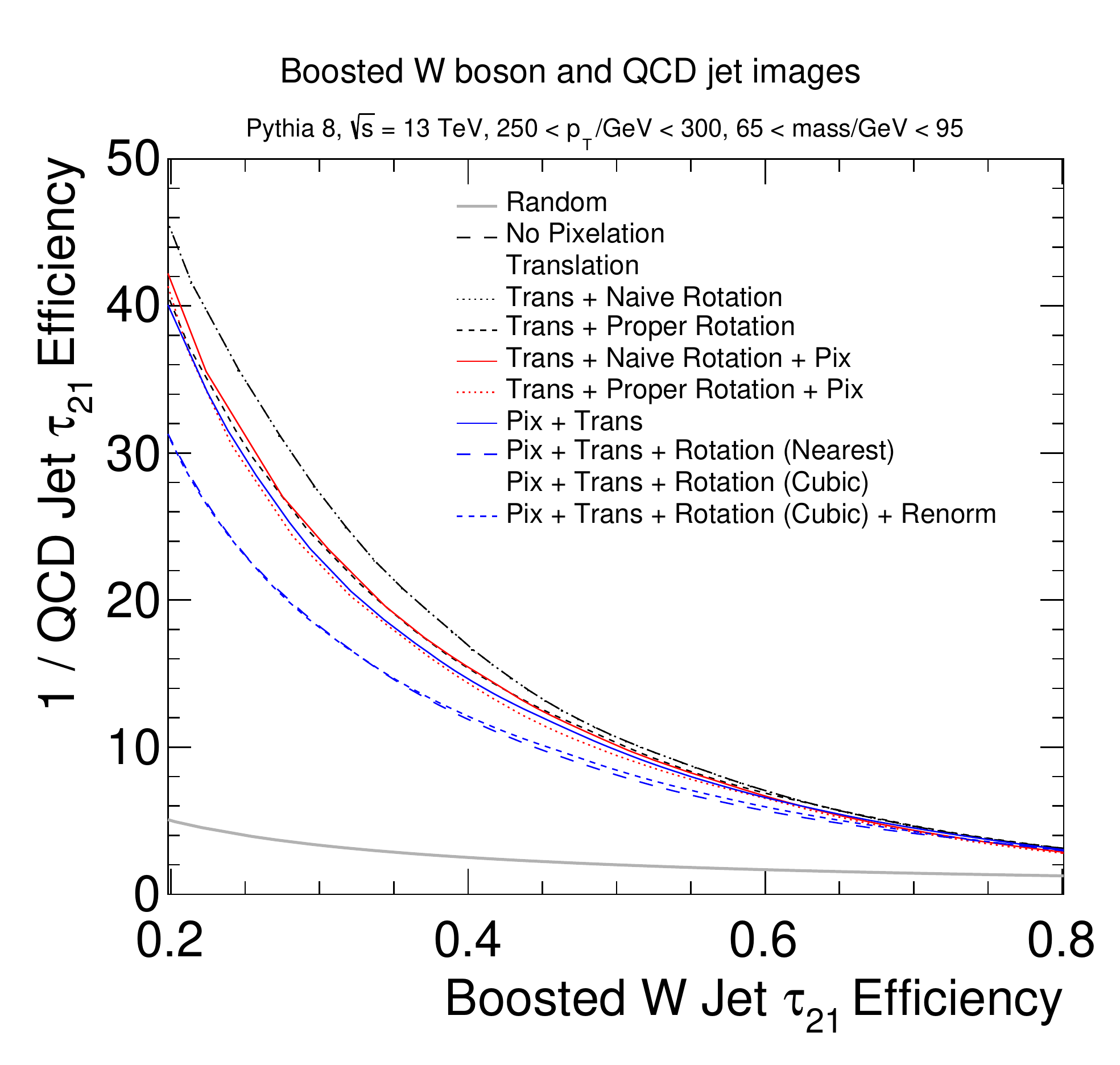}\\
\caption{ROC curves for classifying signal versus background based only on the mass (left) or $n$-subjettiness (right).  Note that in some cases, the preprocessing can actually improve discrimination (but always degrades the information content - see Fig.~\ref{fig:preprocessing:rocs}).}
\label{fig:preprocessing2:rocs}
\end{figure}

\noindent Figure~\ref{fig:preprocessing:rocs} quantifies the information lost by various preprocessing steps, highlighting in particular the rotation step.  A ROC curve is constructed to try to distinguish the preprocessed variable and the unprocessed variable.  If they cannot be distinguished, then there is no loss in information.  Similar plots showing the degradation in signal versus background classification performance are shown in Fig.~\ref{fig:preprocessing2:rocs}.  The best fully preprocessed option for all metrics is the {\it Pix+Trans+Rotation(Cubic)+Renorm}.  This option uses the cubic spline interpolation from Ref.~\cite{deOliveira:2015xxd}, but adds a small additional step that ensures that the sum of the pixel intensities is the same before and after rotation.  This is the procedure that is used throughout the body of the manuscript.

\clearpage
\newpage

\clearpage
\newpage

\bibliography{myrefs.bib}

\end{document}